\documentclass{article}

\usepackage[main, final]{neurips_2026}

\usepackage[utf8]{inputenc}
\usepackage[T1]{fontenc}
\usepackage{hyperref}
\usepackage{url}
\usepackage{booktabs}
\usepackage{amsfonts}
\usepackage{nicefrac}
\usepackage{microtype}
\usepackage{xcolor}
\usepackage{inconsolata}

\usepackage{graphicx}
\usepackage{amsmath}
\usepackage{multirow} 
\usepackage{booktabs}
\usepackage{enumitem}
\usepackage{marvosym}

\usepackage{amsfonts}
\usepackage{makecell} 
\usepackage{colortbl}
\usepackage{xcolor}

\usepackage{tabularx,adjustbox}
\newcolumntype{Y}{>{\raggedright\arraybackslash}X}
\usepackage{subcaption} 
\renewcommand{\arraystretch}{1.05}

\AtBeginDocument{%
  }

\usepackage{listings}
\usepackage{wrapfig}
\usepackage{placeins}
\title{Population-Aligned Persona Generation  for LLM-based Social Simulation}

\author{%
  \textbf{Zhengyu Hu$^{1,2}$ \quad Jianxun Lian$^{2}$\thanks{Corresponding author.} \quad Zheyuan Xiao$^{3}$ \quad Max Xiong$^{4}$} \\
  \textbf{Tianfu Wang$^{2}$ \quad Teng Xiao$^{1}$ \quad Yuxuan Lei$^{2}$ \quad Fengqing Jiang$^{1}$ \quad Kaize Ding$^{5}$} \\
  \textbf{Ziang Xiao$^{6}$ \quad Nicholas Jing Yuan$^{7}$ \quad Xing Xie$^{2}$ \quad Radha Poovendran$^{1}$} \\[0.5em]
  {\normalfont\small $^{1}$University of Washington \quad $^{2}$Microsoft Research Asia} \\
  {\normalfont\small $^{3}$The University of Texas at Austin \quad $^{4}$Duke University} \\
  {\normalfont\small $^{5}$Northwestern University \quad $^{6}$Johns Hopkins University \quad $^{7}$Microsoft}
}

\begin{document}

\maketitle

\begin{abstract}

Recent advances in large language models (LLMs) have enabled large-scale, high-fidelity social simulations, creating new opportunities for computational social science. However, constructing persona sets that faithfully reflect real-world population diversity remains a key challenge. Existing studies often emphasize agentic frameworks and simulation environments, while paying less attention to persona generation and the biases introduced by unrepresentative persona sets. In this paper, we propose a systematic framework for synthesizing high-quality, population-aligned persona sets for LLM-driven social simulation. Our approach begins by leveraging LLMs to generate narrative personas from long-term social media data, followed by rigorous quality assessment to filter out low-fidelity profiles. 
We then apply importance-based selection and optimal transport to align persona responses with human reference distributions, such as the Big Five personality trait distribution.
To address the needs of specific simulation contexts, we further introduce a task-specific module that adapts the globally aligned persona set to targeted subpopulations. Extensive experiments demonstrate that our method significantly reduces population-level bias and enables accurate, flexible social simulation for a wide range of research and policy applications.
\end{abstract}

\section{Introduction}

The growing capacity of large language models (LLMs) to emulate human-like cognitive traits, including aspects of theory of mind and social intelligence~\citep{Huang2024UnderstandingTP,binz2025foundation,strachan2024testing,lu2025tom}, makes them promising, albeit imperfect, tools for advancing social simulation in computational social science~\citep{Fulay2025TheEC,Karvonen2025RobustlyIL,Song2024PredictingUB}.
A central challenge in LLM-based social simulation is constructing persona sets whose \emph{population distribution} matches real-world diversity in traits, behaviors, and psychometrics. 
Yet most work emphasizes agentic frameworks and environments while simplifying persona construction: \citep{ZhangLMQLWLCLWXW25} samples 1{,}000 random Weibo users, \citep{Gao2023S3SS} uses three demographic attributes, and \citep{wang2025yulan} relies on LLM-generated or randomly sampled profiles. \citep{zhang2025socioverse} is more structured but is constrained to 15 demographic dimensions and assumes an \emph{a priori} target group distribution, which rarely holds. Overall, if the persona set is not representative, simulation results risk significant bias.

\begin{wrapfigure}[21]{r}{0.47\columnwidth}
    \vspace{-6pt}
    \centering
    \includegraphics[width=\linewidth]{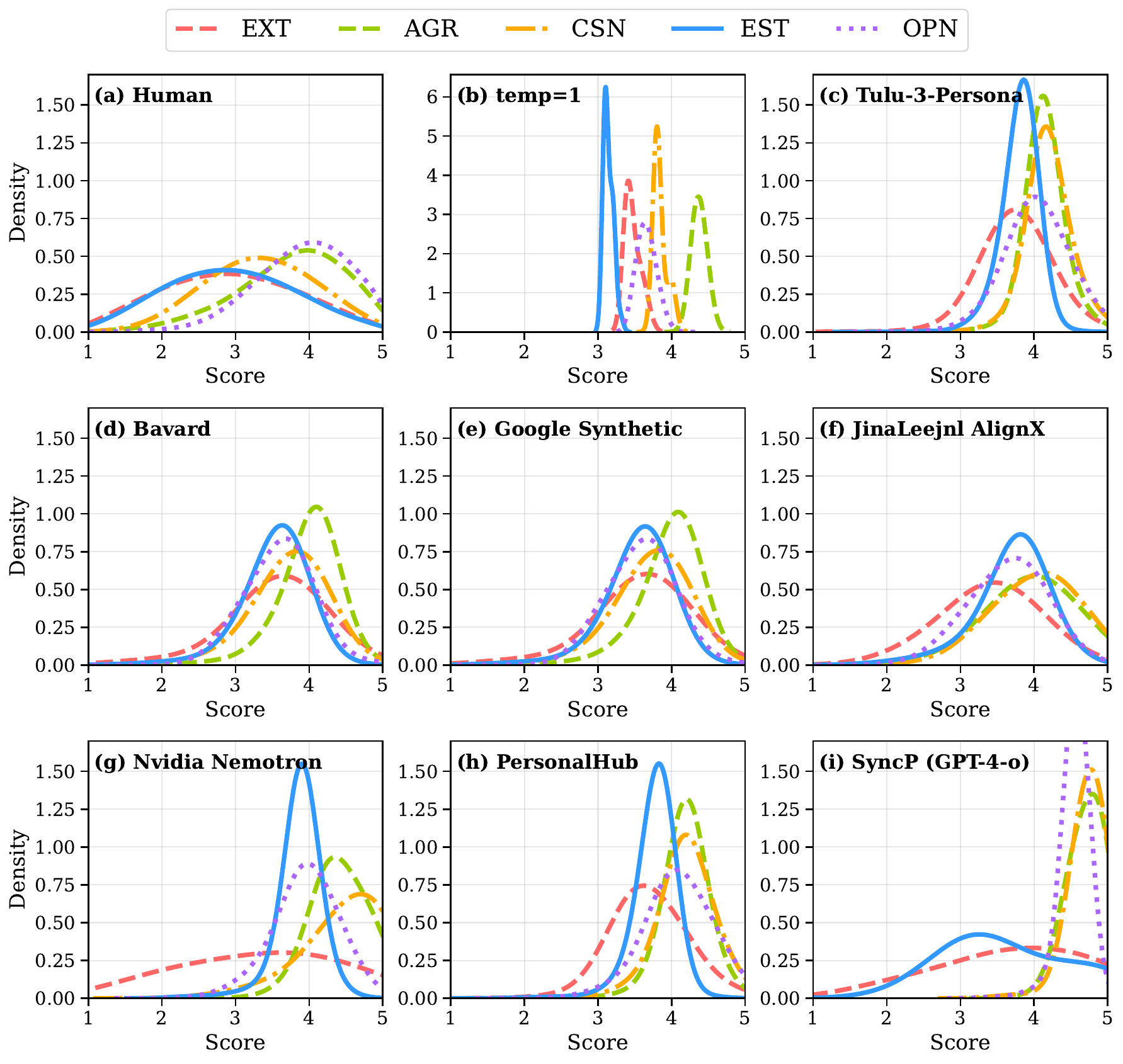}
    \vspace{-6pt}
    \caption{Comparison of Big Five personality trait distributions across real human data, LLM responses without personas, existing persona sets, and GPT-4o-generated personas.}
    \label{fig:intro_image}
    \vspace{-8pt}
\end{wrapfigure}

To illustrate the distortion in social simulation results compared to human data, we use Qwen2.5-72B to role-play humans completing the IPIP inventory~\citep{ipip50} and compare the Big Five personality distributions to a human reference from one million participants across 223 countries/regions~\citep{openpsychometrics2019} (Figure~\ref{fig:intro_image}a). A no-persona baseline at temperature 1.0 yields the distribution shown in Figure~\ref{fig:intro_image}b. We then evaluate six persona sets: Tulu-3-Persona~\citep{Lambert2024TLU3P}, Bavard~\citep{Zhang2018PersonalizingDA}, Google Synthetic~\citep{Jandaghi2023FaithfulPC}, AlignX~\citep{Li2025From1U}, Nvidia Nemotron~\citep{nvidia-Nemotron-Personas}, and PersonalHub~\citep{Chan2024ScalingSD} (Figure~\ref{fig:intro_image}c--h), along with a matched-size GPT-4o persona set at temperature 1.0 (Figure~\ref{fig:intro_image}i). While these persona sets improve over the random baseline, substantial population-level misalignment with real-world distributions persists, underscoring the need for faithful population-level simulation in LLM-based social modeling.

This paper provides a rigorous study on constructing faithful persona sets and examines their impact on social simulation outcomes. We propose a systematic framework for constructing high-quality personas aligned with real-world population distributions, built on three principles. First, we follow the data processing pipeline of HumanLLM~\citep{Lei2026HumanLLMTP} to derive narrative personas from long-term social media histories across blogs~\citep{schler2006effects}, Twitter~\citep{enryu-twitter100m-tweets}, and Reddit~\citep{huggingfacegeclm-reddit-submissions}, capturing nuanced user characteristics beyond rigid demographic schemas. Second, we introduce a distributional alignment strategy that matches persona-generated responses to human references through importance-based selection and optimal transport, using one million IPIP Big Five responses as our global benchmark~\citep{ipip50,goldberg1992development,openpsychometrics2019}. Third, we enable flexible adaptation to subpopulations via embedding-based retrieval and LLM revision, ensuring practical applicability across diverse scenarios. 
Experimental results demonstrate that our framework substantially reduces distributional bias across seven standard evaluation tasks while maintaining strong behavioral consistency in downstream simulations.
To summarize,  our contributions include:

\begin{itemize}
[leftmargin=1.5em, topsep=2pt, itemsep=2pt, parsep=0pt, partopsep=0pt]

    \item We emphasize the importance of population-level persona alignment for LLM-based social simulation, and propose a systematic framework to generate persona sets that accurately reflect real-world population distributions.

    \item We design a two-stage
    alignment method combining Importance-based Selection and Optimal Transport for global distribution matching.

    \item We develop a persona targeting module that supports group-specific persona alignment through embedding retrieval and LLM revision, thereby facilitating its practical application in diverse social science scenarios.

    \item  Through comprehensive evaluation across seven diverse evaluation tasks, we demonstrate that our framework outperforms existing persona sets in distributional alignment, behavioral consistency, and group-specific adaptation.

\end{itemize}

\section{Related Work}

\noindent \textbf{LLM-based Simulation for Social Science.}
LLMs are used to simulate human behavior for analyzing attitudes, preferences, and decision-making~\citep{Park2023GenerativeAI,Chuang2023SimulatingOD,Wang2024SimulatingHD,zhang2025socioverse,wang2025personality, Wang2025LLMpoweredMF}, with most work emphasizing individual-level fidelity (e.g., personalized dialogue user simulation~\citep{Shao2023CharacterLLMAT}, persona style transfer~\citep{Kong2023BetterZR}, reasoning-trait modeling~\citep{Chen2024FromPT}, persona-aware metrics~\citep{Lu2024LargeLM}, and CharacterEval-style role-playing benchmarks~\citep{Tu2024CharacterEvalAC}).
Group-level phenomena have also been studied, including polarization~\citep{Piao2025EmergenceOH}, partisan behaviors~\citep{Chuang2023EvaluatingLA}, and role-playing ordinary individuals~\citep{Ng2024HowWC}.
\citep{wu2025llm} show that LLM simulations often collapse to an ``average persona,'' reducing heterogeneity and emergent group dynamics, and argue for population-level anchoring; yet most methods lack alignment enforcement, leading to biased outcomes.

\noindent \textbf{Persona Modeling and Evaluation.}
Personas are widely used to model agent traits and behaviors in dialogue~\citep{Hong2023MetaGPTMP,Wang2023UnleashingTE}, user modeling~\citep{Li2024PersonalLA,Kaur2025SyntheticVE}, and social simulation~\citep{Cheng2023CoMPosTCA,Hu2025SimulatingRS,Ju2025TrajLLMAM}, with generation ranging from manual curation to LLM-based automation.
Early work encoded speaker traits for consistent dialogue~\citep{Li2016APN} and later conditioned responses on self/partner profiles to improve engagement~\citep{Zhang2018PersonalizingDA}; recent methods use LLMs to simulate historical/fictional characters~\citep{Shao2023CharacterLLMAT} and guide profile generation for personalization~\citep{Zhang2024GuidedPG}.
However, population-aligned persona set construction remains scarce, risking biased simulation outcomes.

\section{Population-Aligned Persona Generation Framework}

\begin{figure*}[h]
\centering
\captionsetup{skip=2pt} 
\includegraphics[width=1.01\linewidth]{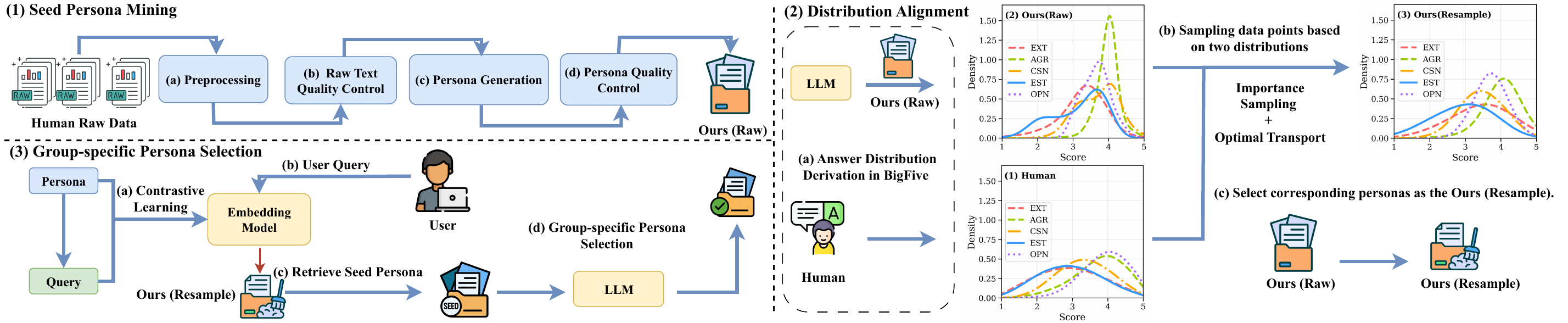}
\caption{
Overview of the Population-Aligned Persona Generation Framework. 
}
\label{fig:framework}
\end{figure*}

As shown in Fig.~\ref{fig:framework}, our framework consists of three key components: (1) \textit{Seed Persona Mining}, which extracts and filters high-quality narrative personas from large-scale human-authored corpora; (2) \textit{Global Distribution Alignment}, which uses a two-stage sampling strategy, Importance-based Selection and Optimal Transport, to align persona distributions with human data; and (3) \textit{Group-specific Persona Construction}, where we support targeted persona generation for specific  tasks.

\subsection{Seed Persona Mining}
\label{subsec:SPG}

We use the Reddit, Twitter, and Blogger subsets of HumanLLM's Cognitive Genome Dataset~\citep{Lei2026HumanLLMTP} and follow its data processing pipeline for seed persona mining. The underlying corpora are the \textit{Blog Authorship Corpus}~\citep{schler2006effects} (681K posts), \textit{Reddit}~\citep{huggingfacegeclm-reddit-submissions} (47.2M submissions from 50 subreddits), and \textit{Twitter}~\citep{enryu-twitter100m-tweets} (88.1M original tweets from CommonCrawl). Together, these sources provide diverse narratives across demographic and behavioral dimensions. The mining pipeline includes four stages:
\vspace{0.3em}

\noindent  \textbf{Preprocessing.} We clean each corpus to ensure data authenticity and narrative richness. For Reddit, we retain only real users with posts satisfying: $\text{score} \ge 2$, $\text{num\_comments} \ge 2$, and length $\ge 300$ characters. For the blog corpus, we discard entries under 30 tokens or lacking first-person pronouns. For Twitter, we remove retweets, short tweets ($<$300 characters), and likely corporate or promotional content. All remaining texts are rewritten using {Llama-3.3-70B-Instruct} to remove HTML tags, URLs, non-English content, and formatting noise. This results in approximately 4M Reddit posts, 500K blog posts, and 1.7M tweets.
\vspace{0.3em}

\noindent \textbf{Raw Text Quality Control.} Each rewritten post is annotated using {Llama-3.3-70B-Instruct} with two labels: a three-level quality score (high, medium, low), and a binary harmfulness label. High-quality content is expected to reflect personal experience, emotions, or social behavior, while excluding boilerplate or ad-like content. Harmful content (e.g., violence, privacy violations) is filtered out. Only posts rated both high-quality and harmless are retained, resulting in final counts of 2.8M Reddit, 368K blog, and 673K Twitter entries.
\vspace{0.3em}

\noindent \textbf{Persona Generation.} We aggregate posts by author and discard users with fewer than six posts. Each user’s posts are concatenated into a single text block $s$, which is passed to {Llama-3.3-70B-Instruct} to generate a grounded persona summary $p_i = \mathcal{G}(s)$,
where the model is prompted to cite evidence from $s$ for factual grounding.

\vspace{0.2em}

\noindent \textbf{Persona Quality Control.} To ensure quality, we use a critic LLM to score each persona $p_i$ on five dimensions: hallucination (faithfulness to source text), coverage (salience of user traits), conciseness (well-formed and within 100 words), relevance (exclusion of generic or promotional content), and overall quality. Each score ranges from 1 to 10. We retain personas with an overall score $>$8 and all other dimensions $>$7. This step filters out approximately 30\% of personas from our source.

The resulting seed pool contains over 160,000 high-fidelity narrative personas covering diverse psychometric and demographic attributes. Appendix~\ref{app:dataproc} provides stage-wise data statistics, filtering thresholds, and prompts; Appendix~\ref{app:curation_audit} reports human evaluations of rewriting and persona generation.

\subsection{Global Distribution Alignment via Two-stage Sampling}
\label{subsec:IS_OT}

Let $\mathcal{P}=\{p_i\}_{i=1}^N$ be the high-quality personas (Section~\ref{subsec:SPG}), and $\mathcal{Q}=\{q_k\}_{k=1}^d$ a set of prompts (e.g., Big Five items). 
Given a language model $\texttt{LLM}$, the response to a persona–question pair is
\begin{equation}
x_{ik} := \texttt{LLM}(p_i \,\|\, q_k),
\end{equation}
with $x_i=(x_{i1},\ldots,x_{id})\in\mathbb{R}^d$. 
The set $\{x_i\}_{i=1}^N$ induces an empirical distribution $r_{\texttt{persona}}$. 
Let $Y=\{y_j\}_{j=1}^M\subset\mathbb{R}^d$ denote human responses, with empirical distribution $r_{\texttt{human}}$.
In this work, we use large-scale human response data from the widely used IPIP Big Five inventory~\citep{ipip50,goldberg1992development,openpsychometrics2019}, spanning one million individuals across 223  countries and regions, as a representative reference for real-world personality distributions.

Our goal is to select $\mathcal{P}'\subseteq\mathcal{P}$ whose induced distribution matches $r_{\texttt{human}}$. 
We use a two-stage sampler: \emph{Importance-based Selection} (IS) retains candidates with the highest human-to-persona density ratios; \emph{Optimal Transport} (OT) then assigns sampling weights based on their transport costs to human responses.
\vspace{0.5em}

\noindent \textbf{Stage 1: Importance-based Selection (IS).}
In the first stage, our goal is to efficiently select a subset of personas whose response distributions significantly overlap with the human distribution. To achieve this, we estimate the densities of $r_{\texttt{human}}$ and $r_{\texttt{persona}}$ via Gaussian-kernel KDE with bandwidth $h$:
\begin{equation}
\begin{split}
\hat{r}_{\texttt{human}}(x) 
&= \frac{1}{M} \sum_{j=1}^{M} \mathcal{N}(x - y_j ; h^2 I), \\[6pt]
\hat{r}_{\texttt{persona}}(x)
&= \frac{1}{N} \sum_{i=1}^{N} \mathcal{N}(x - x_i ; h^2 I),
\end{split}
\end{equation}
where \( I \in \mathbb{R}^{d \times d} \) and \(\mathcal{N}(z; h^2 I)\) is a zero-mean multivariate Gaussian with covariance \( h^2 I \).
Using these estimates, we assign each persona response \( x_i \) an importance weight as the ratio of estimated densities: 
$w_i^{\mathrm{IS}} 
=
\frac{ \hat{r}_{\texttt{human}}(x_i) }
     { \hat{r}_{\texttt{persona}}(x_i) }.
$
This quantifies how much more likely \( x_i \) is under the human distribution than under the original persona distribution. 
Hence, larger \(w_i^{\mathrm{IS}}\) indicate stronger alignment with human responses.
We rank personas by decreasing importance weight. Let $i_{(1)},\ldots,i_{(N)}$ satisfy $w_{i_{(1)}}^{\mathrm{IS}}\ge\cdots\ge w_{i_{(N)}}^{\mathrm{IS}}$. We retain the highest-weight 70\%:
\begin{equation}
N^{\dagger}=\lfloor 0.7N\rfloor,
\qquad
\mathcal{P}^{\dagger}=\{p_{i_{(k)}}:k=1,\ldots,N^{\dagger}\}.
\end{equation}
The corresponding response vectors form $X^{\dagger}=\{x_i:p_i\in\mathcal{P}^{\dagger}\}$.

\vspace{0.5em}

\noindent \textbf{Stage 2: Optimal Transport (OT).}
While IS yields a strong candidate pool, it does not guarantee fine-grained structural alignment with human responses. We therefore solve an entropic OT problem that explicitly accounts for prompt-level importance and matches IS candidates to human data.

Let $\mathcal{P}^{\dagger}$ be the IS candidate set with size $N^{\dagger}$, and let $x_i^{\dagger}\in\mathbb{R}^d$ (persona) and $y_j\in\mathbb{R}^d$ (human) denote response vectors over prompts $\mathcal{Q}=\{q_k\}_{k=1}^{d}$. Assign positive prompt weights $\{\omega_k\}_{k=1}^{d}$ and define the pairwise cost
\begin{equation}
C_{ij}=\sum_{k=1}^{d}\omega_k\bigl(x_{ik}^{\dagger}-y_{jk}\bigr)^2.
\end{equation}
We compute an entropic OT plan
\begin{equation}
\scalebox{0.88}{$\displaystyle
\Gamma^{\star}=\arg\min_{\Gamma\in\Pi(\mathbf{a}^{\dagger},\mathbf{b})}\langle C,\Gamma\rangle
+\varepsilon\sum_{i,j}\Gamma_{ij}(\log\Gamma_{ij}-1),
$}
\end{equation}
where $\langle C,\Gamma\rangle=\sum_{i,j}C_{ij}\Gamma_{ij}$ is the Frobenius inner product; $\varepsilon>0$ controls plan entropy; $\Gamma\in\mathbb{R}_{\ge 0}^{N^{\dagger}\times M}$ transports mass between $N^{\dagger}$ candidates and $M$ human samples; $\mathbf{a}^{\dagger}\in\Delta^{N^{\dagger}-1}$ and $\mathbf{b}\in\Delta^{M-1}$ are uniform marginals over candidates and humans, respectively; and $\Pi(\mathbf{a}^{\dagger},\mathbf{b})=\{\Gamma:\Gamma\mathbf{1}=\mathbf{a}^{\dagger},\Gamma^{\top}\mathbf{1}=\mathbf{b}\}$. We use the Sinkhorn–Knopp algorithm to obtain $\Gamma^{\star}$.
For each candidate, let $a_i^{\dagger}=\sum_j\Gamma_{ij}^{\star}=1/N^{\dagger}$ and define its expected transport cost as $\bar{c}_i=(a_i^{\dagger})^{-1}\sum_j\Gamma_{ij}^{\star}C_{ij}$. We set $w_i^{\mathrm{OT}}=\exp(-\bar{c}_i/\tau)$ with temperature $\tau>0$, then normalize and sample to produce the aligned subset $\mathcal{P}'$:
\begin{equation}
\scalebox{0.65}{$\displaystyle
\pi_i^{\mathrm{OT}} = \frac{w_i^{\mathrm{OT}}}{\sum_{k=1}^{N^{\dagger}} w_k^{\mathrm{OT}}}, 
\qquad
\mathcal{P}' \sim \mathrm{Multinomial}\!\left(N', \{\pi_i^{\mathrm{OT}}\}_{i=1}^{N^{\dagger}}\right).
$}
\end{equation}
Here, $N'$ is the target number of personas in the final OT-aligned set.
Implementation details for both sampling stages are provided in Appendix~\ref{asb:DetailOfTwoStage}.

\subsection{Group-specific Persona Set Construction}
\label{subsec:PSM}

To support adaptive construction of group-specific persona subsets, we train a dedicated embedding model that embeds natural language queries and personas into a shared semantic space. For training data construction, we prompt a LLM~\footnote{All LLMs used in this section are Qwen2.5-72B.}
 to generate descriptive queries from each persona $p_i \in \mathcal{P}'$:
\begin{equation}
q_i = \texttt{LLM}(p_i),
\end{equation}
where each $q_i$ is a natural language description semantically aligned with $p_i$. These $(q_i, p_i)$ pairs serve as positive samples.
To construct negative samples, we first compute  similarity between $q_i$ and all other personas, and select the top-$N$ most similar candidates. 
We also sample $N$ random personas. 
Since both the top-$N$ most similar candidates and the $N$ randomly sampled personas may contain false negatives (i.e., personas that actually match the query), we further use an LLM to filter out any candidates that are semantically aligned with the query.
The embedding model is trained using the following loss:
\begin{equation}
\mathcal{L} = -\log \frac{\exp(\mathrm{sim}(e_q, e_p^+))}{\exp(\mathrm{sim}(e_q, e_p^+)) + \sum_{p^- } \exp(\mathrm{sim}(e_q, e_p^-))},
\label{eq:contrastive}
\end{equation}
where $e_q$ and $e_p$ are the embeddings of the query and persona respectively, and $\mathrm{sim}(\cdot, \cdot)$ denotes cosine similarity.

At inference time, given a user query $q$ (e.g., ``first-year college students in the United States''), we compute its embedding $e_q = \mathrm{Embed}(q)$ and measure similarity against all persona embeddings $\{e_i\}$ in $\mathcal{P}'$. We then select the top-$K$ most similar personas based on cosine similarity to form a seed subset $\mathcal{P}_{\text{seed}}(q)$. This subset provides semantically relevant exemplars for further persona generation.
To generate high-fidelity group-specific personas, we prompt a backbone LLM using the original query and the retrieved seed subset. Specifically, we format the input as a concatenation of $q$ and the selected personas in $\mathcal{P}_{\text{seed}}(q)$, and generate new samples as follows:
\begin{equation}
\mathcal{P}_{\text{group}}(q) = \texttt{LLM}(q \, \| \, \mathcal{P}_{\text{seed}}(q)).
\end{equation}
The resulting set $\mathcal{P}_{\text{group}}(q)$ aligns with user intent, making it applicable to downstream social simulations constrained by specific demographic or behavioral parameters.

\vspace{-2mm}

\section{Experimental Setup}
\label{sec:Exp_setup}

\vspace{-2mm}
\noindent \textbf{Implementation Details.}
Unless stated otherwise, Sections~\ref{sec:population-level}, ~\ref{subsec:individual_consistency} and~\ref{sec:SocSci} follow a unified evaluation setup. 
We use IPIP Big Five~\citep{ipip50} as the global alignment reference because it contains over one million real human responses across 223 countries/regions and captures domain-general personality dimensions; Appendix~\ref{app:why_ipip} discusses the associated in-domain evaluation concern.
For \textit{Without Persona}, each base model answers the questionnaire directly ({DAns}) at temperatures $\{0.0,0.7,1.0\}$, with 5{,}000 samples per temperature to capture sampling variance. 
For \textit{With Persona}, we standardize all persona sets to 5{,}000 personas. 
We consider (i) {SyncP}-type baselines, which synthesize 5{,}000 personas using the corresponding model (e.g., {SyncP (GPT-4o)} at temperature 1.0), and (ii) public persona sets, including {Tulu-3-Persona}~\citep{Lambert2024TLU3P}, {Bavard}~\citep{Zhang2018PersonalizingDA}, {Google Synthetic}~\citep{Jandaghi2023FaithfulPC}, {AlignX}~\citep{Li2025From1U}, {Nvidia Nemotron}~\citep{nvidia-Nemotron-Personas}, and {PersonalHub}~\citep{Chan2024ScalingSD}, from which we uniformly subsample 5{,}000 personas.
To isolate the effect of distribution alignment (vs.\ data source and size), we additionally report {Ours (Raw)} (the full pool from Section~\ref{subsec:SPG}), {Ours (RandomSelect)} (a size-matched random subset from our pool), and {Ours (Resample)} (the aligned set produced by the two-stage sampler in Section~\ref{subsec:IS_OT}, also size-matched to 5{,}000). 
Section~\ref{subsec:trait_vis} follows the same in-domain setup as Section~\ref{sec:population-level}. 
In Section~\ref{subsec:PSM}, we set $N{=}10$. 
For group-specific evaluation (Section~\ref{subsec:group_specific}), we use {Qwen3-Embedding-0.6B}~\citep{qwen3embedding} and further include {Ours (EM w/o train)} (retrieval from {Resample} with an untrained encoder) and {Ours (EM w/ train)} (a contrastively trained encoder with positives and filtered hard negatives as in Eq.~\ref{eq:contrastive}).

\noindent \textbf{Evaluation Datasets.}
We evaluate on seven different datasets: IPIP Big Five~\citep{ipip50}, CFCS~\citep{Strathman1994CFC}, FBPS~\citep{OpenPsychometrics2019Firstborn}, Duckworth (Grit)~\citep{Duckworth2007Grit}, WVS~\citep{Inglehart2020WVS},  YRBSS~\citep{Brener2024OverviewAM} and SocSci~\citep{Kolluri2025FinetuningLF}. Detailed descriptions are in Appendix~\ref{abs:dataset}.
All evaluation datasets use real human responses as evaluation references, rather than synthetic or LLM-generated labels. Detailed dataset statistics are provided in Table~\ref{tab:dataset-stats}.

\noindent \textbf{Evaluation Metrics.}
We assess population-level alignment using Averaged Monotonic Wasserstein (AMW)~\citep{villani2008optimal,wang2024statistical}, Fr\'echet Distance (FD)~\citep{veeramacheneni2023fr,jayasumana2024rethinking}, and Sliced Wasserstein (SW)~\citep{nguyen2022revisiting,lobashev2025color}. We additionally report Maximum Mean Discrepancy (MMD)~\citep{kalinke2022maximum,ren2025few} for IPIP, CFCS, FBPS, and Duckworth, and Energy Distance (ED) for WVS and YRBSS. For individual-level consistency, we compute Pearson correlations across all trait pairs and report the mean absolute error (MAE) between synthetic and human correlation matrices~\citep{pearson_wikipedia,pratelli2025evaluating}.
For Section~\ref{sec:SocSci}, we compute  {Accuracy} and  {Distribution}  Score following~\citep{Kolluri2025FinetuningLF}. Formal definitions appear in Appendix~\ref{app:metrics}. Repeated-run results and significance comparisons appear in Appendix~\ref{app:repeated_runs}.

\section{Results and Analysis}
\label{sec:results}

\vspace{-2mm}

\subsection{Population Level}
\label{sec:population-level}
\begin{table*}[h]
\caption{
In-domain alignment results on the IPIP Big Five dataset for three base models across four metrics
{Lower is better for all metrics.} Best results are \textcolor{red}{red}, second-best \textcolor{blue}{blue}.
}

\resizebox{\linewidth}{!}{

\begin{tabular}{@{}ll|cccc|cccc|cccc|c@{}}
\toprule
\multicolumn{2}{l|}{\textbf{Model Base}}                                                         & \multicolumn{4}{c|}{\textbf{Qwen2.5-72B}}               & \multicolumn{4}{c|}{\textbf{Llama-3-70B}}               & \multicolumn{4}{c|}{\textbf{Phi-4}}                     & \multicolumn{1}{l}{\textbf{Avg.}} \\ \midrule
                                                                           & \textbf{}           & \textbf{AMW} & \textbf{FD} & \textbf{SW} & \textbf{MMD} & \textbf{AMW} & \textbf{FD} & \textbf{SW} & \textbf{MMD} & \textbf{AMW} & \textbf{FD} & \textbf{SW} & \textbf{MMD} & \multicolumn{1}{l}{\textbf{}}     \\ \midrule
\multirow{3}{*}{\begin{tabular}[c]{@{}l@{}}Without\\ Persona\end{tabular}} & DAns (temp=0)       & 0.4111       & 1.4256      & 0.4112      & 0.8631       & 0.6469       & 3.4211      & 0.6962      & 2.1904       & 0.4560       & 1.7358      & 0.4602      & 1.1207       & 1.1532                            \\
                                                                           & DAns (temp=0.7)     & 0.3860       & 1.2309      & 0.3742      & 0.6969       & 0.6271       & 3.0539      & 0.6611      & 1.8805       & 0.3909       & 1.3098      & 0.3960      & 0.8257       & 0.9861                            \\
                                                                           & DAns (temp=1)       & 0.3687       & 1.0947      & 0.3560      & 0.6250       & 0.5946       & 2.9176      & 0.6422      & 1.7867       & 0.3731       & 1.1475      & 0.3744      & 0.7294       & 0.9175                            \\ \midrule
\multirow{10}{*}{\begin{tabular}[c]{@{}l@{}}With\\ Persona\end{tabular}}   & SyncP (Qwen2.5-72B) & 0.2639       & 0.5854      & 0.2792      & 0.3193       & 0.3124       & 0.8279      & 0.3102      & 0.4749       & 0.3284       & 0.8232      & 0.3541      & 0.4282       & 0.4423                            \\
                                                                           & SyncP (Llama-3-70B) & 0.2759       & 0.5832      & 0.2679      & 0.3392       & 0.3996       & 0.8307      & 0.3472      & 0.4421       & 0.3336       & 0.8268      & 0.3383      & 0.4886       & 0.4561                            \\
                                                                           & SyncP (Phi-4)       & 0.2850       & 0.7975      & 0.3363      & 0.4316       & 0.3884       & 1.1620      & 0.4011      & 0.7135       & 0.3357       & 0.9425      & 0.3609      & 0.5767       & 0.5609                            \\
                                                                           & SyncP (GPT-4o)     & 0.2467       & 0.5824      & 0.2679      & 0.2701       & 0.3523       & 0.7263      & 0.3387      & 0.4476       & 0.3249       & 0.7500      & 0.3175      & 0.4147       & 0.4199                            \\
                                                                           & Tulu-3-Persona      & 0.2821       & 0.8400      & 0.3250      & 0.4631       & 0.3276       & 0.9418      & 0.3512      & 0.5941       & 0.2814       & 0.6412      & 0.2788      & 0.3593       & 0.4738                            \\
                                                                           & Bavard              & 0.3069       & 0.7838      & 0.3174      & 0.4669       & 0.3173       & 0.6659      & 0.3476      & 0.3758       & 0.1877       & 0.3759      & 0.2028      & 0.2414       & 0.3825                            \\
                                                                           & Google Synthetic    & 0.3135       & 0.8081      & 0.3246      & 0.4880       & 0.2983       & 0.6715      & 0.3286      & 0.3684       & 0.1878       & 0.3729      & 0.2027      & 0.2394       & 0.3837                            \\
                                                                           & AlignX              & 0.3416       & 0.9393      & 0.3567      & 0.5606       & 0.3319       & 0.6772      & 0.3274      & 0.3481       & 0.1641       & 0.3215      & 0.2163      & 0.2306       & 0.4013                            \\
                                                                           & Nvidia Nemotron     & 0.2645       & 0.8316      & 0.3199      & 0.4414       & 0.3615       & 0.8005      & 0.3177      & 0.4497       & 0.2233       & 0.5119      & 0.2412      & 0.2421       & 0.4171                            \\
                                                                           & PersonalHub         & 0.2982       & 0.9167      & 0.3436      & 0.5303       & 0.3365       & 0.8011      & 0.3268      & 0.5001       & 0.2469       & 0.5620      & 0.2609      & 0.3112       & 0.4529                            \\ \midrule
\multirow{3}{*}{Ours}                                                      
                                                                           & Raw                 & \textcolor{blue}{0.2215}       & 0.4630      & \textcolor{blue}{0.2236}      & \textcolor{blue}{0.2387}       & \textcolor{blue}{0.2693}       & \textcolor{blue}{0.5671}      & \textcolor{blue}{0.2737}      & 0.3354       & \textcolor{blue}{0.1185}       & 0.2385      & 0.1437      & 0.0696       & 0.2636                            \\
                                                                           & RandomSelect        & 0.2221       & \textcolor{blue}{0.4619}      & \textcolor{blue}{0.2236}      & 0.2397       & 0.2696       & 0.5674      & 0.2742      & \textcolor{blue}{0.3320}       & \textcolor{blue}{0.1185}       & \textcolor{blue}{0.2382}      & \textcolor{blue}{0.1435}      & \textcolor{blue}{0.0695}       & \textcolor{blue}{0.2634}                            \\
                                                                           & Resample            & \textcolor{red}{0.1619}       & \textcolor{red}{0.2858}      & \textcolor{red}{0.1504}      & \textcolor{red}{0.0885}       & \textcolor{red}{0.2022}       & \textcolor{red}{0.4088}      & \textcolor{red}{0.2201}      & \textcolor{red}{0.2041}       & \textcolor{red}{0.1130}       & \textcolor{red}{0.2176}      & \textcolor{red}{0.1246}      & \textcolor{red}{0.0569}       & \textcolor{red}{0.1862}                            \\ \bottomrule
\end{tabular}

}

\label{tab:population-level-in-domain}
\end{table*}

\begin{table*}[h]
\caption{Out-of-domain alignment results on CFCS, FBPS, and Duckworth datasets across four metrics. Lower is better for all metrics. Best results are \textcolor{red}{red}, second-best \textcolor{blue}{blue}.}
\resizebox{\linewidth}{!}{
\begin{tabular}{@{}l|cccc|cccc|cccc|r@{}}
\toprule
 & \multicolumn{4}{c|}{\textbf{CFCS}} & \multicolumn{4}{c|}{\textbf{FBPS}} & \multicolumn{4}{c|}{\textbf{Duckworth}} & \multicolumn{1}{l}{\textbf{Avg.}} \\
\midrule
 & \textbf{AMW} & \textbf{FD} & \textbf{SW} & \textbf{MMD} & \textbf{AMW} & \textbf{FD} & \textbf{SW} & \textbf{MMD} & \textbf{AMW} & \textbf{FD} & \textbf{SW} & \textbf{MMD} & \multicolumn{1}{c}{} \\
\midrule
AlignX & 0.2189 & 0.0697 & 0.2457 & 0.1091 & 0.3527 & 1.0808 & 0.3122 & 0.5710 & 0.4083 & 0.4333 & 1.2429 & 0.7111 & 0.4796 \\
Tulu-3-Persona & 0.2755 & 0.0826 & 0.3086 & 0.1450 & 0.2418 & 0.4849 & 0.2716 & 0.2564 & 0.3838 & 0.4009 & 0.7533 & 0.5815 & 0.3488 \\
Bavard & 0.2415 & 0.0870 & 0.2708 & 0.1171 & 0.2773 & 0.6788 & 0.2545 & 0.3717 & 0.3048 & 0.3052 & 0.7038 & 0.4301 & 0.3369 \\
Google Synthetic & 0.2563 & 0.0972 & 0.2873 & 0.1320 & 0.2857 & 0.7112 & 0.2648 & 0.3985 & 0.3141 & 0.3122 & 0.7295 & 0.4524 & 0.3534 \\
Nvidia Nemotron & 0.2514 & 0.1029 & 0.2794 & 0.1178 & 0.2595 & 0.9599 & 0.2817 & 0.5387 & 0.2970 & 0.4054 & 1.1136 & 0.6837 & 0.4409 \\
SyncP (GPT-4o) & 0.3258 & 0.1591 & 0.3546 & 0.2043 & 0.3066 & 0.9925 & 0.3054 & 0.5456 & 0.3287 & 0.3573 & 0.8842 & 0.5241 & 0.4407 \\
PersonalHub & 0.2099 & 0.0975 & 0.2358 & 0.0885 & 0.2453 & 0.6068 & 0.2286 & 0.3075 & 0.3236 & 0.3664 & 0.9190 & 0.5443 & 0.3478 \\
\midrule
Ours (Raw) & 0.1977 & 0.0699 & 0.1837 & \textcolor{blue}{0.0701} & 0.2379 & 0.4642 & 0.2018 & 0.2473 & 0.2515 & 0.2331 & \textcolor{blue}{0.4371} & \textcolor{blue}{0.2753} & 0.2391 \\
Ours (RandomSelect) & \textcolor{blue}{0.1973} & \textcolor{blue}{0.0695} & \textcolor{blue}{0.1814} & 0.0706 & \textcolor{blue}{0.2371} & \textcolor{blue}{0.4624} & \textcolor{blue}{0.2013} & \textcolor{blue}{0.2466} & \textcolor{blue}{0.2508} & \textcolor{blue}{0.2325} & 0.4379 & 0.2761 & \textcolor{blue}{0.2386} \\
Ours (Resample) & \textcolor{red}{0.1954} & \textcolor{red}{0.0614} & \textcolor{red}{0.1726} & \textcolor{red}{0.0642} & \textcolor{red}{0.2223} & \textcolor{red}{0.4220} & \textcolor{red}{0.1900} & \textcolor{red}{0.2186} & \textcolor{red}{0.2450} & \textcolor{red}{0.2324} & \textcolor{red}{0.4257} & \textcolor{red}{0.2623} & \textcolor{red}{0.2260} \\
\bottomrule
\end{tabular}
}
\label{tab:population-level-out-domain}
\end{table*}

\paragraph{In-domain results.}
Table~\ref{tab:population-level-in-domain} reports in-domain alignment results across three open-source LLMs: Qwen2.5-72B, Llama-3-70B, and Phi-4. The task evaluates how closely model outputs match human Big Five trait distributions from the IPIP inventory, which also serves as the reference distribution in our alignment pipeline.
In the \textit{Without Persona} setting, increasing the sampling temperature from 0 to 1 improves alignment across all models; for instance, FD on Qwen2.5-72B drops from 1.43 to 1.09, indicating that decoding randomness introduces more human-like variation. However, even at optimal temperature, these settings perform far worse than persona-conditioned baselines, underscoring the limitations of unconstrained generation.
All \textit{With Persona} baselines yield substantial improvements. Public persona sets (e.g., Tulu-3-Persona, AlignX, PersonalHub) moderately reduce alignment error, while SyncP variants (e.g., GPT-4o-generated personas) perform comparably but do not consistently surpass hand-crafted sets. These gains confirm the value of personas, but also reveal persistent mismatches with population-level ground truth, highlighting the limits of naive persona generation.
Our framework achieves the best results across all models. The \textit{Resample} variant, built with importance-based selection and optimal transport, performs best on all four metrics for each base model. Averaging AMW, FD, SW, and MMD within each model, \textit{Resample} reduces alignment error by 49.8\% on Qwen2.5-72B (vs. SyncP (GPT-4o)), 37.9\% on Llama-3-70B (vs. Google Synthetic), and 45.1\% on Phi-4 (vs. AlignX).
Even our unaligned variants ({Raw} and {RandomSelect}) outperform most public persona sets, demonstrating the strength of our seed pool. Their near-identical performance also confirms that the gains from {Resample} stem from distributional alignment rather than sample size effects.

\paragraph{Out-of-domain results.}

\begin{wraptable}[14]{r}{0.49\columnwidth}
    \vspace{-6pt}
    \centering
    \caption{
    {MAE\textsubscript{corr} between human and synthetic inter-trait correlations.} 
    Lower values indicate better individual-level consistency. 
    Best results are \textcolor{red}{red}, second-best \textcolor{blue}{blue}.
}
    \label{tab:consistency}
    \vspace{-4pt}
    \resizebox{\linewidth}{!}{%
    \begin{tabular}{@{}lccccccccccccccccc@{}}
    \toprule
    \textbf{Method}     & {\color[HTML]{333333} \textbf{BIG5}} & {\color[HTML]{333333} \textbf{CFCS}} & {\color[HTML]{333333} \textbf{FBPS}} & \textbf{Duwo} & \multicolumn{1}{c}{\textbf{Avg.}} \\ \midrule
    AlignX              & 0.5761                               & 0.7944                               & 0.5078                               & 0.5688        & {\color[HTML]{333333} 0.6118}     \\
    Tulu-3-Persona      & 0.4971                               & 1.0521                               & 0.4017                               & 0.4970        & {\color[HTML]{333333} 0.6120}     \\
    Bavard              & 0.4528                               & 0.4066                               & 0.4268                               & 0.4559        & {\color[HTML]{333333} 0.4355}     \\
    Google Synthetic    & 0.4570                               & 0.4382                               & 0.4168                               & 0.4571        & {\color[HTML]{333333} 0.4423}     \\
    Nvidia Nemotron     & 0.8000                               & 1.1935                               & 0.6894                               & 0.8075        & {\color[HTML]{333333} 0.8726}     \\
    SyncP (GPT-4o)     & 1.0412                               & 1.2794                               & 0.9289                               & 1.0443        & {\color[HTML]{333333} 1.0735}     \\
    PersonalHub         & 0.6547                               & 0.9334                               & 0.5045                               & 0.6564        & {\color[HTML]{333333} 0.6873}     \\ \midrule
    Ours (Raw)          & \textcolor{blue}{0.4357}                               & \textcolor{blue}{0.2447}                               & \textcolor{blue}{0.3874}                               & \textcolor{blue}{0.4467}        & {\color[HTML]{333333} \textcolor{blue}{0.3786}}     \\
    Ours (RandomSelect) & 0.4366                               & 0.2473                               & 0.3895                               & 0.4483        & {\color[HTML]{333333} 0.3804}     \\
    Ours (Resample)     & \textcolor{red}{0.4243}                               & \textcolor{red}{0.2300}                               & \textcolor{red}{0.3684}                               & \textcolor{red}{0.4377}        & {\color[HTML]{333333} \textcolor{red}{0.3651}}     \\     \bottomrule
    \end{tabular}
    }
    \vspace{-8pt}
\end{wraptable}

Table~\ref{tab:population-level-out-domain} presents the population-level alignment results on three psychological instruments, CFCS~\citep{Strathman1994CFC}, FBPS~\citep{OpenPsychometrics2019Firstborn}, and Duwo i.e., Duckworth~\citep{Duckworth2007Grit}, that were not used during persona construction. Despite measuring distinct constructs, these datasets share an important property: they are all built on globally distributed human responses, covering 144, 169, and 111 countries respectively. This aligns well with the IPIP Big Five inventory used in our alignment process, which spans one million individuals across 223 countries and serves as a proxy for global human trait distribution.
Across all three tasks and four evaluation metrics (AMW, FD, SW, MMD), {Resample} achieves the lowest average divergence (0.2260), outperforming the strongest baseline (Bavard, 0.3369) by 33\%. On FBPS our method reduces the Fréchet Distance to 0.4220, substantially outperforming Tulu-3-Persona (0.4849), Google Synthetic (0.7112), and SyncP (GPT-4o) (0.9925). Even our unaligned variants, {Raw} (0.2391) and {RandomSelect} (0.2386), consistently outperform all baselines, highlighting the strength of the underlying seed pool. The additional gains from the alignment process validate the necessity of principled population-level sampling in constructing accurate and transferable persona sets.

\begin{table}[h]
\caption{Performance of group-specific persona generation across datasets and regions.
{ED denotes Energy Distance. Lower is better for all metrics.} Best results are \textcolor{red}{red}, second-best \textcolor{blue}{blue}.
} 
\scalebox{0.51
}{
\begin{tabular}{@{}lccccccccccccccccc@{}}
\toprule
                                         & \multicolumn{4}{c}{\textbf{YRBSS}}                                                                                                                 & \multicolumn{4}{c}{\textbf{WVS (East Asia)}}                                                                                                        & \multicolumn{4}{c}{\textbf{WVS (North America)}}                                                                                                    & \multicolumn{4}{c}{\textbf{WVS (Western Europe)}}                                                                                                   & \textbf{Avg.}                 \\ \midrule
\multicolumn{1}{l|}{}                    & \textbf{AMW}                  & \textbf{FD}                   & \textbf{SW}                   & \multicolumn{1}{c|}{\textbf{ED}}                  & \textbf{AMW}                  & \textbf{FD}                    & \textbf{SW}                   & \multicolumn{1}{c|}{\textbf{ED}}                  & \textbf{AMW}                  & \textbf{FD}                    & \textbf{SW}                   & \multicolumn{1}{c|}{\textbf{ED}}                  & \textbf{AMW}                  & \textbf{FD}                    & \textbf{SW}                   & \multicolumn{1}{c|}{\textbf{ED}}                  & \multicolumn{1}{l}{}          \\ \midrule
\multicolumn{1}{l|}{AlignX}              & 0.4676                        & 9.7132                        & 0.5679                        & \multicolumn{1}{c|}{2.0409}                        & 0.8274                        & 21.0512                        & 0.8712                        & \multicolumn{1}{c|}{2.4598}                        & 0.8785                        & 20.4461                        & 0.8914                        & \multicolumn{1}{c|}{2.4495}                        & 0.8595                        & 23.4058                        & 0.8576                        & \multicolumn{1}{c|}{2.0302}                        & 5.6136                        \\
\multicolumn{1}{l|}{Tulu-3-Persona}      & 0.4921                        & 9.0356                        & 0.7252                        & \multicolumn{1}{c|}{1.9259}                        & 0.9282                        & 23.2265                        & 0.9343                        & \multicolumn{1}{c|}{2.6507}                        & 0.8498                        & 20.8217                        & 0.8976                        & \multicolumn{1}{c|}{2.6438}                        & 0.9485                        & 26.8801                        & 0.9615                        & \multicolumn{1}{c|}{2.3601}                        & 6.0176                        \\
\multicolumn{1}{l|}{Bavard}              & 0.4569                        & 9.0103                        & 0.5558                        & \multicolumn{1}{c|}{1.8776}                        & 0.8153                        & 22.7143                        & 0.8190                        & \multicolumn{1}{c|}{2.1368}                        & 0.8489                        & 20.3607                        & 0.8730                        & \multicolumn{1}{c|}{2.4311}                        & 0.8991                        & 24.8798                        & 0.8593                        & \multicolumn{1}{c|}{1.9787}                        & 5.7198                        \\
\multicolumn{1}{l|}{Google Synthetic}    & 0.4716                        & 9.0975                        & 0.5500                        & \multicolumn{1}{c|}{1.8997}                        & 0.8197                        & 21.5079                        & 0.8443                        & \multicolumn{1}{c|}{2.2662}                        & 0.8412                        & 22.6565                        & 0.8757                        & \multicolumn{1}{c|}{2.6811}                        & 0.8818                        & 24.8154                        & 0.8667                        & \multicolumn{1}{c|}{1.9872}                        & 5.8164                        \\
\multicolumn{1}{l|}{Nvidia Nemotron}     & 0.5052                        & 10.9920                       & 0.6999                        & \multicolumn{1}{c|}{2.2383}                        & 0.8845                        & 24.7850                        & 0.8990                        & \multicolumn{1}{c|}{2.6935}                        & 0.8687                        & 20.2640                        & 0.8733                        & \multicolumn{1}{c|}{2.4560}                        & 0.8793                        & 26.9882                        & 0.8832                        & \multicolumn{1}{c|}{2.1464}                        & 6.1910                        \\
\multicolumn{1}{l|}{SyncP (GPT-4o)}     & 0.5473                        & 11.5000                       & 0.7013                        & \multicolumn{1}{c|}{2.4586}                        & 0.9706                        & 31.6886                        & 0.9890                        & \multicolumn{1}{c|}{3.4136}                        & 0.8183                        & 22.5895                        & 0.8881                        & \multicolumn{1}{c|}{2.4783}                        & 1.0129                        & 31.3838                        & 1.0204                        & \multicolumn{1}{c|}{3.0113}                        & 7.2170                        \\
\multicolumn{1}{l|}{PersonalHub}         & 0.4834                        & 9.9214                        & 0.6021                        & \multicolumn{1}{c|}{2.1532}                        & 0.8144                        & 22.8832                        & 0.8262                        & \multicolumn{1}{c|}{2.2022}                        & 0.8204                        & 21.1220                        & 0.8551                        & \multicolumn{1}{c|}{2.7758}                        & 0.8678                        & 24.9372                        & 0.8577                        & \multicolumn{1}{c|}{1.9928}                        & 5.8822                        \\ \midrule
\multicolumn{1}{l|}{Ours (Raw)}          & 0.4421                        & 8.9240                        & 0.5210                        & \multicolumn{1}{c|}{1.7346}                        & {\color[HTML]{3531FF} 0.8003} & {\color[HTML]{3531FF} 20.6011} & {\color[HTML]{3531FF} 0.8078} & \multicolumn{1}{c|}{{\color[HTML]{3531FF} 2.0013}} & 0.8150                        & {\color[HTML]{3531FF} 19.6053} & 0.8465                        & \multicolumn{1}{c|}{2.3168}                        & 0.8458                        & { 22.6507} & 0.8410                        & \multicolumn{1}{c|}{1.9251}                        & 5.3549                        \\
\multicolumn{1}{l|}{Ours (EM w/o train)} & {\color[HTML]{3531FF} 0.2968} & {\color[HTML]{3531FF} 2.5465} & {\color[HTML]{3531FF} 0.3371} & \multicolumn{1}{c|}{{\color[HTML]{3531FF} 0.7598}} & 0.8072                        & 20.6860                        & 0.8166                        & \multicolumn{1}{c|}{2.1309}                        & {\color[HTML]{3531FF} 0.8091} & 20.1998                        & {\color[HTML]{3531FF} 0.8446} & \multicolumn{1}{c|}{{\color[HTML]{3531FF} 2.2947}} & {\color[HTML]{3531FF} 0.8352} &  \color[HTML]{3531FF} 22.0169                        & {\color[HTML]{3531FF} 0.8376} & \multicolumn{1}{c|}{{\color[HTML]{3531FF} 1.8256}} & {\color[HTML]{3531FF} 4.8778} \\
\multicolumn{1}{l|}{Ours (EM w/ train)}   & {\color[HTML]{FE0000} 0.2903} & {\color[HTML]{FE0000} 2.4841} & {\color[HTML]{FE0000} 0.3205} & \multicolumn{1}{c|}{{\color[HTML]{FE0000} 0.7470}} & {\color[HTML]{FE0000} 0.7840} & {\color[HTML]{FE0000} 19.4737} & {\color[HTML]{FE0000} 0.7899} & \multicolumn{1}{c|}{{\color[HTML]{FE0000} 1.9562}} & {\color[HTML]{FE0000} 0.7737} & {\color[HTML]{FE0000} 17.5798} & {\color[HTML]{FE0000} 0.7876} & \multicolumn{1}{c|}{{\color[HTML]{FE0000} 1.8028}} & {\color[HTML]{FE0000} 0.7897} & {\color[HTML]{FE0000} 21.5078} & {\color[HTML]{FE0000} 0.8111} & \multicolumn{1}{c|}{{\color[HTML]{FE0000} 1.7363}} & {\color[HTML]{FE0000} 4.5397} \\ \bottomrule
\end{tabular}
}
\label{tab:persona_specific}
\end{table}

\subsection{Individual-Level Behavioral Consistency}
\label{subsec:individual_consistency}
\vspace{-2mm}

Beyond matching trait distributions, realistic simulation must preserve the internal structure of personality traits \cite{wang2024not}. We evaluate this using mean absolute error (MAE\textsubscript{corr}) between pairwise Pearson correlations in human vs.\ synthetic data (see Section~\ref{sec:Exp_setup}). Table~\ref{tab:consistency} reports results across four instruments: IPIP Big Five~\citep{ipip50}, CFCS~\citep{Strathman1994CFC}, FBPS~\citep{OpenPsychometrics2019Firstborn}, and Duckworth~\citep{Duckworth2007Grit}.
Our proposed {Resample} method consistently achieves the strongest performance, producing personas whose inter-trait correlations most closely mirror those of real human populations. It attains the lowest error on \emph{every} instrument, with an overall MAE\textsubscript{corr} of 0.3651, outperforming the strongest prior baseline (Bavard, 0.4355) by 16.2\%, and  surpassing commonly used sets like AlignX (0.6118) and Tulu-3-Persona (0.6120). These results demonstrate that our alignment pipeline sharpens individual-level consistency, not just population-level fit.
Interestingly, we observe that the full two-stage alignment procedure provides clear improvements even over our own high-quality seed pool. Compared to the unaligned variant ({Raw}), the final {Resample} variant reduces MAE\textsubscript{corr} by 3.9\% on average, confirming that distribution-aware sampling can enhance internal coherence even when the source data is already strong. This is further supported by the minimal difference between {Raw} and {RandomSelect} (0.3786 vs.\ 0.3804), which isolates the effect of sample size control.

\subsection{Group-Specific Persona Adaptation}
\label{subsec:group_specific}

\vspace{-2mm}

\begin{table*}[t]
\caption{{Examples of seed personas and their LLM-generated outputs across different demographic adaptation tasks.} Each output persona reflects the cultural, behavioral, and social context of the respective group.}
\vspace{-2mm}
\centering
\scriptsize
\renewcommand{\arraystretch}{1.05}
\setlength{\tabcolsep}{3.5pt}
\begin{adjustbox}{max width=\textwidth}
\begin{tabularx}{\textwidth}{@{}l YY@{}}
\toprule
\textbf{Setting} & \textbf{Seed Persona} & \textbf{Output Persona} \\
\midrule
YRBSS (U.S. teens) &
The user is a young adult, likely female, living in Penticton, Canada, who values a healthy lifestyle and strives to maintain an active routine, often starting her day with yoga, walks, or other exercises, despite facing challenges with energy and motivation, particularly due to her early work schedule and occasional injuries. &
Alex, 16, lives in a suburban area of California with her parents and younger brother. She is a junior at a public high school, where she excels in academics and is on the track team. Alex starts her day with a 30-minute jog and eats a balanced diet. She has had minor injuries but stays committed to fitness. 
She manages stress through journaling and talking to friends. \\
\midrule
WVS East (Hong Kong) &
The user is a young, likely Asian, possibly Chinese, individual who values cultural heritage and identity, is emotional and sentimental, and is openly gay and appreciative of diversity. &
Alex Wong, 28, lives in Hong Kong and works as a graphic designer. He embraces Chinese heritage in his work, is openly gay, and actively supports LGBTQ+ causes. He visits ancestral sites for emotional grounding, participates in community service, and supports democratic reform. He balances sentimentality with a pragmatic mindset. \\
\midrule
WVS Europe (Germany) &
The user is a 25-year-old married male, living with his wife, earning \$145,000–\$160,000 jointly, focused on saving for a house, and values work-life balance. &
Markus, 25, lives in Munich with his wife Lisa. Both work in tech and earn €130,000–€145,000 annually. Markus prioritizes flexible hours and vacation time. They’re saving for a home and planning for children. He supports welfare policies and environmental volunteering, favoring collective well-being over individualism. \\
\midrule
WVS North America (U.S.) &
The user is a likely young, liberal-minded individual, possibly American, who values critical thinking and challenges norms, with strong views on justice and social responsibility. &
Emily Thompson, 32, lives in suburban Chicago. A marketing professional and moderate Democrat, she volunteers locally, values civic duty, and supports social justice. She votes regularly, discusses reform topics, and values honesty and responsibility. She is optimistic despite student debt and owns two rescue dogs. \\
\bottomrule
\end{tabularx}
\end{adjustbox}
\vspace{-2mm}
\label{tab:seed_output_cases}
\end{table*}

\begin{wrapfigure}[23]{l}{0.47\columnwidth}
    \vspace{-6pt}
    \centering
    \includegraphics[width=\linewidth]{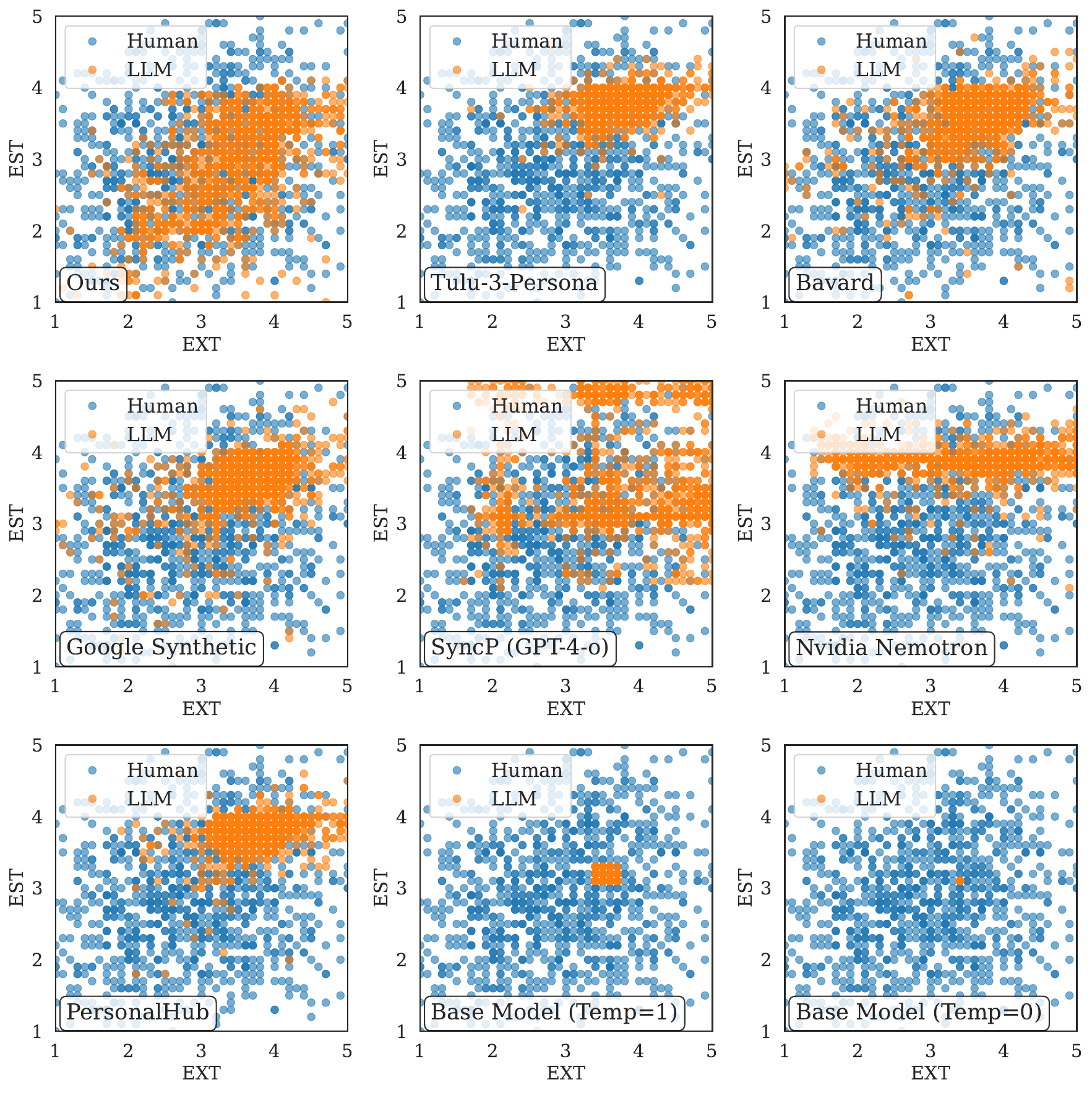}
    \vspace{-6pt}
    \caption{
    Scatter plots comparing the distribution of  EXT and  EST  scores between real human responses (blue) and LLM-generated outputs (orange) across nine settings. 
    }
    \label{fig:scatter_ext_est_image}
    \vspace{-8pt}
\end{wrapfigure}

We assess whether the globally aligned pool $\mathcal{P}'$ of 5,000 personas can be adapted to specific demographic targets using two test sets: \textit{YRBSS}~\citep{Brener2024OverviewAM} and \textit{WVS}~\citep{Inglehart2020WVS}. For each group, we issue a natural-language query (e.g., ``Generate a persona from the Hong Kong region’’) and retrieve the top-$K$ personas from $\mathcal{P}'$ using an embedding model. Specifically, we use {Qwen3-Embedding-0.6B}~\citep{qwen3embedding}, which maps both queries and personas into a shared semantic space. 
The corresponding training hyper-parameters and prompt details for the embedding model can be found in Appendix~\ref{app:query},  ~\ref{app:negfilter} and ~\ref{app:group_prompts}.
Retrieved personas are then revised using Qwen2.5‑72B‑Instruct to produce 1,000 group-specific personas per setting.
We compare: (1) {Raw}; (2) {EM w/o train}, retrieval from {Resample} with an \emph{untrained} encoder; and (3) {EM w/ train}, our contrastively trained encoder. 
Alignment is evaluated using AMW, FD, SW, and Energy Distance (ED), averaged by region.
Table~\ref{tab:persona_specific} shows {EM w/ train} achieves the lowest errors across all regions/metrics, reducing average alignment error by 19.1\% vs.\ AlignX and 6.9\% vs.\ {EM w/o train}. 
Even {EM w/o train} outperforms all external sets, validating the effectiveness of the retrieval approach, and  gains from {EM w/ train} over {EM w/o train} are consistent, indicating contrastive training improves generalization rather than overfitting.
To complement the quantitative results, we further present  four case studies of group-specific persona construction, each comprising a raw seed persona and its LLM-generated output persona after adaptation to the target setting. These outputs follow the prompts detailed in Table~\ref{tab:seed_output_cases}.

\subsection{Trait-Space Visualization}
\label{subsec:trait_vis}

\vspace{-2mm}

To complement our quantitative analysis, we provide a qualitative visualization of personality trait coverage. Following the same in-domain setup as in Section~\ref{sec:population-level}, we plot the joint distribution over two representative Big Five traits: {Extraversion (EXT)} and {Emotional Stability (EST)}. Figure~\ref{fig:scatter_ext_est_image} shows scatter plots of 1,000 samples per setting across nine representative conditions, where the $x$-axis denotes EXT and the $y$-axis denotes EST. Blue points represent real human responses; orange points denote LLM outputs. Each subplot is labeled with the corresponding method.
In the {Temp = 0} setting (bottom right), all responses collapse to a single point due to deterministic decoding. Increasing temperature to 1 adds some dispersion, but the outputs remain narrowly clustered, failing to reflect real-world variability. This shows that sampling alone is insufficient for simulating population-level diversity.
Persona-based approaches (e.g., Tulu-3-Persona, Bavard, Google Synthetic) offer modest improvements, with outputs spreading further into trait space and partially overlapping human data. However, coverage remains limited-especially in the lower-EXT and lower-EST regions-indicating insufficient representation of diverse psychological profiles.
By contrast, our {Ours (Resample)} method (top left) yields outputs that closely match the human distribution, covering the full trait space and successfully reaching underrepresented regions. This strongly supports our quantitative findings in Section~\ref{sec:population-level}, affirming the effectiveness of our two-stage sampling approach.
Although our method achieves substantial improvements in trait alignment, the generated distribution still falls short of fully matching real human data, due to inherent model and data biases. Closing this gap remains a key challenge.

\begin{wrapfigure}[20]{r}{0.47\columnwidth}
    \vspace{-6pt}
    \centering
    \includegraphics[width=1.0\linewidth]{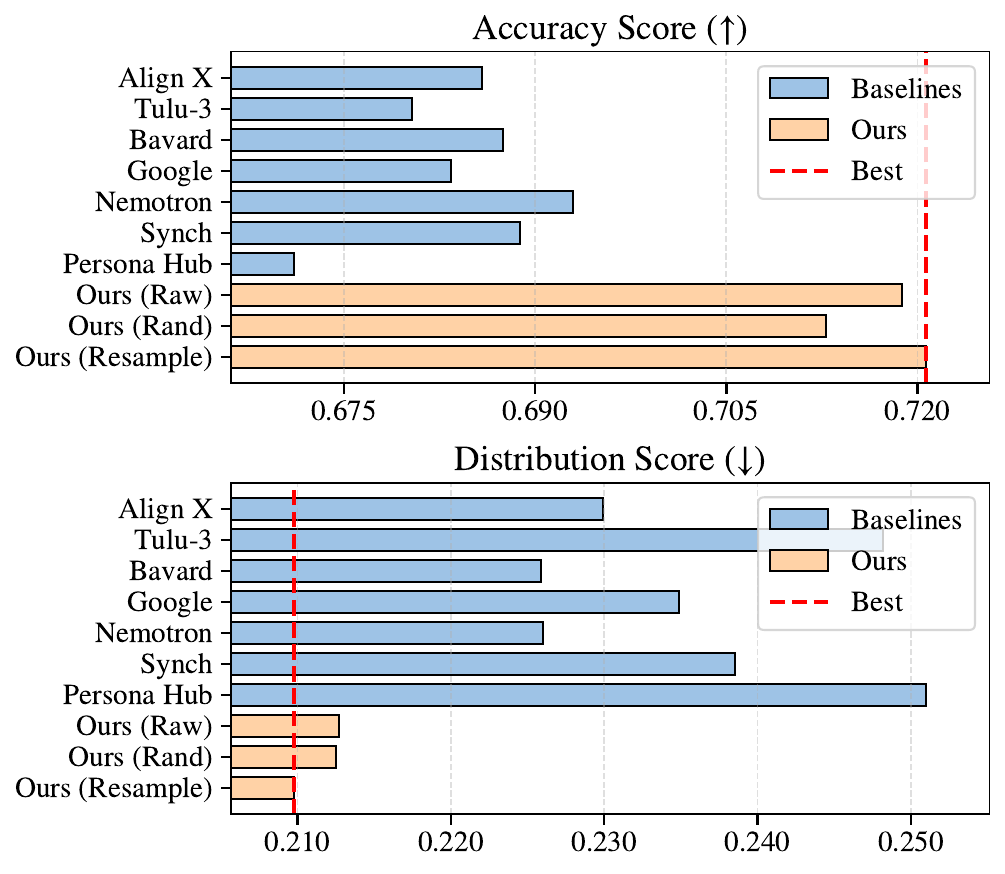}
    \vspace{-6pt}
    \caption{Result for social science experiments. Higher is better for Accuracy Score; lower is better for Distribution Score.}
    \label{fig:socsci_metrics}
    \vspace{-8pt}
\end{wrapfigure}

\subsection{Simulation of Social Science Experiments}
\label{sec:SocSci}
\vspace{-2mm}
We further evaluate persona sets on {SocSci}~\citep{Kolluri2025FinetuningLF}, a benchmark derived from open-source social science experiments, where models are prompted to answer outcome questions conditioned on a persona and an experimental stimulus.
Appendix~\ref{app:behavior_generalization} reports six behavioral games and multi-turn customer simulation.
As shown in Figure~\ref{fig:socsci_metrics}, among existing persona sets, {Nvidia Nemotron} achieves the strongest Accuracy Score (0.6930), while {Bavard} yields the best Distribution Score (0.2259).
In contrast, our personas substantially improve both response-level and condition-level fidelity: {Ours (Raw)} and {Ours (RandomSelect)} already outperform all baselines on both metrics.
Finally, {Ours (Resample)} provides the best overall alignment, reaching the lowest Distribution Score (0.2098) while also attaining the highest Accuracy Score (0.7207).
These gains indicate that population-aligned persona construction transfers beyond psychometric inventories to downstream social science simulations, improving both individual response agreement and condition-level distribution matching.

\section{Ablation Study of Hyperparameters}
\label{app:hyper-ablation}

We conduct a sensitivity analysis for the key hyperparameters in our two-stage global alignment module (Section~\ref{subsec:IS_OT}). 
Unless otherwise specified, we vary one parameter at a time while keeping the others at their default values: KDE bandwidth $h{=}0.20$, IS Top-$k{=}70\%$, OT entropic regularization $\varepsilon{=}0.08$, Sinkhorn iterations $T{=}250$, and target persona size $N'{=}5\mathrm{k}$. 
We report the population-level distributional discrepancy measured by AMW (lower is better) on IPIP Big Five, using Qwen2.5-72B as the base model. 
For reference, we also include a size-matched random subset of our seed pool ({RandomSelect}) and representative existing persona sets (Nvidia Nemotron, AlignX, Google Synthetic), holding these baselines fixed when varying KDE, IS, and OT parameters.

\paragraph{Findings.}
Table~\ref{tab:sensitivity-amw} shows that our method is stable across a broad range of KDE bandwidths and IS Top-$k$ thresholds: AMW varies only slightly (typically within $[0.16, 0.17]$). 
Similarly, OT is robust to $\varepsilon$ and Sinkhorn iterations, suggesting that our entropic OT solver is not overly sensitive to these optimization choices in the studied regime.
Finally, the target persona population size $N'$ has the most visible effect: larger persona sets improve alignment for all methods, and our approach benefits the most, reaching the best AMW at $N'{=}10\mathrm{k}$ while maintaining strong performance at the default $N'{=}5\mathrm{k}$ for computational efficiency.

\begin{table}[t]
\caption{Sensitivity analysis of key hyperparameters in the two-stage sampler on IPIP Big Five (AMW; lower is better) with Qwen2.5-72B.
We vary one parameter at a time and keep others at defaults ($h{=}0.20$, Top-$k{=}70\%$, $\varepsilon{=}0.08$, Sinkhorn iters $T{=}250$, persona size $N'{=}5\mathrm{k}$).
Numbers are reported as ({RandomSelect} / {Ours} / {Nvidia} / {AlignX} / {Google}).}
\centering
\scriptsize
\setlength{\tabcolsep}{4.2pt}

\resizebox{0.8\columnwidth}{!}{
\begin{tabular*}{\columnwidth}{@{\extracolsep{\fill}} lccccc}
\toprule
\textbf{Setting} & \textbf{RandomSelect} & \textbf{Ours} & \textbf{Nvidia} & \textbf{AlignX} & \textbf{Google} \\
\midrule

\multicolumn{6}{@{}l}{\textbf{KDE bandwidth $h$ (default: $0.20$)}} \\
\midrule
$h=0.10$ & 0.22 & 0.17 & 0.26 & 0.34 & 0.31 \\
$\mathbf{h=0.20}$ & 0.22 & 0.16 & 0.26 & 0.34 & 0.31 \\
$h=0.30$ & 0.22 & 0.16 & 0.26 & 0.34 & 0.31 \\
$h=0.50$ & 0.22 & 0.17 & 0.26 & 0.34 & 0.31 \\
\midrule

\multicolumn{6}{@{}l}{\textbf{IS Top-$k$ (default: $70\%$)}} \\
\midrule
Top-$k=55\%$ & 0.22 & 0.17 & 0.26 & 0.34 & 0.31 \\
Top-$k=70\%$ & 0.22 & 0.16 & 0.26 & 0.34 & 0.31 \\
Top-$k=85\%$ & 0.22 & 0.16 & 0.26 & 0.34 & 0.31 \\
Top-$k=95\%$ & 0.22 & 0.17 & 0.26 & 0.34 & 0.31 \\
\midrule

\multicolumn{6}{@{}l}{\textbf{OT entropic regularization $\varepsilon$ (default: $0.08$)}} \\
\midrule
$\varepsilon=0.05$ & 0.22 & 0.16 & 0.26 & 0.34 & 0.31 \\
$\varepsilon=0.08$ & 0.22 & 0.16 & 0.26 & 0.34 & 0.31 \\
$\varepsilon=0.12$ & 0.22 & 0.16 & 0.26 & 0.34 & 0.31 \\
$\varepsilon=0.20$ & 0.22 & 0.17 & 0.26 & 0.34 & 0.31 \\
\midrule

\multicolumn{6}{@{}l}{\textbf{Sinkhorn iterations $T$ (default: $250$)}} \\
\midrule
$T=50$  & 0.22 & 0.17 & 0.26 & 0.34 & 0.31 \\
$T=100$ & 0.22 & 0.17 & 0.26 & 0.34 & 0.31 \\
$T=\mathbf{250}$ & 0.22 & 0.16 & 0.26 & 0.34 & 0.31 \\
$T=500$ & 0.22 & 0.16 & 0.26 & 0.34 & 0.31 \\
\midrule

\multicolumn{6}{@{}l}{\textbf{Target persona size $N'$ (default: $5\mathrm{k}$)}} \\
\midrule
$N'=1\mathrm{k}$   & 0.24 & 0.19 & 0.28 & 0.35 & 0.32 \\
$N'=2.5\mathrm{k}$ & 0.23 & 0.17 & 0.27 & 0.34 & 0.32 \\
$N'=\mathbf{5\mathrm{k}}$   & 0.22 & 0.16 & 0.26 & 0.34 & 0.31 \\
$N'=10\mathrm{k}$  & 0.22 & 0.15 & 0.26 & 0.34 & 0.31 \\
\bottomrule
\end{tabular*}
}

\vspace{-2mm}
\label{tab:sensitivity-amw}
\end{table}

\section{Conclusion}
\vspace{-2mm}
\label{sec:conclusion}
We propose a principled framework for constructing population-aligned persona sets for LLM-based social simulation. Existing approaches often focus on individual-level attributes while neglecting population-level distributional fidelity, leading to biased and unrealistic outcomes. 
Our three-stage pipeline (1) extracts high-quality personas from real-world social media, (2) performs two-stage alignment via importance-based selection and optimal transport, and (3) enables group-specific persona generation.
Across seven diverse evaluation tasks, our method consistently outperforms widely used persona sets in both population alignment and behavioral realism. It generalizes well across questionnaire, survey, and social-science simulation settings and supports high-fidelity group-specific simulation.
These results highlight the importance of distribution-aware persona construction for reliable social modeling.
Future work will explore higher-order simulations for forecasting social trends and evaluating intervention strategies.

\bibliographystyle{plain}
\bibliography{reference}

@article{Huang2024UnderstandingTP,
  title={Understanding the planning of LLM agents: A survey},
  author={Xu Huang and Weiwen Liu and Xiaolong Chen and Xingmei Wang and Hao Wang and Defu Lian and Yasheng Wang and Ruiming Tang and Enhong Chen},
  journal={ArXiv},
  year={2024},
  volume={abs/2402.02716},
  url={https://api.semanticscholar.org/CorpusID:267411892}
}

@article{Fulay2025TheEC,
  title={The Empty Chair: Using LLMs to Raise Missing Perspectives in Policy Deliberations},
  author={Suyash Pradeep Fulay and Deb Roy},
  journal={ArXiv},
  year={2025},
  volume={abs/2503.13812},
  url={https://api.semanticscholar.org/CorpusID:277103981}
}

@inproceedings{Song2024PredictingUB,
  title={Predicting User Behavior in Smart Spaces with LLM-Enhanced Logs and Personalized Prompts},
  author={Yunpeng Song and Jiawei Li and Yiheng Bian and Zhongmin Cai},
  booktitle={AAAI Conference on Artificial Intelligence},
  year={2024},
  url={https://api.semanticscholar.org/CorpusID:274789599}
}

@article{Karvonen2025RobustlyIL,
  title={Robustly Improving LLM Fairness in Realistic Settings via Interpretability},
  author={Adam Karvonen and Samuel Marks},
  journal={ArXiv},
  year={2025},
  volume={abs/2506.10922},
  url={https://api.semanticscholar.org/CorpusID:279318319}
}

@article{Park2023GenerativeAI,
  title={Generative Agents: Interactive Simulacra of Human Behavior},
  author={Joon Sung Park and Joseph C. O’Brien and Carrie J. Cai and Meredith Ringel Morris and Percy Liang and Michael S. Bernstein},
  journal={Proceedings of the 36th Annual ACM Symposium on User Interface Software and Technology},
  year={2023},
  url={https://api.semanticscholar.org/CorpusID:258040990}
}

@inproceedings{Chuang2023SimulatingOD,
  title={Simulating Opinion Dynamics with Networks of LLM-based Agents},
  author={Yun-Shiuan Chuang and Agam Goyal and Nikunj Harlalka and Siddharth Suresh and Robert Hawkins and Sijia Yang and Dhavan Shah and Junjie Hu and Timothy T. Rogers},
  booktitle={NAACL-HLT},
  year={2023},
  url={https://api.semanticscholar.org/CorpusID:265220960}
}

@article{Gao2023S3SS,
  title={S3: Social-network Simulation System with Large Language Model-Empowered Agents},
  author={Chen Gao and Xiaochong Lan and Zhi-jie Lu and Jinzhu Mao and Jing Piao and Huandong Wang and Depeng Jin and Yong Li},
  journal={ArXiv},
  year={2023},
  volume={abs/2307.14984},
  url={https://api.semanticscholar.org/CorpusID:260202947}
}

@article{Wang2024SimulatingHD,
  title={Simulating Human-like Daily Activities with Desire-driven Autonomy},
  author={Yiding Wang and Yuxuan Chen and Fangwei Zhong and Long Ma and Yizhou Wang},
  journal={ArXiv},
  year={2024},
  volume={abs/2412.06435},
  url={https://api.semanticscholar.org/CorpusID:274598179}
}

@article{Wang2025LLMpoweredMF,
  title={LLM-powered Multi-agent Framework for Goal-oriented Learning in Intelligent Tutoring System},
  author={Tianfu Wang and Yi Zhan and Jianxun Lian and Zhengyu Hu and Nicholas Jing Yuan and Qi Zhang and Xing Xie and Hui Xiong},
  journal={Companion Proceedings of the ACM on Web Conference 2025},
  year={2025},
  url={https://api.semanticscholar.org/CorpusID:275920775}
}

@article{Shao2023CharacterLLMAT,
  title={Character-LLM: A Trainable Agent for Role-Playing},
  author={Yunfan Shao and Linyang Li and Junqi Dai and Xipeng Qiu},
  journal={ArXiv},
  year={2023},
  volume={abs/2310.10158},
  url={https://api.semanticscholar.org/CorpusID:264145862}
}

@inproceedings{Kong2023BetterZR,
  title={Better Zero-Shot Reasoning with Role-Play Prompting},
  author={Aobo Kong and Shiwan Zhao and Hao Chen and Qicheng Li and Yong Qin and Ruiqi Sun and Xiaoxia Zhou},
  booktitle={North American Chapter of the Association for Computational Linguistics},
  year={2023},
  url={https://api.semanticscholar.org/CorpusID:260900230}
}

@article{Chen2024FromPT,
  title={From Persona to Personalization: A Survey on Role-Playing Language Agents},
  author={Jiangjie Chen and Xintao Wang and Rui Xu and Siyu Yuan and Yikai Zhang and Wei Shi and Jian Xie and Shuang Li and Ruihan Yang and Tinghui Zhu and Aili Chen and Nianqi Li and Lida Chen and Caiyu Hu and Siye Wu and Scott Ren and Ziquan Fu and Yanghua Xiao},
  journal={Trans. Mach. Learn. Res.},
  year={2024},
  volume={2024},
  url={https://api.semanticscholar.org/CorpusID:269448713}
}

@article{Lu2024LargeLM,
  title={Large Language Models are Superpositions of All Characters: Attaining Arbitrary Role-play via Self-Alignment},
  author={Keming Lu and Bowen Yu and Chang Zhou and Jingren Zhou},
  journal={ArXiv},
  year={2024},
  volume={abs/2401.12474},
  url={https://api.semanticscholar.org/CorpusID:267095369}
}

@inproceedings{Tu2024CharacterEvalAC,
  title={CharacterEval: A Chinese Benchmark for Role-Playing Conversational Agent Evaluation},
  author={Quan Tu and Shilong Fan and Zihang Tian and Rui Yan},
  booktitle={Annual Meeting of the Association for Computational Linguistics},
  year={2024},
  url={https://api.semanticscholar.org/CorpusID:266725287}
}

@article{Piao2025EmergenceOH,
  title={Emergence of human-like polarization among large language model agents},
  author={Jing Piao and Zhihong Lu and Chen Gao and Fengli Xu and Fernando P. Santos and Yong Li and James Evans},
  journal={ArXiv},
  year={2025},
  volume={abs/2501.05171},
  url={https://api.semanticscholar.org/CorpusID:275405906}
}

@article{Hong2023MetaGPTMP,
  title={MetaGPT: Meta Programming for Multi-Agent Collaborative Framework},
  author={Sirui Hong and Xiawu Zheng and Jonathan P. Chen and Yuheng Cheng and Ceyao Zhang and Zili Wang and Steven Ka Shing Yau and Zi Hen Lin and Liyang Zhou and Chenyu Ran and Lingfeng Xiao and Chenglin Wu},
  journal={ArXiv},
  year={2023},
  volume={abs/2308.00352},
  url={https://api.semanticscholar.org/CorpusID:260351380}
}

@inproceedings{Wang2023UnleashingTE,
  title={Unleashing the Emergent Cognitive Synergy in Large Language Models: A Task-Solving Agent through Multi-Persona Self-Collaboration},
  author={Zhenhailong Wang and Shaoguang Mao and Wenshan Wu and Tao Ge and Furu Wei and Heng Ji},
  booktitle={North American Chapter of the Association for Computational Linguistics},
  year={2023},
  url={https://api.semanticscholar.org/CorpusID:259765919}
}

@article{Li2024PersonalLA,
  title={Personal LLM Agents: Insights and Survey about the Capability, Efficiency and Security},
  author={Yuanchun Li and Hao Wen and Weijun Wang and Xiangyu Li and Yizhen Yuan and Guohong Liu and Jiacheng Liu and Wenxing Xu and Xiang Wang and Yi Sun and Rui Kong and Yile Wang and Hanfei Geng and Jian Luan and Xuefeng Jin and Zi-Liang Ye and Guanjing Xiong and Fan Zhang and Xiang Li and Mengwei Xu and Zhijun Li and Peng Li and Yang Liu and Yaqiong Zhang and Yunxin Liu},
  journal={ArXiv},
  year={2024},
  volume={abs/2401.05459},
  url={https://api.semanticscholar.org/CorpusID:266933252}
}

@article{Kaur2025SyntheticVE,
  title={Synthetic Voices: Evaluating the Fidelity of LLM-Generated Personas in Representing People’s Financial Wellbeing},
  author={Arshnoor Kaur and Amanda Aird and Harris Borman and Andrea Nicastro and Anna Leontjeva and Luiz Pizzato and Dan Jermyn},
  journal={Proceedings of the 33rd ACM Conference on User Modeling, Adaptation and Personalization},
  year={2025},
  url={https://api.semanticscholar.org/CorpusID:279323112}
}

@article{Cheng2023CoMPosTCA,
  title={CoMPosT: Characterizing and Evaluating Caricature in LLM Simulations},
  author={Myra Cheng and Tiziano Piccardi and Diyi Yang},
  journal={ArXiv},
  year={2023},
  volume={abs/2310.11501},
  url={https://api.semanticscholar.org/CorpusID:264288848}
}

@article{Hu2025SimulatingRS,
  title={Simulating Rumor Spreading in Social Networks using LLM Agents},
  author={Tianrui Hu and Dimitrios Liakopoulos and Xiwen Wei and Radu Marculescu and Neeraja Jayant Yadwadkar},
  journal={ArXiv},
  year={2025},
  volume={abs/2502.01450},
  url={https://api.semanticscholar.org/CorpusID:276107205}
}

@article{Ju2025TrajLLMAM,
  title={TrajLLM: A Modular LLM-Enhanced Agent-Based Framework for Realistic Human Trajectory Simulation},
  author={Chenlu Ju and Jiaxin Liu and Shobhit Sinha and Hao Xue and Flora D. Salim},
  journal={Companion Proceedings of the ACM on Web Conference 2025},
  year={2025},
  url={https://api.semanticscholar.org/CorpusID:276617707}
}

@article{Chuang2023EvaluatingLA,
  title={Evaluating LLM Agent Group Dynamics against Human Group Dynamics: A Case Study on Wisdom of Partisan Crowds},
  author={Yun-Shiuan Chuang and Siddharth Suresh and Nikunj Harlalka and Agam Goyal and Robert Hawkins and Sijia Yang and Dhavan Shah and Junjie Hu and Timothy T. Rogers},
  journal={ArXiv},
  year={2023},
  volume={abs/2311.09665},
  url={https://api.semanticscholar.org/CorpusID:275357913}
}

@article{Ng2024HowWC,
  title={How Well Can LLMs Echo Us? Evaluating AI Chatbots' Role-Play Ability with ECHO},
  author={Man Tik Ng and Hui Tung Tse and Jen-Tse Huang and Jingjing Li and Wenxuan Wang and Michael R. Lyu},
  journal={ArXiv},
  year={2024},
  volume={abs/2404.13957},
  url={https://api.semanticscholar.org/CorpusID:269293628}
}

@article{Li2016APN,
  title={A Persona-Based Neural Conversation Model},
  author={Jiwei Li and Michel Galley and Chris Brockett and Georgios P. Spithourakis and Jianfeng Gao and William B. Dolan},
  journal={ArXiv},
  year={2016},
  volume={abs/1603.06155},
  url={https://api.semanticscholar.org/CorpusID:2955580}
}

@article{Zhang2018PersonalizingDA,
  title={Personalizing Dialogue Agents: I have a dog, do you have pets too?},
  author={Saizheng Zhang and Emily Dinan and Jack Urbanek and Arthur Szlam and Douwe Kiela and Jason Weston},
  journal={ArXiv},
  year={2018},
  volume={abs/1801.07243},
  url={https://api.semanticscholar.org/CorpusID:6869582}
}

@article{Zhang2024GuidedPG,
  title={Guided Profile Generation Improves Personalization with LLMs},
  author={Jiarui Zhang},
  journal={ArXiv},
  year={2024},
  volume={abs/2409.13093},
  url={https://api.semanticscholar.org/CorpusID:272770339}
}

@article{Lambert2024TLU3P,
  title={T{\"U}LU 3: Pushing Frontiers in Open Language Model Post-Training},
  author={Nathan Lambert and Jacob Daniel Morrison and Valentina Pyatkin and Shengyi Huang and Hamish Ivison and Faeze Brahman and Lester James Validad Miranda and Alisa Liu and Nouha Dziri and Xinxi Lyu and Yuling Gu and Saumya Malik and Victoria Graf and Jena D. Hwang and Jiangjiang Yang and Ronan Le Bras and Oyvind Tafjord and Chris Wilhelm and Luca Soldaini and Noah A. Smith and Yizhong Wang and Pradeep Dasigi and Hanna Hajishirzi},
  journal={ArXiv},
  year={2024},
  volume={abs/2411.15124},
  url={https://api.semanticscholar.org/CorpusID:274192505}
}

@article{Jandaghi2023FaithfulPC,
  title={Faithful Persona-based Conversational Dataset Generation with Large Language Models},
  author={Pegah Jandaghi and XiangHai Sheng and Xinyi Bai and Jay Pujara and Hakim Sidahmed},
  journal={ArXiv},
  year={2023},
  volume={abs/2312.10007},
  url={https://api.semanticscholar.org/CorpusID:266335707}
}

@article{Li2025From1U,
  title={From 1,000,000 Users to Every User: Scaling Up Personalized Preference for User-level Alignment},
  author={Jia-Nan Li and Jian Guan and Songhao Wu and Wei Wu and Rui Yan},
  journal={ArXiv},
  year={2025},
  volume={abs/2503.15463},
  url={https://api.semanticscholar.org/CorpusID:277113478}
}

@software{nvidia-Nemotron-Personas,
  author = {Meyer, Yev and Corneil, Dane},
  title = {{Nemotron-Personas}: Synthetic Personas Aligned to Real-World Distributions
},
  month = {June},
  year = {2025},
  url = {https://huggingface.co/datasets/nvidia/Nemotron-Personas}
}

@misc{ipip50,
  title = {Possible Questionnaire Format for Administering the 50-Item Set of IPIP Big-Five Factor Markers},
  author = {{International Personality Item Pool}},
  howpublished = {\url{https://ipip.ori.org/new_ipip-50-item-scale.htm}},
}

@article{Chan2024ScalingSD,
  title={Scaling Synthetic Data Creation with 1,000,000,000 Personas},
  author={Xin Chan and Xiaoyang Wang and Dian Yu and Haitao Mi and Dong Yu},
  journal={ArXiv},
  year={2024},
  volume={abs/2406.20094},
  url={https://api.semanticscholar.org/CorpusID:270845490}
}

@article{goldberg1992development,
  title={The development of markers for the Big-Five factor structure.},
  author={Goldberg, Lewis R},
  journal={Psychological assessment},
  volume={4},
  number={1},
  pages={26},
  year={1992},
  publisher={American Psychological Association}
}

@misc{openpsychometrics2019,
  title = {Big Five Personality Test using the IPIP Big-Five Factor Markers},
  author = {{Open-Source Psychometrics Project}},
  howpublished = {\url{https://openpsychometrics.org/tests/IPIP-BFFM/}}
}

@inproceedings{ZhangLMQLWLCLWXW25,
  author       = {Zeyu Zhang and
                  Jianxun Lian and
                  Chen Ma and
                  Yaning Qu and
                  Ye Luo and
                  Lei Wang and
                  Rui Li and
                  Xu Chen and
                  Yankai Lin and
                  Le Wu and
                  Xing Xie and
                  Ji{-}Rong Wen},
  editor       = {Luis Chiruzzo and
                  Alan Ritter and
                  Lu Wang},
  title        = {TrendSim: Simulating Trending Topics in Social Media Under Poisoning
                  Attacks with LLM-based Multi-agent System},
  booktitle    = {Findings of the Association for Computational Linguistics: {NAACL}
                  2025, Albuquerque, New Mexico, USA, April 29 - May 4, 2025},
  pages        = {2930--2949},
  publisher    = {Association for Computational Linguistics},
  year         = {2025},
  url          = {https://doi.org/10.18653/v1/2025.findings-naacl.160},
  doi          = {10.18653/V1/2025.FINDINGS-NAACL.160},
  bibsource    = {dblp computer science bibliography, https://dblp.org}
}

@article{wang2025personality,
  title={Personality Structured Interview for Large Language Model Simulation in Personality Research},
  author={Wang, Pengda and Zou, Huiqi and Chen, Hanjie and Sun, Tianjun and Xiao, Ziang and Oswald, Frederick L},
  journal={arXiv preprint arXiv:2502.12109},
  year={2025}
}

@article{wang2025yulan,
  title={Yulan-onesim: Towards the next generation of social simulator with large language models},
  author={Wang, Lei and Gao, Heyang and Bo, Xiaohe and Chen, Xu and Wen, Ji-Rong},
  journal={arXiv preprint arXiv:2505.07581},
  year={2025}
}

@article{zhang2025socioverse,
  title={Socioverse: A world model for social simulation powered by llm agents and a pool of 10 million real-world users},
  author={Zhang, Xinnong and Lin, Jiayu and Mou, Xinyi and Yang, Shiyue and Liu, Xiawei and Sun, Libo and Lyu, Hanjia and Yang, Yihang and Qi, Weihong and Chen, Yue and others},
  journal={arXiv preprint arXiv:2504.10157},
  year={2025}
}

@article{binz2025foundation,
  title={A foundation model to predict and capture human cognition},
  author={Binz, Marcel and Akata, Elif and Bethge, Matthias and Br{\"a}ndle, Franziska and Callaway, Fred and Coda-Forno, Julian and Dayan, Peter and Demircan, Can and Eckstein, Maria K and {\'E}ltet{\H{o}}, No{\'e}mi and others},
  journal={Nature},
  pages={1--8},
  year={2025},
  publisher={Nature Publishing Group UK London}
}

@article{strachan2024testing,
  title={Testing theory of mind in large language models and humans},
  author={Strachan, James WA and Albergo, Dalila and Borghini, Giulia and Pansardi, Oriana and Scaliti, Eugenio and Gupta, Saurabh and Saxena, Krati and Rufo, Alessandro and Panzeri, Stefano and Manzi, Guido and others},
  journal={Nature Human Behaviour},
  volume={8},
  number={7},
  pages={1285--1295},
  year={2024},
  publisher={Nature Publishing Group UK London}
}

@article{lu2025tom,
  title={Tom-rl: Reinforcement learning unlocks theory of mind in small llms},
  author={Lu, Yi-Long and Zhang, Chunhui and Song, Jiajun and Fan, Lifeng and Wang, Wei},
  journal={arXiv e-prints},
  pages={arXiv--2504},
  year={2025}
}

@article{Duckworth2007Grit,
  title   = {Grit: Perseverance and Passion for Long-Term Goals},
  author  = {Duckworth, Angela L. and Peterson, Christopher and Matthews, Michael D. and Kelly, Dennis R.},
  journal = {Journal of Personality and Social Psychology},
  year    = {2007},
  volume  = {92},
  number  = {6},
  pages   = {1087--1101},
  doi     = {10.1037/0022-3514.92.6.1087}
}

@article{Strathman1994CFC,
  title   = {The Consideration of Future Consequences: Weighing Immediate and Distant Outcomes of Behavior},
  author  = {Strathman, Alan and Gleicher, Faye and Boninger, David S. and Edwards, Charles S.},
  journal = {Journal of Personality and Social Psychology},
  year    = {1994},
  volume  = {66},
  number  = {4},
  pages   = {742--752},
  doi     = {10.1037/0022-3514.66.4.742}
}

@misc{OpenPsychometrics2019Firstborn,
  title        = {Firstborn Personality Scale Responses},
  author       = {{OpenPsychometrics.org}},
  year         = {2019},
  howpublished = {Online},
  url          = {https://openpsychometrics.org/tests/birthorder/development/}
}

@misc{Inglehart2020WVS,
  title        = {World Values Survey: Wave 7 Country-Pooled Datafile},
  author       = {Inglehart, Ronald and Haerpfer, Christian and Moreno, Ana and Welzel, Christian and Kizilova, Katerina and Diez-Medrano, Juan and Lagos, Max and Norris, Pippa and Ponarin, Eduard and Puranen, Boris},
  year         = {2020},
  howpublished = {Online Dataset},
  url          = {http://www.worldvaluessurvey.org/WVSDocumentationWV7.jsp}
}

@article{wu2025llm,
  title={LLM-Based Social Simulations Require a Boundary},
  author={Wu, Zengqing and Peng, Run and Ito, Takayuki and Xiao, Chuan},
  journal={arXiv preprint arXiv:2506.19806},
  year={2025}
}

@book{villani2008optimal,
  title={Optimal transport: old and new},
  author={Villani, C{\'e}dric and others},
  volume={338},
  year={2008},
  publisher={Springer}
}

@article{wang2024statistical,
  title={Statistical and computational guarantees of kernel max-sliced wasserstein distances},
  author={Wang, Jie and Boedihardjo, March and Xie, Yao},
  journal={arXiv preprint arXiv:2405.15441},
  year={2024}
}

@article{lobashev2025color,
  title={Color Conditional Generation with Sliced Wasserstein Guidance},
  author={Lobashev, Alexander and Larchenko, Maria and Guskov, Dmitry},
  journal={arXiv preprint arXiv:2503.19034},
  year={2025}
}

@article{nguyen2022revisiting,
  title={Revisiting sliced Wasserstein on images: From vectorization to convolution},
  author={Nguyen, Khai and Ho, Nhat},
  journal={Advances in Neural Information Processing Systems},
  volume={35},
  pages={17788--17801},
  year={2022}
}

@inproceedings{ren2025few,
  title={Few-shot LLM Synthetic Data with Distribution Matching},
  author={Ren, Jiyuan and Du, Zhaocheng and Wen, Zhihao and Jia, Qinglin and Dai, Sunhao and Wu, Chuhan and Dong, Zhenhua},
  booktitle={Companion Proceedings of the ACM on Web Conference 2025},
  pages={432--441},
  year={2025}
}

@article{kalinke2022maximum,
  title={Maximum mean discrepancy on exponential windows for online change detection},
  author={Kalinke, Florian and Heyden, Marco and Gntuni, Georg and Fouch{\'e}, Edouard and B{\"o}hm, Klemens},
  journal={arXiv preprint arXiv:2205.12706},
  year={2022}
}

@article{veeramacheneni2023fr,
  title={Fr$\backslash$'echet Wavelet Distance: A Domain-Agnostic Metric for Image Generation},
  author={Veeramacheneni, Lokesh and Wolter, Moritz and Kuehne, Hildegard and Gall, Juergen},
  journal={arXiv preprint arXiv:2312.15289},
  year={2023}
}

@inproceedings{jayasumana2024rethinking,
  title={Rethinking fid: Towards a better evaluation metric for image generation},
  author={Jayasumana, Sadeep and Ramalingam, Srikumar and Veit, Andreas and Glasner, Daniel and Chakrabarti, Ayan and Kumar, Sanjiv},
  booktitle={Proceedings of the IEEE/CVF Conference on Computer Vision and Pattern Recognition},
  pages={9307--9315},
  year={2024}
}

@misc{pearson_wikipedia,
  title = {Pearson correlation coefficient},
  howpublished = {Wikipedia, accessed 2025‑07‑25},
  note = {“Measures linear correlation... between two sets of data…”},
  url = {https://en.wikipedia.org/wiki/Pearson_correlation_coefficient}
}

@article{pratelli2025evaluating,
  title={Evaluating the Simulation of Human Personality-Driven Susceptibility to Misinformation with LLMs},
  author={Pratelli, Manuel and Petrocchi, Marinella},
  journal={arXiv preprint arXiv:2506.23610},
  year={2025}
}

@inproceedings{schler2006effects,
    title={Effects of age and gender on blogging.},
    author={Schler, Jonathan and Koppel, Moshe and Argamon, Shlomo and Pennebaker, James W},
    booktitle={AAAI spring symposium: Computational approaches to analyzing weblogs},
    volume={6},
    pages={199--205},
    year={2006}
}

@dataset{enryu-twitter100m-tweets,
  author = {Enryu},
  title = {{twitter100m\_tweets}: A Large-Scale Dataset of Tweets and Users for Political and Sentiment Analysis},
  year = {2023},
  month = {May},
  url = {https://huggingface.co/datasets/enryu43/twitter100m_tweets},
  }

@dataset{huggingfacegeclm-reddit-submissions,
  author = {Lefourrier, Clément},
  title = {{REDDIT\_submissions}: High-Quality Submissions from 50 Subreddits (2006–2023)},
  year = {2023},
  month = {January},
  url = {https://huggingface.co/datasets/HuggingFaceGECLM/REDDIT_submissions}
  
}

@article{qwen3embedding,
  title={Qwen3 Embedding: Advancing Text Embedding and Reranking Through Foundation Models},
  author={Zhang, Yanzhao and Li, Mingxin and Long, Dingkun and Zhang, Xin and Lin, Huan and Yang, Baosong and Xie, Pengjun and Yang, An and Liu, Dayiheng and Lin, Junyang and Huang, Fei and Zhou, Jingren},
  journal={arXiv preprint arXiv:2506.05176},
  year={2025}
}

@article{wang2024not,
  title={Not yet: Large language models cannot replace human respondents for psychometric research},
  author={Wang, Pengda and Zou, Huiqi and Yan, Zihan and Guo, Feng and Sun, Tianjun and Xiao, Ziang and Zhang, Bo},
  journal={OSF Preprint: https://doi. org/10.31219/osf. io/rwy9b},
  year={2024}
}

@article{Kolluri2025FinetuningLF,
  title={Finetuning LLMs for Human Behavior Prediction in Social Science Experiments},
  author={Akaash Kolluri and Shengguang Wu and Joon Sung Park and Michael S. Bernstein},
  journal={ArXiv},
  year={2025},
  volume={abs/2509.05830},
  url={https://api.semanticscholar.org/CorpusID:281202869}
}

@article{xie2024can,
  title={Can large language model agents simulate human trust behavior?},
  author={Xie, Chengxing and Chen, Canyu and Jia, Feiran and Ye, Ziyu and Lai, Shiyang and Shu, Kai and Gu, Jindong and Bibi, Adel and Hu, Ziniu and Jurgens, David and others},
  journal={Advances in neural information processing systems},
  volume={37},
  pages={15674--15729},
  year={2024}
}

@inproceedings{moon2024virtual,
    title = "Virtual Personas for Language Models via an Anthology of Backstories",
    author = "Moon, Suhong  and
      Abdulhai, Marwa  and
      Kang, Minwoo  and
      Suh, Joseph  and
      Soedarmadji, Widyadewi  and
      Behar, Eran Kohen  and
      Chan, David M.",
    editor = "Al-Onaizan, Yaser  and
      Bansal, Mohit  and
      Chen, Yun-Nung",
    booktitle = "Proceedings of the 2024 Conference on Empirical Methods in Natural Language Processing",
    month = nov,
    year = "2024",
    address = "Miami, Florida, USA",
    publisher = "Association for Computational Linguistics",
    url = "https://aclanthology.org/2024.emnlp-main.1110/",
    doi = "10.18653/v1/2024.emnlp-main.1110",
    pages = "19864--19897"
}

@inproceedings{rahimzadeh2026synthia,
    title = "Synthia: Scalable Grounded Persona Generation from Social Media Data",
    author = "Rahimzadeh, Vahid  and
      Moosavi Monazzah, Erfan  and
      Pilehvar, Mohammad Taher  and
      Yaghoobzadeh, Yadollah",
    editor = "Liakata, Maria  and
      Moreira, Viviane P.  and
      Zhang, Jiajun  and
      Jurgens, David",
    booktitle = "Proceedings of the 64th Annual Meeting of the {A}ssociation for {C}omputational {L}inguistics (Volume 1: Long Papers)",
    month = jul,
    year = "2026",
    address = "San Diego, California, United States",
    publisher = "Association for Computational Linguistics",
    url = "https://aclanthology.org/2026.acl-long.2134/",
    doi = "10.18653/v1/2026.acl-long.2134",
    pages = "45983--46006",
    ISBN = "979-8-89176-390-6"
}

@inproceedings{lu2026multiturn,
    title = "Can {LLM} Agents Simulate Multi-Turn Human Behavior? Evidence from Real Online Customer Behavior Data",
    author = "Lu, Yuxuan  and
      Huang, Jing  and
      Han, Yan  and
      Yao, Bingsheng  and
      Bei, Sisong  and
      Xie, Yaochen  and
      Sang, Yisi  and
      He, Qi  and
      Wang, Dakuo",
    editor = "Liakata, Maria  and
      Moreira, Viviane P.  and
      Zhang, Jiajun  and
      Jurgens, David",
    booktitle = "Proceedings of the 64th Annual Meeting of the {A}ssociation for {C}omputational {L}inguistics (Volume 1: Long Papers)",
    month = jul,
    year = "2026",
    address = "San Diego, California, United States",
    publisher = "Association for Computational Linguistics",
    url = "https://aclanthology.org/2026.acl-long.2034/",
    doi = "10.18653/v1/2026.acl-long.2034",
    pages = "43961--43977",
    ISBN = "979-8-89176-390-6"
}

@Book{Lei2026HumanLLMTP,
 author = {Yuxuan Lei and Tian-Fu Wang and Jian-Xun Lian and Zheng-Yu Hu and De-Fu Lian and Xing Xie},
 booktitle = {Knowledge Discovery and Data Mining},
 journal = {Proceedings of the 32nd ACM SIGKDD Conference on Knowledge Discovery and Data Mining V.1},
 title = {HumanLLM: Towards Personalized Understanding and Simulation of Human Nature},
 year = {2026}
}

@article{Brener2024OverviewAM,
  author = {Brener, Nancy D. and Mpofu, Jonetta J. and
            Krause, Kathleen H. and Everett Jones, Sherry and
            Thornton, Jemekia E. and Myles, Zachary and
            Harris, William A. and Chyen, David and
            Lim, Connie and Arrey, Loredona and
            Mbaka, Cecily K. and Trujillo, Lindsay and
            Shanklin, Shari L. and Smith-Grant, Jennifer and
            Whittle, Lisa and McKinnon, Izraelle I. and
            Washington, Malaika and Queen, Barbara E. and
            Roberts, Alice M.},
  title = {Overview and Methods for the {Youth Risk Behavior
           Surveillance System}---{United States}, 2023},
  journal = {MMWR Supplements},
  year = {2024},
  volume = {73},
  number = {4},
  pages = {1--12},
  doi = {10.15585/mmwr.su7304a1},
  url = {https://www.cdc.gov/mmwr/volumes/73/su/su7304a1.htm}
}
\clearpage

\appendix

\section{Further Implementation Details}
\label{app:further-implet-detail}
\subsection{Experimental Setup}
\label{app:Experimental-Setup}
All experiments were conducted on a high-performance computing server equipped with an AMD EPYC 7V12 64-Core Processor, 512 GB of system memory, and 96 logical CPUs. The system includes eight NVIDIA A100 GPUs, each with 80 GB of memory.
Seed persona mining uses Llama-3.3-70B-Instruct. The response models use the following HuggingFace checkpoints: for Llama-3-70B, we use {meta-llama/Meta-Llama-3-70B-Instruct}\footnote{\url{https://huggingface.co/meta-llama/Meta-Llama-3-70B-Instruct}}; for Qwen2.5-72B, we use {Qwen/Qwen2.5-72B-Instruct}\footnote{\url{https://huggingface.co/Qwen/Qwen2.5-72B-Instruct}}; and for Phi-4, we use {microsoft/phi-4}\footnote{\url{https://huggingface.co/microsoft/phi-4}}.
Our seed persona mining pipeline draws from three open-source datasets. Reddit submissions are taken from this collection\footnote{\url{https://huggingface.co/datasets/HuggingFaceGECLM/REDDIT_submissions}}, covering 50 topical subreddits. The Twitter corpus consists of original tweets from 100M users, hosted here\footnote{\url{https://huggingface.co/datasets/enryu43/twitter100m_users}}. The blog data is based on the Blog Authorship Corpus\footnote{\url{https://huggingface.co/datasets/barilan/blog_authorship_corpus}}, which includes entries from over 19,000 users.
For population-level persona sampling via Importance-based Selection and Optimal Transport (Section~\ref{subsec:IS_OT}), we apply a Gaussian KDE with bandwidth 0.20 and retain the 70\% of candidate personas with the highest importance weights. OT weights are computed in batches of size 10{,}000 with entropic regularization $\varepsilon=0.08$ (scaled to the median cost). The Sinkhorn algorithm is run for 250 iterations.
For a detailed ablation and sensitivity analysis of these hyperparameters, please refer to Appendix~\ref{app:hyper-ablation}.
Finally, for embedding model training, we follow the SWIFT recipe from the Qwen3-Embedding repository, available at this link\footnote{\url{https://github.com/QwenLM/Qwen3-Embedding/blob/main/docs/training/SWIFT.md}}.
Code is available at \url{https://github.com/PersonaGen/PersonaGen}.

\subsection{Notation}

Table~\ref{tab:notA} and Table~\ref{tab:notB} summarize the notations used in our paper.

\begin{table*}[t]
\caption{Summary of notation (A): Seed mining, LLM responses, and KDE-based Importance-based Selection.}
\centering
\scriptsize
\setlength{\tabcolsep}{3.5pt}
\begin{adjustbox}{max width=\linewidth}
\begin{tabularx}{\linewidth}{@{}lX@{}}
\toprule
\textbf{Symbol} & \textbf{Description} \\
\midrule
\multicolumn{2}{@{}l}{\textbf{Part I: Basic objects \& LLM-induced responses}} \\
\midrule
$\mathcal{P}=\{p_i\}_{i=1}^{N}$ & Set of textual personas (Sec.~\ref{subsec:SPG}) \\
$N$ & Number of personas in $\mathcal{P}$ \\
$s$ & Aggregated long-form user text (concatenated posts) \\
$\mathcal{G}$ & LLM persona summarizer mapping $s \mapsto p_i$ \\
$p_i=\mathcal{G}(s)$ & Narrative persona synthesized from $s$ \\
$\mathcal{Q}=\{q_k\}_{k=1}^{d}$ & Psychometric items (e.g., IPIP Big Five) \\
$d$ & Number of items / response dimension \\
\texttt{LLM} & Frozen LLM mapping $(p_i,q_k)$ to a scalar response \\
$\|$ & Textual concatenation operator \\
$x_{ik}$ & Model response to $(p_i,q_k)$ \\
$x_i\in\mathbb{R}^d$ & Response vector for persona $p_i$ \\
$X\in\mathbb{R}^{N\times d}$ & Persona response matrix with rows $x_i$ \\
$Y=\{y_j\}_{j=1}^{M}$ & Human response vectors from surveys \\
$M$ & Number of human response samples \\
$r_{\texttt{persona}}$ & Empirical distribution of $\{x_i\}_{i=1}^N$ \\
$r_{\texttt{human}}$ & Empirical distribution of $\{y_j\}_{j=1}^M$ \\
\midrule
\multicolumn{2}{@{}l}{\textbf{Part II: KDE-based Importance-based Selection}} \\
\midrule
$h$ & Bandwidth parameter for KDE \\
$I$ & Identity matrix \\
$\mathcal{N}(z;h^2 I)$ & Multivariate Gaussian kernel density \\
$\hat r_{\texttt{persona}}$ & KDE estimate of $r_{\texttt{persona}}$ \\
$\hat r_{\texttt{human}}$ & KDE estimate of $r_{\texttt{human}}$ \\
$w_i^{\mathrm{IS}}$ & IS weight: $\hat r_{\texttt{human}}(x_i)/\hat r_{\texttt{persona}}(x_i)$ \\
$i_{(k)}$ & Index of the persona with the $k$-th highest importance weight \\
$N^{\dagger}$ & Number of candidates retained by IS, $\lfloor 0.7N\rfloor$ by default \\
$\mathcal{P}^{\dagger}$ & Candidate persona subset after IS \\
$X^{\dagger}$ & Response vectors of $\mathcal{P}^{\dagger}$ \\
\bottomrule
\end{tabularx}
\end{adjustbox}
\vspace{-2mm}
\label{tab:notA}
\end{table*}

\begin{table*}[t]
\caption{Summary of notation (B): Entropic OT alignment and Group-specific construction.}
\centering
\scriptsize
\setlength{\tabcolsep}{3.5pt}
\begin{adjustbox}{max width=\linewidth}
\begin{tabularx}{\linewidth}{@{}lX@{}}
\toprule
\textbf{Symbol} & \textbf{Description} \\
\midrule
\multicolumn{2}{@{}l}{\textbf{Part III: Optimal Transport alignment}} \\
\midrule
$\omega_k$ & Item weight for prompt $q_k$ \\
$C\in\mathbb{R}^{N^{\dagger}\times M}$ & Cost matrix, $C_{ij}=\sum_{k}\omega_k(x^{\dagger}_{ik}-y_{jk})^2$ \\
$\varepsilon$ & Entropic regularization parameter in OT \\
$\mathbf{a}^{\dagger},\,\mathbf{b}$ & Uniform marginals on candidates / human samples \\
$\mathbf{1}_n$ & All-ones vector of length $n$ \\
$\Pi(\mathbf{a}^{\dagger},\mathbf{b})$ & Feasible couplings with given marginals \\
$\langle C,\Gamma\rangle$ & Frobenius inner product $\sum_{ij} C_{ij}\Gamma_{ij}$ \\
$K$ & Gibbs kernel: $K_{ij}=\exp(-C_{ij}/\varepsilon)$ \\
$u, v$ & Sinkhorn scaling vectors \\
$\Gamma,\,\Gamma^\star$ & Transport plan / optimal plan \\
$\operatorname{diag}(\cdot)$ & Diagonal matrix formed from a vector \\
$a_i^{\dagger}$ & Candidate marginal mass, $\sum_j\Gamma_{ij}^{\star}=1/N^{\dagger}$ \\
$\bar c_i$ & Expected transport cost, $(a_i^{\dagger})^{-1}\sum_j\Gamma_{ij}^{\star}C_{ij}$ \\
$\tau$ & Temperature for OT-derived sampling weights \\
$w_i^{\mathrm{OT}}$ & OT-derived weight for $x_i^{\dagger}$ \\
$\pi_i^{\mathrm{OT}}$ & Sampling probability induced by $w_i^{\mathrm{OT}}$ \\
$N'$ & Final number of personas sampled after OT \\
$\mathcal{P}'$ & Final population-aligned persona subset \\
\midrule
\multicolumn{2}{@{}l}{\textbf{Part IV: Group-specific retrieval \& generation}} \\
\midrule
$q$ & Natural-language group/query \\
$\mathrm{Embed}(\cdot)$ & Learned embedding model for queries/personas \\
$e_q,\,e_i,\,e_p$ & Embeddings of $q$, persona $p_i$, and a generic persona \\
$\mathrm{sim}(\cdot,\cdot)$ & Cosine similarity between embeddings \\
$K_{\text{ret}}$ & Top-$K$ retrieval size (distinct from OT kernel $K$) \\
$\mathcal{P}_{\text{seed}}(q)$ & Retrieved seed personas for $q$ \\
$\mathcal{P}_{\text{group}}(q)$ & Generated group-specific personas conditioned on seeds \\
$e_p^{+}$ & Positive persona embedding in contrastive loss \\
$p^{-}$ & Negative personas in contrastive loss \\
$\mathcal{L}$ & Contrastive loss:
$-\log \frac{e^{\mathrm{sim}(e_q,e_p^{+})}}{e^{\mathrm{sim}(e_q,e_p^{+})}+\sum_{p^-} e^{\mathrm{sim}(e_q,e_p^-)}}$ \\
\bottomrule
\end{tabularx}
\end{adjustbox}
\vspace{-2mm}
\label{tab:notB}
\end{table*}

\subsection{Datasets}
\label{abs:dataset}

\noindent \textbf{IPIP Big Five Test}~\citep{ipip50}.
The Big Five Inventory consists of 50 self‑descriptive statements (e.g., "I see myself as someone who is talkative") rated on a 5-point Likert scale.
It quantifies five broad personality dimensions-Openness, Conscientiousness, Extraversion, Agreeableness, and Neuroticism-providing stable trait profiles.
The instrument is used to predict outcomes in areas such as job performance, relationship quality, and health behaviors, and to examine correlations between personality and cognitive, emotional, or social variables.

\noindent \textbf{CFCS}~\citep{Strathman1994CFC}. The Consideration of Future Consequences Scale (CFCS) is a 12‑item self‑report instrument rated on a 5‑point Likert scale designed to quantify individual differences in the extent to which people consider distant versus immediate outcomes of their actions. The CFCS score predicts a wide range of real‑world behaviors, such as health maintenance, financial planning, environmental stewardship, and academic persistence, making it a versatile benchmark for assessing temporal orientation and decision‑making dynamics in both clinical and applied research contexts.

\noindent \textbf{FBPS}~\citep{OpenPsychometrics2019Firstborn}. The Firstborn Personality Scale is a 76-item measure designed to capture personality traits statistically associated with being firstborn, without explicitly referencing birth order. The items reflect a profile marked by intellectual engagement, verbal interests, competitiveness, assertiveness, and cognitive intensity-traits such as enjoying complex reading, using advanced vocabulary, following politics, and striving to outdo others. Developed through large-sample analysis, the FBPS offers a concise snapshot of a driven, literate, and analytically minded personality style.

\noindent \textbf{Duckworth.}~\citep{Duckworth2007Grit}. The Duckworth Grit Scale assesses perseverance and passion for long‑term goals through 12 self‑report items (e.g., "I finish whatever I begin," "Setbacks don’t discourage me"), providing a validated measure of sustained effort and consistency of interest. In this paper, Duckworth serves as a self‑regulation benchmark to verify that our persona alignment method captures realistic distributions of grit and motivational persistence .

\noindent \textbf{WVS}~ \citep{Inglehart2020WVS}. The World Values Survey administers 200-300 questions on political attitudes, social norms, religious beliefs, and subjective well‑being (e.g., "Overall, how happy are you?") across over 100 countries every 5-10 years. It tracks longitudinal changes in values related to democracy, gender roles, economic security, and quality of life. WVS supports comparative research in sociology, political science, and international development.

\noindent \textbf{YRBSS}~\citep{Brener2024OverviewAM}. We use the 2023 national Youth Risk Behavior Survey, which includes 20,103 U.S. high school students and is conducted by the CDC to assess health-risk behaviors. Topics include tobacco/alcohol/drug use, sexual activity, mental health (e.g., suicidality), nutrition, physical activity, and violence exposure. The goal is to monitor trends and inform public health interventions targeting adolescents.

\noindent \textbf{SocSci}~\citep{Kolluri2025FinetuningLF}. SocSci is a large-scale dataset of 2.9M individual responses from 400,491 participants across 210 open-source social science experiments, with rich demographics and standardized text descriptions of questions, conditions, and responses, used to finetune and evaluate LLMs for predicting human response distributions and demographic bias.

\begin{table*}[t]
\caption{Statistics of datasets used for alignment and evaluation.}
\label{tab:dataset-stats}
\centering
\scriptsize
\renewcommand{\arraystretch}{1.05}

\begin{tabular*}{\textwidth}{@{\extracolsep{\fill}} l r r r @{}}
\toprule
\textbf{Dataset} & \textbf{\# Questions} &
\makecell{\textbf{\# Human} \\ \textbf{Responses}} &
\makecell{\textbf{\# Countries} \\ \textbf{and Regions}} \\
\midrule
Big Five (IPIP)     & 50  & 1,015,341 & 223 \\
CFCS                & 12  &   15,034  & 144 \\
FBPS                & 76  &   41,841  & 169 \\
Duckworth           & 12  &    4,270  & 111 \\
YRBSS (2023)        & 107 &   20,103  &   1 \\
WVS (Full)          & 294 &   97,221  &  66 \\
WVS -- East Asia     & 250 &   11,593  &   7 \\
WVS -- North America & 250 &    8,355  &   3 \\
WVS -- Europe        & 250 &    6,282  &   3 \\
SocSci              & 5,998 & 2,900,000 & 1 \\
\bottomrule
\end{tabular*}
\vspace{-1mm}
\end{table*}

\subsection{Dataset Statistics}
\label{sec:dataset-statistics}

Table~\ref{tab:dataset-stats} summarizes all datasets used in our study. For the World Values Survey (WVS), we construct three regional subsets to evaluate the {Group-Specific Persona Adaptation} module, as described in Section~\ref{subsec:group_specific}. These subsets correspond to the full set of countries available within each target region in the WVS dataset: {East Asia}, including China, Japan, South Korea, Hong Kong, Mongolia, Taiwan, and Macao; {North America}, including Canada, United States, and Mexico; and {Europe}, including Germany, Netherlands, and United Kingdom.

\subsection{Query Generation from Personas for Embedding Training}
\label{app:query}

\noindent\textbf{Motivation.}
To construct positive \emph{(query, persona)} pairs for the contrastive encoder (§\ref{subsec:PSM}), we automatically derive two natural language queries from each persona $p_i\!\in\!\mathcal{P}'$. The first is a \emph{detailed} query that captures fine-grained persona traits, and the second is a \emph{broad} query that summarizes only high-level characteristics. 
We generate both queries using LLM prompting, with one prompt designed for detailed descriptions (see Prompt~\ref{lst:prompt_longquery}) and another for coarse-grained abstractions (see Prompt~\ref{lst:prompt_shortquery}). Each query is paired with the original persona for training the embedding model using the contrastive objective defined in Eq.~\ref{eq:contrastive}.

\begin{lstlisting}[caption={Prompt for detailed (long) query generation},label={lst:prompt_longquery}]
You are a helpful assistant tasked with generating queries based on detailed persona descriptions. 
Your goal is to generate concise queries that describe the key traits of the persona, such as age, gender, 
emotional tone, and significant interests or lifestyle. The query should be specific but short, capturing the 
persona's core characteristics without going into excessive detail.

Persona: "The user is a young, likely female, living in a town she affectionately calls 'peach town', working 
multiple jobs, including one at a gym and another at a cleaners, and values relaxation and enjoyment of simple 
pleasures, like good food and sunny days, while navigating life's changes and emotional ups and downs with an 
introspective and slightly moody tone."
Generated query: "Generate a young female persona who works multiple jobs, values relaxation and simple pleasures, 
and has an introspective and slightly moody tone."

Your generated query should capture the persona's essential traits in a clear and concise manner, avoiding 
unnecessary complexity.

Only return the generated query, nothing else.

Based on the following persona description, generate a concise query that describes a persona with similar traits 
and characteristics. Focus on the persona's key traits like age, gender, emotional tone, and major activities or 
interests, keeping the query short and to the point.

Persona description:
{persona}

Generated query:
Only return the generated query, nothing else.
\end{lstlisting}
\begin{lstlisting}[caption={Prompt for broad (short) query generation},label={lst:prompt_shortquery}]
You are a helpful assistant tasked with generating broader queries based on persona descriptions. 
Your goal is to generate simple and concise queries that describe the basic traits of the persona, such as age, 
gender, and general interests or activities. The query should be high-level, focusing on the core attributes, and 
should be short without diving into personal or emotional details.

Persona: "The user is a young, likely female, living in a town she affectionately calls 'peach town', working 
multiple jobs, including one at a gym and another at a cleaners, and values relaxation and enjoyment of simple 
pleasures, like good food and sunny days, while navigating life's changes and emotional ups and downs with an 
introspective and slightly moody tone."
Generated query: "Generate a young female persona who works multiple jobs."

Your generated query should focus on broad characteristics and general interests, and should be brief without 
over-explaining.

Only return the generated query, nothing else.

Based on the following persona description, generate a broad, concise query that describes a persona with similar 
traits and characteristics. Focus on the basic attributes such as age, gender, and major interests or lifestyle, 
keeping the query short and general.

Persona description:
{persona}

Generated query:
Only return the generated query, nothing else.
\end{lstlisting}

\noindent\textbf{Training usage.}
For each persona we generate one detailed query with Prompt~\ref{lst:prompt_longquery} and one broad query with Prompt~\ref{lst:prompt_shortquery}. Both are encoded and paired with the same persona embedding in the contrastive loss of Eq.~\eqref{eq:contrastive}, while hard negatives are mined as described in the main text. 
Full tuning hyper-parameter details are provided in Table~\ref{tab:embedding_hyperparams}.

\begin{table}[t]
\caption{\textbf{Contrastive embedding model fine-tuning hyper-parameters.}}
\centering
\renewcommand{\arraystretch}{1.05}
\setlength{\tabcolsep}{4pt}
\begin{adjustbox}{max width=\linewidth}
\begin{tabular}{@{}l l@{}}
\toprule
\textbf{Hyper-parameter}      & \textbf{Assignment} \\
\midrule
Number of epochs              & 2 \\
Batch size per device         & 4 \\
Gradient accumulation         & 4 \\
Effective batch size          & 32 \\
Learning rate                 & 6e-6 \\
Precision                     & FP16 \\
Optimizer                     & Deepspeed (ZeRO-3) \\
\bottomrule
\end{tabular}
\end{adjustbox}
\vspace{-2mm}
\label{tab:embedding_hyperparams}
\end{table}

\subsection{LLM-based Filtering of Semantically Matched Negatives}
\label{app:negfilter}

To improve the quality of negative sampling during training (§\ref{subsec:PSM}), we apply an additional filtering step to eliminate false negatives personas that are semantically aligned with the query but incorrectly treated as negatives. Specifically, for each query $q_i$, we examine two sets of candidate negatives: the top-$N$ most similar personas (in embedding space) and $N$ randomly sampled personas. Both sets may contain cases that match the query, so we prompt a large language model to verify alignment. Given a query-persona pair $(q_i, \tilde{p})$, the LLM is asked to return a strict binary response: \texttt{YES} if the persona clearly matches the query, and \texttt{NO} otherwise. Only those pairs labeled \texttt{NO} are retained as valid negatives. The prompt template used for this filtering step is shown in Prompt~\ref{lst:prompt_negfilter}.

\begin{lstlisting}[caption={Prompt for LLM-based semantic match filtering},label={lst:prompt_negfilter}]
You are an expert evaluator tasked with determining whether a user persona matches a given query or description.

Your job is to carefully analyze if the persona accurately represents someone who would be described by the query. Consider:

Key characteristics mentioned in the query

Interests, behaviors, and preferences

Demographics if mentioned

Overall alignment between query and persona

Respond with ONLY "YES" if the persona matches the query, or "NO" if it doesn't match.
Be strict in your evaluation - only respond "YES" if there is a clear and strong alignment.


Query: Generate a crafty individual who values convenience and quality, frequently purchases art and sewing supplies, and enjoys DIY projects.

Persona: The user is a crafty individual who values quality and functionality, often purchasing items for themselves and as gifts for others, including artists and sewers, indicating a supportive and generous nature, with a keen interest in sewing, drawing, and art supplies.

Does this persona match the query? Respond with only "YES" or "NO".
\end{lstlisting}

\subsection{Prompts for Group-Specific Persona Adaptation}
\label{app:group_prompts}
\noindent\textbf{Prompt Details.}
We adapt personas from the globally aligned pool $\mathcal{P}'$ to region- or group-specific targets by prompting an LLM to revise a \emph{seed persona} under task-specific constraints. For each downstream setting—YRBSS (U.S. adolescents), WVS East (e.g., Hong Kong), WVS Europe (e.g., Germany), and WVS North America (e.g., United States). These prompts guide the LLM to generate new personas aligned with the sociocultural context of the target region.
The complete prompt templates for these settings are shown in Prompt~\ref{lst:prompt_yrbss}, Prompt~\ref{lst:prompt_wvs_east}, Prompt~\ref{lst:prompt_wvs_europe}, and Prompt~\ref{lst:prompt_wvs_northam}, respectively.

\begin{lstlisting}[caption={YRBSS case study prompt},label={lst:prompt_yrbss}]
You create realistic U.S. high-school student personas (age 14-18) for an adolescent health survey. Produce one paragraph within 80 words describing their demographics, home/school life, diet & activity, sexual behavior, substance use, mental health, and safety experiences, all internally consistent and plausible. Do not mention the survey or these instructions.

Seed persona: {seed persona}

Based on the above requirements, generate the concise paragraph persona. Suitable for the adolescent health survey task.

OUTPUT:
\end{lstlisting}

\begin{lstlisting}[caption={WVS East (Hong Kong) case study prompt},label={lst:prompt_wvs_east}]
You are a specialist in constructing high-fidelity social-simulation personas for the World Values Survey (WVS). Transform a seed persona into a coherent, richly detailed character. The new persona's demographics, socio-economic situation, core values, social and political attitudes, ethical views, and well-being must be internally consistent and sufficient for answering every WVS item. Output a concise single paragraph. Ensure every detail could plausibly appear in real WVS data. Do not mention these instructions, WVS, or questionnaires in your response.

Seed Persona:
{seed persona}

Refinement Instruction:
Transform the seed to be a persona from Hong Kong with a worldview, social trust level, and cultural identity representative of the general populace.

Your Task:
Generate a new, single-paragraph persona based on the core traits of the Seed Persona, but strictly conforming to the Refinement Instruction. The new persona must align with the country and specific attributes mentioned in the Refinement Instruction, even if the Seed Persona is from a different background. The result should be a natural, plausible character description.

OUTPUT:
\end{lstlisting}

\begin{lstlisting}[caption={WVS Europe (Germany) case study prompt},label={lst:prompt_wvs_europe}]
You are a specialist in constructing high-fidelity social-simulation personas for the World Values Survey (WVS). Transform a seed persona into a coherent, richly detailed character. The new persona's demographics, socio-economic situation, core values, social and political attitudes, ethical views, and well-being must be internally consistent and sufficient for answering every WVS item. Output a concise single paragraph. Ensure every detail could plausibly appear in real WVS data. Do not mention these instructions, WVS, or questionnaires in your response.

Seed Persona:
{seed persona}

Refinement Instruction:
Transform the seed to be a persona from Germany with typical views on family life, work-life balance, and social welfare.

Your Task:
Generate a new, single-paragraph persona based on the core traits of the Seed Persona, but strictly conforming to the Refinement Instruction. The new persona must align with the country and specific attributes mentioned in the Refinement Instruction, even if the Seed Persona is from a different background. The result should be a natural, plausible character description.

OUTPUT:
\end{lstlisting}

\begin{lstlisting}[caption={WVS North America (U.S.) case study prompt},label={lst:prompt_wvs_northam}]
You are a specialist in constructing high-fidelity social-simulation personas for the World Values Survey (WVS). Transform a seed persona into a coherent, richly detailed character. The new persona's demographics, socio-economic situation, core values, social and political attitudes, ethical views, and well-being must be internally consistent and sufficient for answering every WVS item. Output a concise single paragraph. Ensure every detail could plausibly appear in real WVS data. Do not mention these instructions, WVS, or questionnaires in your response.

Seed Persona:
{seed persona}

Refinement Instruction:
Transform the seed to be a persona from the United States with mainstream perspectives on national identity, civic engagement, and social justice.

Your Task:
Generate a new, single-paragraph persona based on the core traits of the Seed Persona, but strictly conforming to the Refinement Instruction. The new persona must align with the country and specific attributes mentioned in the Refinement Instruction, even if the Seed Persona is from a different background. The result should be a natural, plausible character description.

OUTPUT:
\end{lstlisting}

\section{Evaluation Metrics: Definitions and Rationale}
\label{app:metrics}

In this appendix, we provide the rationale and formal definitions for the evaluation metrics used in Section~\ref{sec:Exp_setup}, covering population-level alignment, individual-level consistency, and downstream social-science simulation.

\subsection{Metric Selection Rationale}
Our metric suite is designed to evaluate complementary notions of fidelity in LLM-based social simulation. 
For population-level alignment, the central question is whether the distribution induced by a persona set matches the distribution of real human responses, rather than whether individual samples merely appear plausible. 
We therefore adopt a diverse set of distribution-level discrepancies that capture different aspects of mismatch between human and synthetic populations. 
AMW provides an interpretable marginal comparison by measuring trait-wise Wasserstein distance and averaging across traits, making it sensitive to shifts in each response dimension~\citep{wang2024statistical}. 
SW extends this idea to high-dimensional response vectors by comparing one-dimensional projections along multiple random directions, thereby capturing broader geometric differences in the joint distribution~\citep{nguyen2022revisiting,lobashev2025color}. 
FD summarizes first- and second-order discrepancies through a Gaussian approximation, offering a compact measure of mean and covariance mismatch~\citep{veeramacheneni2023fr,jayasumana2024rethinking}. 
MMD measures the distance between kernel mean embeddings~\citep{kalinke2022maximum,ren2025few}. For WVS and YRBSS, ED compares between-population and within-population Euclidean distances. 
Using these metrics together reduces dependence on any single assumption and yields a more robust assessment of population-level alignment.
For individual-level consistency, matching marginal or joint distributions alone does not guarantee that synthetic respondents preserve realistic relationships among traits. 
We therefore compare inter-trait Pearson correlation structures between human and synthetic responses using $\mathrm{MAE}_{\mathrm{corr}}$. 
This metric evaluates whether the simulated population preserves trait co-variation patterns that are important for downstream social simulation, where behaviors often depend on combinations of correlated attributes rather than isolated traits.
For the SocSci benchmark, we report Accuracy Score and Distribution Score following the benchmark's original protocol to ensure comparability with prior work~\citep{Kolluri2025FinetuningLF}. 
The Accuracy Score measures response-level agreement with observed human answers, while the Distribution Score measures condition-level alignment between predicted and human response distributions. 
Together, these two metrics evaluate both individual prediction fidelity and aggregate experimental fidelity in downstream social-science simulations.

\subsection{Population-level Alignment Metrics}
We compare LLM-generated and human response distributions using AMW, FD, SW, and either MMD (psychometric instruments) or ED (WVS and YRBSS):

\noindent  \textbf{Averaged Monotonic Wasserstein (AMW)}~\citep{villani2008optimal,wang2024statistical} measures 1D Wasserstein-1 distance across each trait and averages over all \(N\) traits:
\begin{equation}
\scalebox{0.75}{$\displaystyle
\mathrm{AMW} = \frac{1}{N} \sum_{t=1}^{N} \int_{-\infty}^{\infty} \bigl|F_{P_t}(x) - F_{Q_t}(x)\bigr| \, dx,
$}
\end{equation}
where \(F_{P_t}\), \(F_{Q_t}\) are empirical CDFs of human and synthetic responses on trait \(t\).

\noindent  \textbf{Fréchet Distance (FD)}~\citep{veeramacheneni2023fr,jayasumana2024rethinking} compares two multivariate Gaussian distributions:
\begin{equation}
\scalebox{0.75}{$\displaystyle
\mathrm{FD} = \|\mu_P - \mu_Q\|_2^2 + \mathrm{Tr}(\Sigma_P + \Sigma_Q - 2(\Sigma_P \Sigma_Q)^{1/2}),
$}
\end{equation}
where \(\mu_P, \mu_Q\) are mean vectors and \(\Sigma_P, \Sigma_Q\) are covariance matrices of human and synthetic samples.

\noindent  \textbf{Sliced Wasserstein Distance (SW)}~\citep{nguyen2022revisiting,lobashev2025color}  projects high-dimensional samples onto \(K\) random directions \(v_i\), then averages 1D Wasserstein distances:
\begin{equation}
\mathrm{SW} = \frac{1}{K} \sum_{i=1}^{K} W_1(\langle X, v_i \rangle, \langle Y, v_i \rangle),
\end{equation}
where \(\langle \cdot, v_i \rangle\) denotes projection.

\noindent  \textbf{Maximum Mean Discrepancy (MMD)}~\citep{kalinke2022maximum, ren2025few} measures distributional gap in a reproducing kernel Hilbert space (RKHS):
\begin{equation}
\mathrm{MMD} = \left\| \mathbb{E}_{x \sim P}[\phi(x)] - \mathbb{E}_{y \sim Q}[\phi(y)] \right\|_{\mathcal{H}},
\end{equation}
where \(\phi\) is the kernel mapping and \(\mathcal{H}\) is the RKHS.

\noindent \textbf{Energy Distance (ED)} is used for WVS and YRBSS. For independent draws $X,X'\sim P$ and $Y,Y'\sim Q$, define
\begin{equation}
\mathcal{E}(P,Q)
=2\mathbb{E}\|X-Y\|_2-\mathbb{E}\|X-X'\|_2-\mathbb{E}\|Y-Y'\|_2.
\end{equation}
We report $\mathcal{E}(P,Q)$ for WVS and $\sqrt{\mathcal{E}(P,Q)}$ for YRBSS, estimated from pairwise Euclidean distances.

\subsection{Individual-level Consistency Metric}
To evaluate whether inter-trait correlations are preserved, we compute Pearson correlation coefficients across all \(N\) trait pairs from both human and synthetic responses, and report their mean absolute error:
\begin{equation}
\mathrm{MAE}_{\text{corr}} = \frac{1}{N} \sum_{i=1}^{N} \left| \hat{\rho}_i - \rho_i^{\text{human}} \right|,
\end{equation}
where \( \rho_i^{\text{human}} \) and \( \hat{\rho}_i \) denote the Pearson correlation coefficient~\citep{pearson_wikipedia,pratelli2025evaluating} for the \(i\)-th trait pair in human and synthetic data, respectively.

\subsection{SocSci Benchmark Metrics}

\paragraph{Accuracy Score.}
For each evaluated instance \(m=1,\dots,M\), let \(\hat r_m\) be the model-predicted response and \(r_m\) the ground-truth human response. Let \(r_{\min}(m)\) and \(r_{\max}(m)\) be the minimum/maximum values of the response scale for that question. We define
\begin{equation}
\scalebox{0.82}{$\displaystyle
\mathrm{Acc}=1-\frac{1}{M}\sum_{m=1}^{M}\frac{\lvert \hat r_m-r_m\rvert}{r_{\max}(m)-r_{\min}(m)},
$}
\end{equation}
where larger \(\mathrm{Acc}\) indicates better prediction.

\paragraph{Distribution Score.}
For each evaluated condition, outcome pair \(s\in\mathcal{S}\), let \(\hat R_s\) and \(R_s\) denote the multisets of model-predicted and human responses, respectively. Using the response scale \([r_{\min}(s), r_{\max}(s)]\), we standardize \(x'=(x-r_{\min}(s))/(r_{\max}(s)-r_{\min}(s))\), yielding \(\hat R'_s\) and \(R'_s\). We compute Wasserstein-1 distance \(W_1(\hat R'_s, R'_s)\) and average across all \(|\mathcal{S}|\) pairs:
\begin{equation}
\scalebox{0.7}{$\displaystyle
\mathrm{Dist}=\frac{1}{|\mathcal{S}|}\sum_{s\in\mathcal{S}} W_1\!\Bigl(\hat R'_s, R'_s\Bigr),
\quad
x'=\frac{x-r_{\min}(s)}{r_{\max}(s)-r_{\min}(s)},
$}
\end{equation}
where smaller \(\mathrm{Dist}\) indicates better distributional alignment.

\section{Reference Population and Representativeness}
\label{app:why_ipip}

We assess representativeness through population-level personality response distributions.
IPIP provides a standardized, public-domain Big Five reference with more than one million respondents across 223 countries/regions~\citep{ipip50,goldberg1992development,openpsychometrics2019}.
The reference consists of English-language responses from self-selected online participants.

IPIP measures in-domain alignment; CFCS, FBPS, Duckworth, WVS, YRBSS, and SocSci evaluate transfer to other psychological constructs, survey responses, and behavioral outcomes.
Appendix~\ref{app:cross_instrument_reference} compares alternative alignment instruments, and Appendix~\ref{app:response_robustness} examines internal response structure and prompt variation.

\section{Cross-Instrument Reference Ablation}
\label{app:cross_instrument_reference}

Our global alignment uses IPIP Big Five responses as the reference distribution. 
We choose IPIP not because Big Five traits fully determine all downstream behaviors, but because it provides a widely used basic-trait representation and the largest public psychometric reference in our setting, covering more than one million respondents from 223 countries/regions~\citep{ipip50,goldberg1992development,openpsychometrics2019}. 
Still, matching only in the IPIP space may raise concerns about transfer to other constructs. 
We therefore rerun the same two-stage alignment procedure using three smaller instruments, CFCS, FBPS, and Duckworth, as alternative reference distributions, and evaluate each aligned set on held-out instruments.
As shown in Table~\ref{tab:cross_reference_alignment}, IPIP-based alignment performs best on all held-out instruments and metrics, outperforming alignments based on the smaller alternative questionnaires. 
This suggests that the broad coverage of the IPIP reference provides a more stable global population prior than narrower instruments. 
The smaller gains from {Raw} to {Resample} in the out-of-domain setting are therefore expected: CFCS, FBPS, and Duckworth measure constructs related to, but not identical with, Big Five traits. 
Nevertheless, the consistent cross-instrument improvements indicate that IPIP alignment does not merely overfit to IPIP items, but improves transferable population-level coverage. 
When task-specific reference data are available, our group-specific adaptation module can further incorporate such target constraints beyond this global prior.

\begin{table*}[h]
\caption{
Cross-instrument reference ablation with different alignment references. 
Each block evaluates transfer to a held-out psychometric instrument. 
Lower is better for all metrics.
}
\centering
\scriptsize
\setlength{\tabcolsep}{5pt}
\begin{tabular}{@{}llcccc@{}}
\toprule
\textbf{Test} & \textbf{Reference} & \textbf{AMW} & \textbf{FD} & \textbf{SW} & \textbf{MMD} \\
\midrule
\multirow{3}{*}{CFCS}
& FBPS      & 0.1976 & 0.0685 & 0.1974 & 0.0715 \\
& Duckworth & 0.1990 & 0.0725 & 0.1990 & 0.0713 \\
& IPIP      & 0.1954 & 0.0614 & 0.1726 & 0.0642 \\
\midrule
\multirow{3}{*}{FBPS}
& CFCS      & 0.2340 & 0.4653 & 0.2080 & 0.2400 \\
& Duckworth & 0.2442 & 0.4750 & 0.2050 & 0.2550 \\
& IPIP      & 0.2223 & 0.4220 & 0.1900 & 0.2186 \\
\midrule
\multirow{3}{*}{Duckworth}
& CFCS      & 0.2531 & 0.2363 & 0.4330 & 0.2727 \\
& FBPS      & 0.2487 & 0.2354 & 0.4353 & 0.2735 \\
& IPIP      & 0.2450 & 0.2324 & 0.4257 & 0.2623 \\
\bottomrule
\end{tabular}
\vspace{-1mm}
\label{tab:cross_reference_alignment}
\end{table*}

\section{Distributional Alignment and Downstream Outcomes}
\label{app:downstream_utility}

For an $L$-Lipschitz outcome function $f$, the difference between human and persona expectations satisfies
\[
\left|\mathbb{E}_{r_{\texttt{human}}}[f]-\mathbb{E}_{r_{\texttt{persona}}}[f]\right|
\leq L\,W_1(r_{\texttt{human}},r_{\texttt{persona}}).
\]
This bound motivates measuring population mismatch in policy and healthcare simulations.

In SocSci~\citep{Kolluri2025FinetuningLF}, each simulated response is conditioned on a persona, an experimental condition, and an outcome question, and is compared with observed human responses.
We hypothesize that grounded personas transfer to these tasks because they encode values, preferences, and behavioral tendencies from long-term human-authored text, while alignment improves the coverage of these characteristics.
Appendices~\ref{app:repeated_runs} and~\ref{app:external_sampling} compare random sampling and IS+OT across five runs.

\section{Bias in Group-Specific Persona Adaptation}
\label{app:group_bias}

The adaptation prompt instructs the model to preserve the retrieved seed persona's core traits while adapting its description to the target group (Appendix~\ref{app:group_prompts}).
We audit 210 seed--revised persona pairs, sampling 70 each from the Hong Kong, Germany, and United States WVS settings.
Four annotators independently assess trait preservation and unsupported or stereotypical cultural, political, religious, behavioral, and value-related additions on a 1--5 scale.
We average their scores and flag pairs below 4 as potential stereotype leakage.
GPT-5.5 evaluates the same pairs independently.

Table~\ref{tab:group_audit} reports potential leakage in 9/210 pairs (4.3\%) under human evaluation and 12/210 (5.7\%) under GPT-5.5 evaluation.

\begin{table}[htbp]
\centering
\caption{Stereotype-leakage audit of group-specific adaptation. \emph{No leakage} denotes a score of at least 4; \emph{Potential} denotes a score below 4.}
\label{tab:group_audit}
\small
\setlength{\tabcolsep}{6pt}
\begin{tabular}{@{}lcccc@{}}
\toprule
& \multicolumn{2}{c}{Human} & \multicolumn{2}{c}{GPT-5.5} \\
\cmidrule(lr){2-3}\cmidrule(l){4-5}
WVS setting & No leakage & Potential & No leakage & Potential \\
\midrule
Hong Kong & 68/70 & 2/70 & 66/70 & 4/70 \\
Germany & 67/70 & 3/70 & 65/70 & 5/70 \\
United States & 66/70 & 4/70 & 67/70 & 3/70 \\
\midrule
Overall & 201/210 & 9/210 & 198/210 & 12/210 \\
\bottomrule
\end{tabular}
\end{table}

\paragraph{Response-model control.}
On the WVS Japan subset, replacing Meta-Llama-3-70B with the Japanese-adapted Llama-3-Swallow-70B-Instruct-v0.1 under the same evaluation setup reduces AMW from 0.7804 to 0.7621 and FD from 19.5930 to 18.7672.

\begin{lstlisting}[caption={Group-adaptation evaluation prompt.},label={lst:audit_group_adaptation}]
Evaluate whether the Revised Persona preserves the Seed Persona without introducing unsupported or stereotypical group attributes. Consider:
1. Whether the core traits and characteristics of the Seed Persona are preserved.
2. Whether newly added group-specific details are relevant, natural, and plausible.
3. Whether unsupported demographic, cultural, political, religious, behavioral, or value-related attributes are introduced.
4. Whether the persona is overgeneralized or reduced to a group stereotype.
Score:
5 = Fully faithful; core traits are preserved and no stereotypical attributes are introduced.
4 = Mostly faithful; minor additions with no meaningful stereotype leakage.
3 = Some unsupported or overgeneralized group-related attributes are introduced.
2 = Clear stereotype leakage or substantial alteration of the seed persona.
1 = Severe stereotyping or replacement of the individual with a group prototype.
Scores 4-5 indicate no detectable stereotype leakage. Scores 1-3 indicate potential stereotype leakage.
Return only JSON:
{"score": 1, "no_detectable_leakage": false, "issues": "none or a brief description", "reasoning": "brief explanation"}
Target Group: {Target Group}
Seed Persona: {Seed Persona}
Revised Persona: {Revised Persona}
\end{lstlisting}

\section{Limitations}
\label{app:limi}

The seed corpora and IPIP reference reflect internet-accessible populations, with limited coverage of people who have little online presence.
IPIP additionally reflects English-language, self-selected participation, and Big Five captures one part of human diversity.
Narrative personas retain information expressed in source posts; attributes absent from those posts cannot be reliably recovered, which constrains group-specific adaptation.
Filtering harmful content can also underrepresent adversity and social tension.

LLM rewriting, generation, and adaptation can introduce unsupported attributes.
The audits in Appendices~\ref{app:curation_audit} and~\ref{app:group_bias} identify content shifts and stereotype leakage in a subset of the sampled pairs.

\section{Relation to SocioVerse and Seed Persona Ablation}
\label{app:socioverse_ablation}

SocioVerse~\citep{zhang2025socioverse} is closely related to our work in that both methods construct a candidate pool and then sample agents from it. 
The main differences lie in the agent representation, the alignment target, and the sampling objective.
SocioVerse represents each agent as an account-level unit with a user ID, associated posts, and a fixed set of inferred attributes:
\begin{equation}
U_i = (\mathrm{user\ ID}, P_i, \mathbf{z}_i),
\end{equation}
where $P_i$ denotes the user's posts and $\mathbf{z}_i$ contains demographic, socioeconomic, political, personality, and interest labels. 
Its sampling procedures operate over low-dimensional demographic or economic variables. 
For example, the U.S. election setting uses IPF to match marginals over gender, race, age group, ideology, and partisanship; the breaking-news setting samples from an observed joint distribution over a small number of demographic variables; and the China economic survey setting samples from region-level aggregates and a parametric income prior.

Our framework uses a different representation and alignment space. 
Each seed persona is a natural-language profile distilled from long-term behavioral traces and written for LLM role-play. 
Instead of aligning demographic marginals or aggregate statistics, we align persona-conditioned LLM responses in a continuous psychometric response space. 
Given a persona $p_i$ and inventory items $\mathcal{Q}$, the simulator produces a response vector $x_i \in \mathbb{R}^d$, which is directly matched to individual-level human response vectors in the same item space. 
This enables distribution-level alignment of simulated behavior rather than only demographic balancing.
Algorithmically, SocioVerse uses task-specific sampling rules such as IPF, identical distribution sampling, and parametric mixture sampling. 
These methods are appropriate for matching discrete marginals, low-dimensional joints, or aggregate moments. 
In contrast, our two-stage sampler first uses KDE-based importance-based selection to select candidates with high overlap with the human response distribution, and then applies optimal transport for fine-grained global matching in the continuous response space. 
Additional empirical comparisons with IPF and IDS are provided in Appendix~\ref{app:sampling_baselines}.
We also ablate two components of the seed persona mining pipeline under the same IPIP Big Five setting with Qwen2.5-72B used in Table~\ref{tab:population-level-in-domain}. 
As shown in Table~\ref{tab:seed_persona_ablation}, both text rewriting and persona quality control improve the quality of the initial seed pool before distributional alignment.
Removing text rewriting causes the largest degradation, indicating that cleaning noisy social-media text is important for stable persona extraction. 
Removing persona quality control also worsens alignment, suggesting that source-grounded persona generation can still produce generic or weakly grounded profiles. 
These results show that effective population alignment depends not only on the final sampler, but also on constructing a reliable seed persona pool.
\begin{table}[t]
\centering
\caption{Ablation study of the seed persona mining pipeline on IPIP Big Five with Qwen2.5-72B. Lower is better for both metrics.}
\label{tab:seed_persona_ablation}
\scriptsize
\setlength{\tabcolsep}{6pt}
\renewcommand{\arraystretch}{1.08}
\begin{tabular*}{\columnwidth}{@{\extracolsep{\fill}}lcc@{}}
\toprule
\textbf{Method} & \textbf{AMW} & \textbf{FD} \\
\midrule
Full pipeline & \textbf{0.2215} & \textbf{0.4630} \\
w/o text rewriting & 0.3231 & 0.7834 \\
w/o persona quality filter & 0.2673 & 0.5726 \\
\bottomrule
\end{tabular*}
\vspace{-1mm}
\end{table}

\section{Comparison with Sampling Baselines}
\label{app:sampling_baselines}

We further compare our two-stage sampler with two sampling strategies used in SocioVerse~\citep{zhang2025socioverse}: iterative proportional fitting (IPF) and identical distribution sampling (IDS). 
All methods are evaluated under the same Qwen2.5-72B in-domain setting as Table~\ref{tab:population-level-in-domain}, using IPIP Big Five responses as the human reference distribution. 
We report AMW and FD, where lower values indicate better population-level alignment.

\begin{table}[t]
\caption{Comparison with SocioVerse sampling baselines on IPIP Big Five under the Qwen2.5-72B setting.}
\centering
\scriptsize
\setlength{\tabcolsep}{8pt}
\begin{tabular}{@{}lcc@{}}
\toprule
\textbf{Sampler} & \textbf{AMW} & \textbf{FD} \\
\midrule
IPF~\citep{zhang2025socioverse} & 0.2063 & 0.3675 \\
IDS~\citep{zhang2025socioverse} & 0.1985 & 0.3446 \\
Ours (IS)    & 0.1834 & 0.3241 \\
Ours (IS+OT) & \textbf{0.1619} & \textbf{0.2858} \\
\bottomrule
\end{tabular}
\vspace{-1mm}
\label{tab:sampling_baselines}
\end{table}

As shown in Table~\ref{tab:sampling_baselines}, our IS-only variant already outperforms both IPF and IDS, indicating that density-ratio-based selection provides a stronger first-stage alignment signal for the persona response space. 
Adding the OT stage further improves both AMW and FD, showing that fine-grained transport-based matching complements the coarse distributional correction from IS. 
Since applying OT directly to the full seed pool would require constructing a large pairwise cost matrix between all candidate personas and human responses, our two-stage design uses IS to first reduce the candidate space and then applies OT for more precise alignment.

\section{Aligned Pool Size Sensitivity}
\label{app:seed_pool_size}

The 136M raw texts are cleaned, aggregated by author, and filtered into a pool of over 160K personas.
For group-specific evaluation, we vary the aligned pool size $N'$ over 1K, 5K, 10K, and 80K personas, using 5K by default and keeping the remaining pipeline fixed.
Table~\ref{tab:seedpool_size_amw} reports AMW on YRBSS and the WVS subsets.
Increasing the pool from 1K to 5K reduces YRBSS AMW from 0.433 to 0.290; the 10K and 80K pools yield similar scores of 0.287 and 0.285.

For importance-based selection, we retain the $N^{\dagger}=\lfloor 0.7N\rfloor$ candidates with the highest density-ratio weights.
Retention ratios from 55\% to 95\% give similar results in Table~\ref{tab:sensitivity-amw}.
Table~\ref{tab:seed_persona_ablation} reports the filtering ablation.

\begin{table}[t]
\caption{Aligned pool size sensitivity on downstream subgroup alignment (AMW; lower is better).
}
\centering
\scriptsize
\setlength{\tabcolsep}{4.2pt}

\begin{tabular*}{\columnwidth}{@{\extracolsep{\fill}} lcccc}
\toprule
\textbf{Setting} & $N'{=}1\mathrm{k}$ & $N'{=}5\mathrm{k}$ & $N'{=}10\mathrm{k}$ & $N'{=}80\mathrm{k}$ \\
\midrule
\multicolumn{5}{@{}l}{\textbf{Aligned pool size $N'$ (default: $5\mathrm{k}$)}} \\
\midrule
\textbf{YRBSS} & 0.433 & 0.290 & 0.287 & 0.285 \\
\textbf{WVS (East Asia)} & 0.896 & 0.784 & 0.782 & 0.781 \\
\textbf{WVS (North America)} & 0.822 & 0.774 & 0.771 & 0.770 \\
\textbf{WVS (Western Europe)} & 0.878 & 0.790 & 0.786 & 0.783 \\
\bottomrule
\end{tabular*}

\vspace{-2mm}
\label{tab:seedpool_size_amw}
\end{table}

\section{Rationale for Questionnaire-based Evaluation}
\label{app:why_questionnaires}

We choose psychometric and survey-style benchmarks because they provide standardized prompts, controlled response scales, and large human reference distributions. 
These properties are well suited to the core goal of this work: testing whether a persona set matches real population-level variation. 
Given the same questionnaire items, synthetic persona responses can be directly compared with human responses using distributional metrics such as AMW, FD, SW, MMD, and ED, while avoiding confounders from environment design.
The selected benchmarks cover both broad personality traits and downstream social attributes. 
IPIP captures domain-general Big Five personality variation~\citep{ipip50,goldberg1992development,openpsychometrics2019}; CFCS, FBPS, and Duckworth test transfer to additional psychometric constructs~\citep{Strathman1994CFC,OpenPsychometrics2019Firstborn,Duckworth2007Grit}; and WVS and YRBSS provide survey-style population references for values, attitudes, and adolescent health behaviors~\citep{Inglehart2020WVS,Brener2024OverviewAM}. 
Together, these datasets allow controlled evaluation of whether persona alignment transfers across related but distinct human-response distributions.
Long-horizon social simulations are also important, but they introduce additional factors such as interaction rules, incentives, memory, goals, and environment dynamics. 
In such settings, it is harder to isolate whether a mismatch arises from the persona distribution itself or from the simulation framework. 
For this reason, questionnaire-based evaluation serves as a controlled testbed for population-level persona alignment. 
To further examine transfer beyond questionnaires, we additionally evaluate on SocSci~\citep{Kolluri2025FinetuningLF} and six behavioral tasks adopted from~\citep{xie2024can}; see Section~\ref{sec:SocSci} and Appendix~\ref{app:behavior_generalization}. 
These additional experiments test whether the benefits of population alignment extend beyond questionnaire responses to downstream social-science and behavioral simulations.

\section{Behavioral Generalization Across Contexts}
\label{app:behavior_generalization}

To further validate that population-aligned personas induce consistent behavior simulation, we additionally evaluate our personas on six behavioral tasks adopted from~\citep{xie2024can} using Llama3-8B as the simulator. These tasks cover both financial decision-making settings (Trust, Dictator, MAP Trust, Risky Dictator) and social-interaction settings (LPG, LGG). Following prior work, we report the deviation between the simulated outcomes and the corresponding human reference distributions (lower is better). As shown in Table~\ref{tab:behavior-generalization}, our method achieves the lowest deviation across all six tasks compared to representative existing persona sets, indicating that the distributional alignment obtained in our framework transfers to behavioral domains and yields more human-faithful population behavior across diverse contexts.

\begin{table}[t]
\caption{Behavioral generalization on six tasks (deviation from human reference; lower is better).}
\centering
\scriptsize
\setlength{\tabcolsep}{4.2pt}
\begin{tabular*}{\columnwidth}{@{\extracolsep{\fill}} lcccc}
\toprule
\textbf{Game} & \textbf{Ours} & \textbf{Nvidia} & \textbf{AlignX} & \textbf{Google} \\
\midrule
Trust          & 0.05 & 0.08 & 0.07 & 0.09 \\
Dictator       & 1.53 & 2.23 & 1.75 & 2.19 \\
MAP Trust      & 0.11 & 0.26 & 0.19 & 0.22 \\
Risky Dictator & 0.22 & 0.26 & 0.24 & 0.25 \\
LPG            & 0.12 & 0.24 & 0.27 & 0.26 \\
LGG            & 0.19 & 0.39 & 0.31 & 0.40 \\
\bottomrule
\end{tabular*}
\vspace{-2mm}
\label{tab:behavior-generalization}
\end{table}

\subsection{Multi-Turn Customer Behavior}
\label{app:customer_behavior}

We evaluate persona-conditioned simulation on the multi-turn customer behavior benchmark~\citep{lu2026multiturn}.
Table~\ref{tab:customer_behavior} reports Action Accuracy and Session Outcome F1.
Ours (Resample) achieves an Action Accuracy of 23.84 and a Session Outcome F1 of 56.61, the highest values among the evaluated methods.

\begin{table}[htbp]
\centering
\caption{Multi-turn customer behavior simulation. Higher is better.}
\label{tab:customer_behavior}
\small
\begin{tabular}{@{}lcc@{}}
\toprule
Persona source & Action Accuracy & Session Outcome F1 \\
\midrule
AlignX & 20.72 & 51.86 \\
Bavard & 22.08 & 54.73 \\
Google Synthetic & 21.91 & 54.25 \\
Nvidia Nemotron & 20.54 & 51.29 \\
Ours (Resample) & 23.84 & 56.61 \\
\bottomrule
\end{tabular}
\end{table}

\section{Privacy De-identification}
\label{sec:privacy}

Before Stage\,2, we insert a lightweight \emph{de-identification} step to ensure that no sensitive personal information (PII) is exposed when personas are inspected or released.  
For every persona in the {Ours (raw data)}  reported in Table~\ref{tab:population-level-in-domain}, we first run a GPT-4o PII detector.  
The detector prompt is shown in Listing~\ref{lst:detector_prompt}.  
The model considers as sensitive any full name, street-level home address, email address, phone number, or bank/credit-card number.  
Across the corpus, \textbf{99.31\%} of personas are flagged \texttt{no}, indicating that no PII is present.  
Listing~\ref{lst:yes_case} provides an example that triggers a \texttt{yes} response, while Listing~\ref{lst:no_case} illustrates a typical \texttt{no}.  
For every record flagged \texttt{yes}, we invoke a second prompt (Listing~\ref{lst:rewrite_prompt}) that rewrites the persona by replacing only the sensitive fields with realistic but fictitious values while preserving style and meaning.  
A de-identified output corresponding to the \texttt{yes} case is presented in Listing~\ref{lst:rewrite_example}.  
We further conducted a human study on all these rewritten \texttt{yes} responses, involving five independent annotators.  
All cases were verified to have the identified PII correctly removed or substituted, with no observed semantic inconsistencies, indicating the effectiveness of the de-identification process.

\begin{lstlisting}[caption={PII detection prompt},label={lst:detector_prompt}]
You are tasked with checking persona text for sensitive personal information.
Sensitive items are defined as follows: 
- Full names (at least first and last name together, not just a first name or a generic reference). 
- Home addresses (street-level detail, not just city or country). 
- Email addresses (must contain '@' with a valid-looking format). 
- Phone numbers (must contain at least 7 digits, optionally with separators like '-', '()', or spaces). 
- Bank account or credit card numbers (must contain a sequence of at least 8 digits, e.g., full or partial account numbers). 
Do NOT treat the following as sensitive: 
- Generic mentions of 'phone number', 'email', 'credit card', 'parents', etc. without specific values. 
- General locations (city, country, region). 
- Public figure names (e.g., celebrities, politicians, authors). 
Instructions: 
1. If you detect any concrete sensitive information as defined above, output only: yes. 
2. If none is found, output only: no. 
Output must be a single word: yes or no.


Persona text:

\end{lstlisting}

\begin{lstlisting}[caption={Detector output: yes},label={lst:yes_case}]
The user is a young, likely married, British mother in her mid-to-late twenties, living with her husband, Mr. Spid, and their young daughter, Baby Spid. She values her relationships and family, but also enjoys her independence, music, and socializing with friends, often expressing herself through sarcastic humor and frustration
\end{lstlisting}

\begin{lstlisting}[caption={Detector output: no},label={lst:no_case}]
The user is a young male, likely in a relationship, who values simplicity and frugality, but also enjoys spending time with his girlfriend and having fun, often finding joy in everyday moments, with a carefree and lighthearted tone, living in the Lower Mainland, possibly in Canada, given the mention of T&T and Tim Horton's.
\end{lstlisting}

\begin{lstlisting}[caption={PII rewrite prompt},label={lst:rewrite_prompt}]
You are tasked with de-identifying the given persona text by replacing sensitive items
with realistic but fictitious substitutes.
Sensitive items to REPLACE: full names, home addresses, email addresses, phone numbers,
and bank account numbers.
Do NOT replace nationality or general location (city, country, region).
Requirements:
+ Preserve the original style, tone, and intent.
+ Do not leak or partially reveal the original sensitive values.
+ Generate substitutes that look plausible but are fabricated and not traceable to real individuals.
\end{lstlisting}

\begin{lstlisting}[caption={Rewritten persona (de-identified)},label={lst:rewrite_example}]
The user is a young, likely married, British mother in her mid-to-late twenties, living with her husband, Tom Ridley, and their young daughter, Lila Ridley. She values her relationships and family, but also enjoys her independence, music, and socializing with friends, often expressing herself through sarcastic humor and frustration.
\end{lstlisting}

\subsection{Data and Model Release}
\label{app:data_release}

We will release the trained embedding model, source-data download links, and code for reproducing the processed dataset.
We have consulted our institution's research-compliance committee and initiated IRB review of the persona-derived datasets.
The code and embedding model can be released independently of persona data; dataset-derived artifacts will follow the corresponding institutional determinations.
The Blog Authorship Corpus, Reddit, and Twitter sources are reviewed separately, with additional procedures for the latter two.

\section{Seed Persona Mining: Data and Prompts}
\label{app:dataproc}

\noindent\textbf{Corpora and Scale.}
We use the Reddit, Twitter, and Blogger subsets of HumanLLM's Cognitive Genome Dataset~\citep{Lei2026HumanLLMTP} and follow its data processing pipeline. Specifically, we use: (1) the \emph{Blog Authorship Corpus}, which contains 681K posts and roughly 140 million words from over 19,000 bloggers; (2) Reddit submissions drawn from 50 topical subreddits covering lifestyle, identity, health, and culture, totaling 47.2 million posts; and (3) Twitter data comprising 88.1 million original tweets sourced from CommonCrawl archives. Collectively, these corpora capture diverse first-person perspectives spanning various demographics, interests, and socioemotional expressions.

\noindent\textbf{Preprocessing.}
We apply source-specific filters to remove noise while preserving personal expression. For Reddit, we discard posts authored by bots, deleted accounts, or system users. We retain only posts with engagement indicators ($\text{score}\ge2$, $\text{num\_comments}\ge2$) and content length exceeding 300 characters. For blogs, we remove segments shorter than 30 tokens or lacking first-person pronouns, which tend to be non-narrative. For Twitter, we exclude retweets, posts under 300 characters, and those authored by suspected corporate or promotional accounts (identified heuristically). The remaining texts are passed to \emph{Llama-3.3-70B-Instruct} with a specialized cleaning prompt (Prompt~\ref{lst:prompt_rewrite}) that removes HTML tags, URLs, non-English snippets, mentions, and other noise while preserving tone and syntax. This step yields approximately 4.0M Reddit posts, 500K blog entries, and 1.7M tweets.

\noindent\textbf{Raw Data Quality Control.}
Each cleaned text is further annotated using \emph{Llama-3.3-70B-Instruct} via Prompt~\ref{lst:prompt_quality}, which assigns: (1) a three-level \texttt{<quality>} tag—\emph{high}, \emph{medium}, or \emph{low}, reflecting narrative depth and relevance to human thought or behavior; and (2) a binary \texttt{<harmless>} tag indicating whether the content avoids sensitive or unethical material (e.g., hate speech, violence, privacy violations). Only texts labeled both high quality and harmless are retained. This filtering process excludes approximately 30.0\% of Reddit data, 26.4\% of blogs, and 60.4\% of tweets, resulting in a curated corpus of 2.8M Reddit posts, 368K blog entries, and 673K tweets.

\noindent\textbf{Persona Generation.}
For each retained user, we aggregate their posts and remove authors with fewer than six entries. The remaining user-level documents are concatenated into a single segment $s$ and passed to \emph{Llama-3.3-70B-Instruct} using Prompt~\ref{lst:prompt_persona}, which extracts a narrative persona paragraph. The persona is written in third person and integrates salient demographic, psychological, and behavioral traits, but only if they are explicitly supported by the source text. If the input fails to support persona synthesis, the model outputs \texttt{NULL} to avoid hallucination.

\noindent\textbf{Persona Quality Control.}
Each generated persona is scored by a critic LLM using the evaluation rubric in Prompt~\ref{lst:prompt_critic}. Five dimensions are rated on a 1-10 scale: \emph{hallucination} (factual grounding), \emph{coverage} (extent of salient information captured), \emph{conciseness and clarity}, \emph{relevance} (exclusion of generic/promotional content), and an \emph{overall} score. We retain only those personas where the overall score exceeds 8, and each sub-score exceeds 7. This step filters out 26.4\% of Reddit personas, 26.1\% from blogs, and 28.2\% from Twitter.

\noindent\textbf{Final Seed Pool.}
The resulting seed persona pool comprises over 160,000 high-fidelity, diverse narrative personas, spanning a broad spectrum of demographic dimensions. These personas serve as the foundation for downstream alignment and targeted generation modules.

\noindent\textbf{Prompts Detail.}
To ensure reproducibility, we include the full text of all LLM prompts used during the pipeline—cleaning (Prompt~\ref{lst:prompt_rewrite}), quality control (Prompt~\ref{lst:prompt_quality}), persona generation (Prompt~\ref{lst:prompt_persona}), and persona evaluation (Prompt~\ref{lst:prompt_critic}).

\begin{lstlisting}[caption={Prompt A: rewrite},label={lst:prompt_rewrite}]
You are a text cleaning assistant.
Your task is to clean a Reddit blog post by removing noisy or irrelevant elements while fully preserving the original content's meaning, sentence structure, punctuation, and tone.

Apply the following cleaning rules:
1. Remove truly extraneous characters and encoded text, including:
    - Unicode symbols like \u00a3, \u2026, \u0caa\u0ccd, etc.
    - HTML entities like &amp;, &gt;, #x200B, etc.
    - Repetitive filler patterns like . . ., \n.\n.\n, or similar.
2. Delete all URLs, including both raw links (https://..., http://...) and markdown-style links ([text](url)).
3. Remove all hashtags (e.g., #LifeGoals, #AI2024) unless they are clearly essential to the post's meaning.
4. Remove all user mentions (e.g., @elonmusk, @user123). If the mention is critical for understanding or its removal breaks the sentence structure, substitute an appropriate placeholder (e.g., "a friend," "the author") based on context.
5. Strip out emojis, meme-style text, ASCII art, and decorative symbols that do not contribute meaning.
6. Discard non-English text or phrases only if they are not integrated into natural English usage. Common loanwords or cultural terms (e.g., "sushi," "futbol") should be preserved if they are relevant.

Important Guidelines:
- Do not remove or alter meaningful punctuation such as parentheses (), dollar signs $, question marks ?, or exclamation marks ! if used naturally.
- Do not rephrase, paraphrase, or rewrite any part of the original.
- Maintain all meaningful sentence boundaries and user expression styles.
- If a removal creates broken syntax or unclear meaning, minimally repair the sentence to ensure readability without altering the author's voice.
- Output the cleaned text only-no explanations or notes.

Here is the Reddit blog post to clean:
{reddit}
\end{lstlisting}

\begin{lstlisting}[caption={Prompt B: raw data quality control},label={lst:prompt_quality}]
You are a helpful assistant that helps me determine the data quality of a blog. The background is that I want to collect blogs which contain human behaviors or human thoughts, so that I can further study social science based on the collected data in the next step. However, as you may know, blogs from the Internet contain various types of content, and many of them are irrelevant to my goal, so I need to filter them out.

Typically, a blog's quality is high if it records detailed events of a human, reflects human life, mentions social behaviors, or reveals the author's thoughts or feelings about something.

A blog's quality is medium if it only briefly mentions some content related to human behaviors or thoughts, but lacks enough context to understand a complete story or perspective.

A blog's quality is low if it has nothing to do with human behaviors or thoughts, such as ads, job posts, company descriptions, fictional plots, random word dumps, and other irrelevant types. Additionally, a blog is also low quality if it is filled with non-English words, URLs, mentions (e.g., @user), hashtags, special characters (such as Unicode symbols or HTML entities), or emojis, which suggest noise or lack of meaningful personal expression.

In addition to tagging the quality, please also determine whether the blog is harmless. A blog is considered harmless if it does not contain unethical or sensitive content such as violence, pornography, or privacy violations. If such content exists, the harmless tag should be no; otherwise, it should be yes.

So your task is to tag the blog in two aspects:

A quality tag, which can be either "high", "medium", or "low".

A harmless tag, which can be either "yes" or "no".

Please output both tags in the following XML format, and do not include any other words or explanations:
<output><quality>...</quality><harmless>...</harmless></output>

Below is the user's blog post:
{reddit}
\end{lstlisting}

\begin{lstlisting}[caption={Prompt C: persona\_generation},label={lst:prompt_persona}]
Analyze the provided blog posts to create a concise and vivid user persona in a single cohesive paragraph (no more than 100 words). Seamlessly weave together important aspects of the user's persona, such as demographics (age, gender, profession, nationality, location, marital status), key personality traits, core values, interests, and emotional tone, only when these aspects are explicitly supported by the raw blogs. Extract persona based on the user's salient events or thoughts.  

Exclude generic, promotional, or repetitive content in the raw blogs (e.g., product ads, event schedules, technical knowledge re-post).  
Write in third-person (e.g., 'The user is a...'), avoiding lists or bullet points. If the blogs' content cannot support high-quality persona extraction, simply output "NULL".

Now the task begins. Below are the person's raw blogs. Please only output the result. Do not include any additional explanatory text.
{reddit}
\end{lstlisting}

\begin{lstlisting}[caption={Prompt D: persona\_quality\_control},label={lst:prompt_critic}]
**Evaluate the quality of a generated user persona based on the raw posts and the extracted persona paragraph provided. Consider the following aspects in your evaluation:**    

1. **Hallucination:** Does the persona strictly reflect information that is explicitly supported by the raw posts? Score 10 for no hallucinations (all facts directly mentioned), 8 if a few facts are very confidently inferred, down to 1 if most facts are neither mentioned nor plausibly inferred.  
2. **Coverage:** Does the persona cover the main salient aspects of the user that appear in the raw posts (e.g., demographics, values, interests, emotional tone)? Are important elements missing?  
3. **Conciseness and Clarity:** Is the persona concise, cohesive, and written in clear third-person prose within the 100-word limit? Does it avoid redundancy, listing, and repetition?  
4. **Relevance:** Has generic, promotional, or repetitive content from the raw posts been appropriately excluded from the persona?    
5. **Overall:** Give an overall score for the persona based on the above aspects and your holistic judgment.  

**Output your evaluation in the following XML-like format:**    

<data>  
  <explanation> [Your brief analysis and justification.] </explanation>  
  <hallucination> [Hallucination score: integer 1-10] </hallucination>  
  <coverage> [Coverage score: integer 1-10] </coverage>  
  <conciseness> [Conciseness and clarity score: integer 1-10] </conciseness>  
  <relevance> [Relevance score: integer 1-10] </relevance>  
  <overall> [Overall score: integer 1-10] </overall>  
</data>  

**Instructions:**    
Base all ratings and analysis strictly on the given raw posts and generated persona. Do not include any content or facts unsupported by the original posts. If the persona reports "NULL", give the 1 to all scores.  

Now the task begins.
\end{lstlisting}

\noindent\textbf{Data Statistics.}
Table~\ref{tab:appendix-stats} summarizes the data retained at each processing stage. ``Raw'', ``Preproc'', and ``Raw QC'' count texts before processing, after preprocessing, and after text quality control, respectively. ``Persona Gen'' and ``Persona QC'' correspond to the user counts after data synthesis and quality control in HumanLLM. The final seed pool contains 160,409 personas from Reddit, Twitter, and blogs.

\begin{table*}[h]
\centering
\caption{Seed-persona data statistics. Counts in Raw, Raw QC, Persona Gen, and Persona QC are from HumanLLM~\citep{Lei2026HumanLLMTP} (Table A1).\label{tab:appendix-stats}}
\small
\renewcommand{\arraystretch}{1.12}
\begin{tabular}{lrrrrr}
\toprule
\textbf{Source} & \textbf{Raw} & \textbf{Preproc (S1)} & \textbf{Raw QC (S2)} & \textbf{Persona Gen (S3)} & \textbf{Persona QC (S4)} \\
\midrule
Reddit  & 47,200,000 & 4,000,000 & 2,800,000 & 102,861 & 75,661 \\
Twitter & 88,100,000 & 1,700,000 &   673,000 &  75,243 & 54,010 \\
Blog    &   681,000  &   500,000 &   368,000 &  41,607 & 30,738 \\
\midrule
\textbf{Total} & 135,981,000 & 6,200,000 & 3,841,000 & 219,711 & 160,409 \\
\bottomrule
\end{tabular}
\end{table*}

\subsection{LLM-Assisted Curation and External Validation}
\label{app:llm_curation}

The cleaning prompt instructs the model to preserve source meaning, sentence structure, punctuation, and tone while removing formatting noise (Listing~\ref{lst:prompt_rewrite}).
The subsequent prompts filter low-information content, generate source-supported profiles, and score their faithfulness and coverage (Listings~\ref{lst:prompt_quality}--\ref{lst:prompt_critic}).
We evaluate the resulting personas against human-reference psychometric, survey, and behavioral datasets.

\subsection{Human Evaluation of Rewriting and Persona Generation}
\label{app:curation_audit}

We audit 210 original--rewritten post pairs and 210 source-post--persona pairs, sampling 70 pairs per stage from each of Reddit, blogs, and Twitter.
Four annotators independently score each pair from 1 to 5 based on source faithfulness, information coverage, and unsupported content.
We average their ratings and classify a pair as faithful when its mean score is at least 4.
GPT-5.5 independently evaluates the same pairs using the same rubric.
Table~\ref{tab:curation_audit} reports faithful rewriting in 207 of 210 cases and faithful persona generation in 208 of 210 cases under human evaluation.

\begin{table}[htbp]
\centering
\caption{Source-faithfulness audit. Each stage includes 210 pairs; scores below 4 indicate a potential content shift.}
\label{tab:curation_audit}
\small
\begin{tabular}{@{}llcc@{}}
\toprule
Stage & Evaluator & Faithful ($\geq4$) & Potential shift ($<4$) \\
\midrule
\multirow{2}{*}{Post rewriting}
& Human & 207/210 & 3/210 \\
& GPT-5.5 & 208/210 & 2/210 \\
\multirow{2}{*}{Persona generation}
& Human & 208/210 & 2/210 \\
& GPT-5.5 & 208/210 & 2/210 \\
\bottomrule
\end{tabular}
\end{table}

\paragraph{Evaluation prompts.}
The following prompts define the criteria used by the annotators and the independent LLM evaluator.

\begin{lstlisting}[caption={Post-rewriting evaluation prompt.},label={lst:audit_rewriting}]
Evaluate whether the Rewritten Text preserves the information in the Original Text without introducing unsupported content. Consider:
1. Whether the original facts, experiences, opinions, emotions, and behavioral information are preserved.
2. Whether unsupported demographic, psychological, behavioral, political, religious, or value-related attributes are introduced.
3. Whether meaningful information relevant to persona generation is removed or altered.
Removing URLs, HTML, mentions, emojis, or formatting noise is acceptable unless it changes meaningful content.
Score:
5 = Fully faithful; only irrelevant noise is removed.
4 = Mostly faithful; minor edits with no meaningful information change.
3 = Some meaningful information is omitted or altered.
2 = Clear information loss or unsupported content is introduced.
1 = Severe distortion or hallucination.
Scores 4-5 indicate faithful rewriting. Scores 1-3 indicate a potential rewriting-induced shift.
Return only JSON:
{"score": 1, "faithful": false, "issues": "none or a brief description", "reasoning": "brief explanation"}
Original Text: {Original Text}
Rewritten Text: {Rewritten Text}
\end{lstlisting}

\begin{lstlisting}[caption={Persona-generation evaluation prompt.},label={lst:audit_generation}]
Evaluate whether the Generated Persona faithfully represents the Source Posts without introducing unsupported content. Consider:
1. Whether the attributes and claims in the persona are supported by the source posts.
2. Whether important information about the user is preserved.
3. Whether unsupported, contradictory, or overly generic claims are introduced.
Score:
5 = Fully faithful; major claims are supported and important information is covered.
4 = Mostly faithful; minor omissions or imprecision, with no meaningful unsupported claims.
3 = Some important information is omitted or weakly supported.
2 = Clear hallucinations, contradictions, or substantial information loss.
1 = Severely unfaithful or unrelated to the source posts.
Scores 4-5 indicate a faithful persona. Scores 1-3 indicate a potential generation-induced shift.
Return only JSON:
{"score": 1, "faithful": false, "issues": "none or a brief description", "reasoning": "brief explanation"}
Source Posts: {Source Posts}
Generated Persona: {Generated Persona}
\end{lstlisting}

\section{Two-Stage Distribution Alignment: Details}
\label{asb:DetailOfTwoStage}

\noindent \textbf{Stage 1: Importance-based Selection (IS).}
In the first stage, our goal is to efficiently select a subset of personas whose response distributions significantly overlap with the human distribution. To achieve this, we estimate the probability densities of both the human response distribution \( r_{\texttt{human}} \) and the original persona distribution \( r_{\texttt{persona}} \) via Kernel Density Estimation (KDE). Specifically, employing a Gaussian kernel with bandwidth \( h > 0 \), we obtain the density estimates:
\begin{equation}
\begin{split}
\hat{r}_{\texttt{human}}(x) 
&= \frac{1}{M} \sum_{j=1}^{M} \mathcal{N}(x - y_j ; h^2 I), \\[6pt]
\hat{r}_{\texttt{persona}}(x)
&= \frac{1}{N} \sum_{i=1}^{N} \mathcal{N}(x - x_i ; h^2 I),
\end{split}
\end{equation}
where \( I \in \mathbb{R}^{d \times d} \) denotes the identity matrix, and \(\mathcal{N}(z; h^2 I)\) represents a multivariate Gaussian density centered at zero with covariance \( h^2 I \).
Using these KDE-based estimates, we calculate an importance weight for each persona response vector \( x_i \) as the ratio of the estimated densities: 
$w_i^{\mathrm{IS}} 
=
\frac{ \hat{r}_{\texttt{human}}(x_i) }
     { \hat{r}_{\texttt{persona}}(x_i) }.
$
Intuitively, the weight \( w_i^{\mathrm{IS}} \) quantifies the relative likelihood of observing the response vector \( x_i \) under the human response distribution compared to the original persona distribution. Therefore, personas with higher importance weights indicate a stronger alignment with human responses.
We rank personas by decreasing importance weight. Let $i_{(1)},\ldots,i_{(N)}$ satisfy $w_{i_{(1)}}^{\mathrm{IS}}\ge\cdots\ge w_{i_{(N)}}^{\mathrm{IS}}$. We retain the highest-weight 70\%:
\begin{equation}
N^{\dagger}=\lfloor 0.7N\rfloor,
\qquad
\mathcal{P}^{\dagger}=\{p_{i_{(k)}}:k=1,\ldots,N^{\dagger}\}.
\end{equation}
The corresponding response vectors form $X^{\dagger}=\{x_i:p_i\in\mathcal{P}^{\dagger}\}$.

\vspace{0.5em}

\noindent \textbf{Stage 2:  Optimal Transport (OT).}
While Importance-based Selection identifies a promising subset of personas, it does not guarantee fine-grained alignment with the detailed structure of the human response distribution. Therefore, we introduce a second stage leveraging entropic Optimal Transport to precisely match the selected personas to the human data at a granular level.

Recall that each persona response vector \( x_i^{\dagger} \in \mathbb{R}^d \) and human response vector \( y_j \in \mathbb{R}^d \) comprise responses to a set of \( d \)  prompts \( \mathcal{Q} = \{ q_k \}_{k=1}^{d} \). To explicitly account for the importance of individual questionnaire items during alignment, we introduce a set of weights \( \{ \omega_k \}_{k=1}^{d} \), with \( \omega_k > 0 \) denoting the importance of prompt \( q_k \). Thus, we construct the cost matrix \( C \in \mathbb{R}^{N^{\dagger} \times M} \) as a weighted sum of squared differences:
\begin{equation}
C_{ij} = \sum_{k=1}^{d} \omega_k \left(x_{ik}^{\dagger} - y_{jk}\right)^2.
\end{equation}

Our goal is then to identify an optimal probabilistic matching (transport plan) \( \Gamma \in \mathbb{R}^{N^{\dagger}\times M} \) between the two sets of vectors, such that the total weighted transport cost is minimized while maintaining uniform marginal distributions. 
We formulate the following entropic-regularized OT optimization problem:
\begin{equation}
\scalebox{0.80}{$\displaystyle
\Gamma^{\star} 
=
\arg\min_{\Gamma \in \Pi(\mathbf{a}^{\dagger}, \mathbf{b})}
\left\langle C, \Gamma \right\rangle 
+ \varepsilon \sum_{i,j} \Gamma_{ij} \left( \log \Gamma_{ij} - 1 \right),
$}
\end{equation}
where \(\varepsilon > 0\) is a regularization parameter that controls the entropy of the resulting transport plan. Each entry \(\Gamma_{ij} \in \mathbb{R}_{\ge 0}\) denotes the fraction of probability mass assigned to matching persona \(x_i^{\dagger}\) with human response \(y_j\), with higher values indicating closer alignment.

We define the marginals \(\mathbf{a}^{\dagger} \in \mathbb{R}^{N^{\dagger}}\) and \(\mathbf{b} \in \mathbb{R}^M\) as uniform distributions over the candidate personas and the real human samples, respectively:
\begin{equation}
\mathbf{a}^{\dagger} = \frac{1}{N^{\dagger}} \mathbf{1}_{N^{\dagger}}, 
\quad
\mathbf{b} = \frac{1}{M} \mathbf{1}_{M},
\end{equation}
where \(N^{\dagger}\) is the number of candidate personas after importance-based selection, and \(M\) is the number of empirical human response vectors from survey data. Here, \(\mathbf{1}_M \in \mathbb{R}^{M}\) and \(\mathbf{1}_{N^{\dagger}} \in \mathbb{R}^{N^{\dagger}}\) are vectors of all ones.
Given these marginals, the feasible set \(\Pi(\mathbf{a}^{\dagger}, \mathbf{b})\) enforces marginal constraints on the transport plan:
\begin{equation}
\scalebox{0.75}{$\displaystyle
\Pi(\mathbf{a}^{\dagger}, \mathbf{b})
=
\left\{
\Gamma \in \mathbb{R}_{\ge 0}^{N^{\dagger} \times M}
\;\middle|\;
\Gamma \mathbf{1}_M = \mathbf{a}^{\dagger}, \;
\Gamma^\top \mathbf{1}_{N^{\dagger}} = \mathbf{b}
\right\}.
$}
\end{equation}
The constraint \(\Gamma \mathbf{1}_M = \mathbf{a}^{\dagger}\) ensures that each persona distributes its probability mass across all human samples, while \(\Gamma^\top \mathbf{1}_{N^{\dagger}} = \mathbf{b}\) ensures that each human sample receives the correct total mass.

To solve the OT problem efficiently, we employ the Sinkhorn-Knopp iterative algorithm. Specifically, we first define the Gibbs kernel \( K \in \mathbb{R}^{N^{\dagger} \times M} \) by exponentiating the negative normalized cost matrix:
\begin{equation}
K_{ij} = \exp\left( - \frac{C_{ij}}{\varepsilon} \right),
\end{equation}
where each \( K_{ij} \) represents the unnormalized probability mass assigned to transporting from persona \( i \) to human \( j \), inversely weighted by the cost. Let \( u \in \mathbb{R}^{N^{\dagger}} \) and \( v \in \mathbb{R}^M \) be scaling vectors that normalize rows and columns of the transport plan, respectively. The iterative updates are:
\begin{equation}
u \leftarrow \frac{\mathbf{a}^{\dagger}}{K v}, \quad v \leftarrow \frac{\mathbf{b}}{K^{\top} u},
\end{equation}
where \( K v \) denotes matrix-vector multiplication, yielding a vector in \( \mathbb{R}^{N^{\dagger}} \) whose \( i \)-th element is \( \sum_j K_{ij} v_j \), and similarly \( K^{\top} u \in \mathbb{R}^M \) denotes the column-wise normalization step. These updates ensure that the resulting plan \(\Gamma \) satisfies the marginal constraints.
At convergence, we obtain the optimal transport plan \(\Gamma^{\star}\) as:
\begin{equation}
\Gamma^{\star} = \mathrm{diag}(u) \cdot K \cdot \mathrm{diag}(v),
\end{equation}
where \(\mathrm{diag}(u)\) and \(\mathrm{diag}(v)\) denote diagonal matrices with \(u\) and \(v\) on their diagonals, respectively.
Finally, we compute a cost-aware OT importance weight for each candidate persona \(x_i^{\dagger}\) by taking the expected transport cost under \(\Gamma^{\star}\) and mapping it through a temperature-controlled exponential:
\begin{equation}
\scalebox{0.75}{$\displaystyle
a_i^{\dagger} \;=\; \sum_{j=1}^{M} \Gamma_{ij}^{\star} \;=\; \frac{1}{N^{\dagger}}, 
\qquad
w_i^{\mathrm{OT}}
\;=\;
\exp\!\left(
-\frac{1}{\tau\, a_i^{\dagger}}
\sum_{j=1}^{M} \Gamma_{ij}^{\star}\, C_{ij}
\right),
$}
\end{equation}
where \(\tau>0\) is a temperature hyperparameter. After normalization, we sample the final aligned subset \( \mathcal{P}' \subseteq \mathcal{P}^{\dagger} \) of size \( N' \):
\begin{equation}
\scalebox{0.65}{$\displaystyle
\pi_i^{\mathrm{OT}} = \frac{w_i^{\mathrm{OT}}}{\sum_{k=1}^{N^{\dagger}} w_k^{\mathrm{OT}}}, 
\qquad
\mathcal{P}' \sim \mathrm{Multinomial}\!\left(N', \{\pi_i^{\mathrm{OT}}\}_{i=1}^{N^{\dagger}}\right).
$}
\end{equation}

\section{Repeated-Run Evaluation}
\label{app:repeated_runs}

We repeat the psychometric and SocSci evaluations with 1,000 personas per method and five random seeds (0--4), using the questionnaires and metrics described in Section~\ref{sec:Exp_setup}.
Tables~\ref{tab:repeated_fd}--\ref{tab:repeated_socsci} report means and standard deviations; all scores are multiplied by 100.
The $p$-values compare the size-matched RandomSelect and Resample variants.
Resample improves FD on IPIP, CFCS, and FBPS, and reduces $\mathrm{MAE}_{\mathrm{corr}}$ on all three out-of-domain datasets.
The reported OOD and SocSci differences satisfy $p<0.05$ except for Duckworth FD ($p=0.413$), where the two means are nearly identical.

\begin{table}[htbp]
\centering
\caption{FD across five runs ($\times 100$; lower is better). IPIP is the alignment reference; the other instruments are held out.}
\label{tab:repeated_fd}
\small
\setlength{\tabcolsep}{4pt}
\begin{tabular}{@{}lcccc@{}}
\toprule
Method & IPIP & CFCS & FBPS & Duckworth \\
\midrule
Tulu-3-Persona & $83.83\pm1.37$ & $8.33\pm0.67$ & $48.59\pm0.76$ & $40.01\pm0.59$ \\
Bavard & $78.26\pm0.94$ & $8.76\pm0.58$ & $68.01\pm0.98$ & $30.45\pm0.48$ \\
Google Synthetic & $80.68\pm1.04$ & $9.78\pm0.58$ & $71.26\pm1.06$ & $31.14\pm0.59$ \\
Nvidia Nemotron & $83.00\pm1.28$ & $10.34\pm0.43$ & $96.12\pm0.98$ & $40.45\pm0.62$ \\
\midrule
Ours (Raw) & $46.21\pm0.64$ & $7.02\pm0.20$ & $46.51\pm0.64$ & $23.26\pm0.32$ \\
Ours (RandomSelect) & $46.09\pm0.80$ & $6.99\pm0.25$ & $46.35\pm0.80$ & $23.19\pm0.40$ \\
Ours (Resample) & $28.50\pm0.59$ & $6.18\pm0.30$ & $42.32\pm0.87$ & $23.17\pm0.48$ \\
\bottomrule
\end{tabular}
\end{table}

\begin{table}[htbp]
\centering
\caption{OOD comparisons across five runs (scores $\times 100$; lower is better).}
\label{tab:repeated_ood_tests}
\small
\setlength{\tabcolsep}{6pt}
\begin{tabular}{@{}llccc@{}}
\toprule
Dataset & Metric & RandomSelect & Resample & $p$-value \\
\midrule
\multirow{3}{*}{CFCS}
& AMW & $19.69\pm0.27$ & $19.50\pm0.32$ & 0.0103 \\
& FD & $6.99\pm0.25$ & $6.18\pm0.30$ & 0.0018 \\
& $\mathrm{MAE}_{\mathrm{corr}}$ & $24.67\pm0.41$ & $22.94\pm0.46$ & 0.0003 \\
\midrule
\multirow{3}{*}{FBPS}
& AMW & $23.67\pm0.32$ & $22.18\pm0.36$ & 0.0001 \\
& FD & $46.35\pm0.80$ & $42.32\pm0.87$ & 0.0001 \\
& $\mathrm{MAE}_{\mathrm{corr}}$ & $38.87\pm0.58$ & $36.75\pm0.66$ & 0.0007 \\
\midrule
\multirow{3}{*}{Duckworth}
& AMW & $25.13\pm0.34$ & $24.56\pm0.40$ & 0.0121 \\
& FD & $23.19\pm0.40$ & $23.17\pm0.48$ & 0.4130 \\
& $\mathrm{MAE}_{\mathrm{corr}}$ & $44.91\pm0.71$ & $43.86\pm0.83$ & 0.0114 \\
\bottomrule
\end{tabular}
\end{table}

\begin{table}[htbp]
\centering
\caption{SocSci results across five runs (scores $\times 100$). The last row compares RandomSelect with Resample.}
\label{tab:repeated_socsci}
\small
\begin{tabular}{@{}lcc@{}}
\toprule
Method & Accuracy $\uparrow$ & Distribution $\downarrow$ \\
\midrule
Google Synthetic & $68.4\pm0.52$ & $23.6\pm0.39$ \\
Nvidia Nemotron & $69.5\pm0.61$ & $22.5\pm0.34$ \\
Ours (Raw) & $71.7\pm0.28$ & $21.2\pm0.19$ \\
Ours (RandomSelect) & $71.6\pm0.36$ & $21.2\pm0.24$ \\
Ours (Resample) & $72.3\pm0.31$ & $21.0\pm0.18$ \\
\midrule
$p$-value & 0.0001 & 0.0403 \\
\bottomrule
\end{tabular}
\end{table}

\FloatBarrier
\section{IS+OT on External Persona Collections}
\label{app:external_sampling}

We apply IS+OT to Bavard, Google Synthetic, and Nvidia Nemotron.
For each collection, we compare five random subsets of 1,000 personas with size-matched IS+OT subsets.
Tables~\ref{tab:external_ood}--\ref{tab:external_socsci} report mean $\pm$ standard deviation.
IS+OT reduces FD in every evaluated psychometric and WVS setting and improves both SocSci metrics.

\begin{table}[htbp]
\centering
\caption{IS+OT on external collections: held-out psychometric FD ($\times 100$; lower is better).}
\label{tab:external_ood}
\small
\setlength{\tabcolsep}{4pt}
\begin{tabular}{@{}llccc@{}}
\toprule
Persona collection & Selection & CFCS & FBPS & Duckworth \\
\midrule
\multirow{2}{*}{Bavard}
& Random & $8.76\pm0.58$ & $68.01\pm0.98$ & $30.45\pm0.48$ \\
& IS+OT & $7.64\pm0.62$ & $62.10\pm1.10$ & $27.42\pm0.52$ \\
\multirow{2}{*}{Google Synthetic}
& Random & $9.78\pm0.58$ & $71.26\pm1.06$ & $31.14\pm0.59$ \\
& IS+OT & $8.25\pm0.61$ & $65.06\pm1.15$ & $28.11\pm0.63$ \\
\multirow{2}{*}{Nvidia Nemotron}
& Random & $10.34\pm0.43$ & $96.12\pm0.98$ & $40.45\pm0.62$ \\
& IS+OT & $8.64\pm0.48$ & $87.76\pm1.12$ & $35.42\pm0.68$ \\
\bottomrule
\end{tabular}
\end{table}

\begin{table}[htbp]
\centering
\caption{IS+OT on external collections: WVS FD ($\times 100$; lower is better).}
\label{tab:external_wvs}
\small
\setlength{\tabcolsep}{3pt}
\begin{tabular}{@{}llccc@{}}
\toprule
Persona collection & Selection & East Asia & North America & Western Europe \\
\midrule
\multirow{2}{*}{Bavard}
& Random & $2269.17\pm24.13$ & $2033.71\pm21.17$ & $2490.76\pm31.00$ \\
& IS+OT & $2155.71\pm27.10$ & $1962.02\pm24.06$ & $2326.22\pm34.18$ \\
\multirow{2}{*}{Google Synthetic}
& Random & $2147.76\pm32.38$ & $2262.66\pm26.83$ & $2484.93\pm37.82$ \\
& IS+OT & $2040.37\pm35.02$ & $2149.53\pm29.14$ & $2320.68\pm40.31$ \\
\multirow{2}{*}{Nvidia Nemotron}
& Random & $2474.93\pm38.12$ & $2023.15\pm29.24$ & $2702.71\pm43.42$ \\
& IS+OT & $2221.18\pm41.05$ & $1921.99\pm32.06$ & $2447.57\pm46.10$ \\
\bottomrule
\end{tabular}
\end{table}

\begin{table}[htbp]
\centering
\caption{IS+OT on external collections: SocSci scores ($\times 100$).}
\label{tab:external_socsci}
\small
\begin{tabular}{@{}llcc@{}}
\toprule
Persona collection & Selection & Accuracy $\uparrow$ & Distribution $\downarrow$ \\
\midrule
\multirow{2}{*}{Bavard}
& Random & $68.5\pm0.46$ & $22.7\pm0.31$ \\
& IS+OT & $70.8\pm0.41$ & $21.8\pm0.27$ \\
\multirow{2}{*}{Google Synthetic}
& Random & $68.4\pm0.52$ & $23.6\pm0.39$ \\
& IS+OT & $70.3\pm0.34$ & $22.1\pm0.15$ \\
\multirow{2}{*}{Nvidia Nemotron}
& Random & $69.5\pm0.61$ & $22.5\pm0.34$ \\
& IS+OT & $71.2\pm0.45$ & $21.8\pm0.38$ \\
\bottomrule
\end{tabular}
\end{table}

\FloatBarrier
\section{Demographic Subgroup Evaluation}
\label{app:demographic_subgroups}

We evaluate male and female respondents separately in the WVS East Asia and North America subsets, using the group-specific setup in Section~\ref{subsec:group_specific}.
Table~\ref{tab:gender_subgroups} reports all four distributional metrics.
The trained embedding variant achieves the lowest error in each subgroup.
In North America, all methods have higher errors for female respondents than for male respondents.

\begin{table}[htbp]
\centering
\caption{Gender subgroup evaluation on WVS. ED denotes Energy Distance. Lower is better for all metrics.}
\label{tab:gender_subgroups}
\small
\setlength{\tabcolsep}{6pt}
\begin{tabular}{@{}lrrrr@{}}
\toprule
Method & AMW & FD & SW & ED \\
\midrule
\multicolumn{5}{@{}l}{\textbf{East Asia: Male}} \\
AlignX & 0.8957 & 23.0869 & 0.9428 & 2.6813 \\
Bavard & 0.8792 & 24.8543 & 0.8826 & 2.3175 \\
Google Synthetic & 0.8904 & 23.6874 & 0.9154 & 2.4814 \\
Nvidia Nemotron & 0.9635 & 27.4403 & 0.9802 & 2.9617 \\
PersonalHub & 0.8811 & 25.1228 & 0.8946 & 2.4056 \\
Ours (Raw) & 0.8513 & 22.1628 & 0.8578 & 2.1396 \\
Ours (EM w/o train) & 0.8473 & 21.8899 & 0.8571 & 2.2455 \\
Ours (EM w/ train) & 0.8124 & 20.2959 & 0.8174 & 2.0335 \\
\midrule
\multicolumn{5}{@{}l}{\textbf{East Asia: Female}} \\
AlignX & 0.8611 & 22.0011 & 0.9016 & 2.5417 \\
Bavard & 0.8520 & 23.7957 & 0.8514 & 2.2197 \\
Google Synthetic & 0.8500 & 22.3788 & 0.8722 & 2.3304 \\
Nvidia Nemotron & 0.9145 & 25.6449 & 0.9232 & 2.7575 \\
PersonalHub & 0.8481 & 23.8890 & 0.8546 & 2.2704 \\
Ours (Raw) & 0.8261 & 21.3118 & 0.8314 & 2.0550 \\
Ours (EM w/o train) & 0.8279 & 21.2749 & 0.8345 & 2.1769 \\
Ours (EM w/ train) & 0.7986 & 19.8789 & 0.8036 & 1.9861 \\
\midrule
\multicolumn{5}{@{}l}{\textbf{North America: Male}} \\
AlignX & 0.9537 & 22.3580 & 0.9609 & 2.6341 \\
Bavard & 0.9238 & 22.4006 & 0.9439 & 2.6245 \\
Google Synthetic & 0.9104 & 24.6749 & 0.9388 & 2.8703 \\
Nvidia Nemotron & 0.9349 & 21.9509 & 0.9329 & 2.6097 \\
PersonalHub & 0.8883 & 23.0671 & 0.9203 & 2.9838 \\
Ours (Raw) & 0.8701 & 21.0872 & 0.8975 & 2.4556 \\
Ours (EM w/o train) & 0.8518 & 21.3744 & 0.8855 & 2.4018 \\
Ours (EM w/ train) & 0.8042 & 18.3422 & 0.8151 & 1.8669 \\
\midrule
\multicolumn{5}{@{}l}{\textbf{North America: Female}} \\
AlignX & 0.9961 & 23.6918 & 1.0077 & 2.8023 \\
Bavard & 0.9602 & 23.4570 & 0.9841 & 2.7709 \\
Google Synthetic & 0.9566 & 26.3533 & 0.9950 & 3.0801 \\
Nvidia Nemotron & 0.9931 & 23.6889 & 0.9957 & 2.8411 \\
PersonalHub & 0.9325 & 24.5051 & 0.9681 & 3.1766 \\
Ours (Raw) & 0.8989 & 21.9700 & 0.9327 & 2.5732 \\
Ours (EM w/o train) & 0.8748 & 22.1392 & 0.9113 & 2.4952 \\
Ours (EM w/ train) & 0.8186 & 18.7884 & 0.8331 & 1.9145 \\
\bottomrule
\end{tabular}
\end{table}

\FloatBarrier
\section{Comparison with SYNTHIA}
\label{app:synthia}

We evaluate 1,000 personas from each of SYNTHIA~\citep{rahimzadeh2026synthia} and our aligned set using Qwen2.5-72B and the questionnaires and metrics in Section~\ref{sec:Exp_setup}.
Table~\ref{tab:synthia} reports MMD on IPIP and three held-out instruments.
Our aligned personas obtain lower MMD on all four datasets.

\begin{table}[htbp]
\centering
\caption{Comparison with SYNTHIA using Qwen2.5-72B. Lower MMD is better.}
\label{tab:synthia}
\small
\begin{tabular}{@{}lcccc@{}}
\toprule
Method & IPIP & CFCS & FBPS & Duckworth \\
\midrule
SYNTHIA & 0.3516 & 0.1168 & 0.3595 & 0.4372 \\
Ours & 0.0885 & 0.0642 & 0.2186 & 0.2623 \\
\bottomrule
\end{tabular}
\end{table}

\FloatBarrier
\section{Response Structure and Prompt Robustness}
\label{app:response_robustness}

\paragraph{Internal response structure.}
Following prior work~\citep{moon2024virtual}, we report the Frobenius norm and the absolute Cronbach's $\alpha$ gap from the human reference ($\alpha=0.8717$), using 1,000 personas per method and the IPIP protocol in Section~\ref{sec:Exp_setup}.
Our personas yield smaller discrepancies on both measures (Table~\ref{tab:response_structure}).

\begin{table}[htbp]
\centering
\caption{Internal response structure relative to humans. Lower is better.}
\label{tab:response_structure}
\small
\begin{tabular}{@{}lcc@{}}
\toprule
Method & Frobenius norm & $\alpha$ gap \\
\midrule
AlignX & 11.98 & 0.0852 \\
Bavard & 12.35 & 0.0879 \\
Google Synthetic & 12.53 & 0.0891 \\
Ours & 9.80 & 0.0697 \\
\bottomrule
\end{tabular}
\end{table}

\paragraph{Trait-pair visualization.}
EXT and EST in Figure~\ref{fig:scatter_ext_est_image} were selected at random from the five Big Five traits.

\paragraph{Prompt variation.}
We sample 1,000 personas from each of Google Synthetic, Nvidia Nemotron, and Ours (Resample), and repeat IPIP evaluation with five seeds (0--4).
We compare the original protocol, a semantically equivalent prompt, and reversed option ordering.
Table~\ref{tab:prompt_robustness} shows that the method ranking is unchanged across these conditions.

\begin{table}[htbp]
\centering
\caption{IPIP FD under prompt variation (mean $\pm$ standard deviation, $\times 100$; lower is better).}
\label{tab:prompt_robustness}
\small
\begin{tabular}{@{}lccc@{}}
\toprule
Method & Original & Paraphrased & Reversed \\
\midrule
Google Synthetic & $80.68\pm1.04$ & $81.35\pm1.09$ & $82.16\pm1.16$ \\
Nvidia Nemotron & $83.00\pm1.28$ & $82.71\pm1.34$ & $84.54\pm1.41$ \\
Ours (Resample) & $28.50\pm0.59$ & $28.02\pm0.64$ & $29.71\pm0.70$ \\
\bottomrule
\end{tabular}
\end{table}

\FloatBarrier
\section{Frontier-Model Evaluation}
\label{app:frontier_model}

We evaluate GPT-5.5 and Qwen2.5-72B with 1,000 randomly sampled Nvidia Nemotron personas per model, using all 50 IPIP items.
Table~\ref{tab:frontier_model} reports the results.
GPT-5.5 yields lower errors on all four metrics.

\begin{table}[htbp]
\centering
\caption{IPIP evaluation of Nvidia Nemotron personas with two response models. Lower is better.}
\label{tab:frontier_model}
\small
\begin{tabular}{@{}lcccc@{}}
\toprule
Response model & AMW & FD & SW & MMD \\
\midrule
Qwen2.5-72B & 0.2645 & 0.8316 & 0.3199 & 0.4414 \\
GPT-5.5 & 0.2481 & 0.7014 & 0.2617 & 0.3572 \\
\bottomrule
\end{tabular}
\end{table}

\end{document}